\documentclass[10pt,journal,compsoc]{IEEEtran}
\usepackage{graphicx}
\usepackage{longtable}
\usepackage{caption}
\usepackage{amsmath,amsthm}
\usepackage{multirow}
\usepackage{multicol}
\usepackage{booktabs}
\usepackage{url}
\usepackage{pifont}
\usepackage{color}
\usepackage{supertabular}

\theoremstyle{definition}
\newtheorem{definition}{Definition}[section]

\ifCLASSOPTIONcompsoc
  \usepackage[nocompress]{cite}
\else
  \usepackage{cite}
\fi
\ifCLASSINFOpdf
\else
\fi
\newcommand{\sm}[1]{{#1}}
\newcommand{\bb}[1]{{#1}}
\newcommand{\ad}[1]{{#1}}
\newcommand{\gc}[1]{{#1}}
\newcommand{\rev}[1]{{#1}}
\newcommand{\minor}[1]{{#1}}
\newcommand{\myparagraph}[1]{\smallskip \noindent \textbf{#1}}

\usepackage{amsfonts}

\usepackage[normalem]{ulem}

\begin{document}
%
\title{Domain Knowledge Alleviates Adversarial Attacks in Multi-Label Classifiers}
  \author{Stefano~Melacci,~\IEEEmembership{Member,~IEEE,}
        Gabriele~Ciravegna,
        Angelo~Sotgiu, 
        Ambra~Demontis,~\IEEEmembership{Member,~IEEE,} 
        Battista~Biggio,~\IEEEmembership{Member,~IEEE,}
        Marco~Gori,~\IEEEmembership{Fellow,~IEEE,}
        and~Fabio~Roli,~\IEEEmembership{Fellow,~IEEE}
\IEEEcompsocitemizethanks{\IEEEcompsocthanksitem Stefano Melacci and Marco Gori are with the Department
of Information Engineering and Mathematics, University of Siena, Siena,
Italy. Marco Gori is also with MAASAI, Universitè Côte d’Azur, Nice, France. Gabriele Ciravegna is with the Department of Information Engineering, University of Florence, Florence, Italy. Angelo Sotgiu, Ambra Demontis, Battista Biggio, and Fabio Roli are with the Department of Information and Electric Engineering, University of Cagliari, Cagliari, Italy.\protect\\
E-mail: mela@diism.unisi.it, gabriele.ciravegna@unifi.it,
\{angelo.sotgiu, ambra.demontis, battista.biggio\}@unica.it, marco.gori@unisi.it, roli@unica.it}
\thanks{\textbf{Accepted for publication in IEEE Transactions on Pattern Analysis and Machine Intelligence, DOI: 10.1109/TPAMI.2021.3137564.} © 20XX IEEE.  Personal use of this material is permitted.  Permission from IEEE must be obtained for all other uses, in any current or future media, including reprinting/republishing this material for advertising or promotional purposes, creating new collective works, for resale or redistribution to servers or lists, or reuse of any copyrighted component of this work in other works.}}

%
%

\markboth{IEEE Transactions on Pattern Analysis and Machine Intelligence (accepted), DOI: 10.1109/TPAMI.2021.3137564}%
{Shell \MakeLowercase{\textit{et al.}}: Bare Demo of IEEEtran.cls for Computer Society Journals}
%



\IEEEtitleabstractindextext{%
\begin{abstract}
Adversarial attacks on machine learning-based classifiers, along with defense mechanisms, have been widely studied in the context of single-label classification problems.
In this paper, we shift the attention to multi-label classification, where the availability of domain knowledge on the relationships among the considered classes may offer a natural way to spot incoherent predictions, i.e., predictions associated to adversarial examples lying outside of the training data distribution.
We explore this intuition in a framework in which first-order logic knowledge is converted into constraints and injected into a semi-supervised learning problem. Within this setting, the constrained classifier learns to fulfill the domain knowledge over the marginal distribution, and can naturally reject samples with incoherent predictions.
Even though our method does not exploit any knowledge of attacks during training, our experimental analysis surprisingly unveils that domain-knowledge constraints can help detect adversarial examples effectively, especially if such constraints are not known to the attacker. 
\sm{We show how to implement an adaptive attack exploiting knowledge of the constraints and, in a specifically-designed setting, we provide experimental comparisons with popular state-of-the-art attacks.}
We believe that our approach  may provide a significant step towards designing more robust multi-label classifiers.
\end{abstract}

}

\maketitle

\IEEEdisplaynontitleabstractindextext

%
\IEEEpeerreviewmaketitle

\IEEEraisesectionheading{
\section{Introduction}\label{sec:intro}
}

\ad{Despite the impressive results reported in many different application domains, machine-learning algorithms have been shown to be easily misled by adversarial examples, i.e., input samples carefully perturbed to cause misclassifications at test time \cite{szegedy14-iclr}, \cite{biggio13-ecml}.}
In the last few years, a growing number of studies on the properties of adversarial attacks\footnote{Even though every attack is adversarial by definition, we use the term \emph{adversarial attacks} here to refer to the attack algorithms used to generate adversarial examples against machine-learning models.}
and of the corresponding defenses have been produced by the scientific community~\cite{papernot2016transferability,goodfellow2018making,carlini2019evaluating,shafahi2018are,sotgiu2020deep} (see \cite{miller2020adversarial,biggio2018wild} for an in-depth review on this topic). 
Most of the existing approaches work in the fully-supervised learning setting, either proposing methods for improving classifier robustness by modifying the learning algorithm to explicitly account for the presence of adversarial data perturbations~\cite{goodfellow15-iclr,papernot2016distillation,sinha2018certifying}, or developing specific detection mechanisms for adversarial examples~\cite{carlini2017adversarial,ma2018characterizing,samangouei2018defensegan,pang2018towards,lee2018simple,miller2020adversarial}. \ad{Those approaches are developed against a specific class of attacks and usually are not robust against adversarial examples generated with different techniques~\cite{araujo20-arxiv}.}
Only a few approaches \bb{leverage also unlabeled data to improve adversarial robustness}~\cite{miyato16,park2018adversarial,akcay2018ganomaly,carmon2019unlabeled,miyato2018virtual,zhai2019adversarially,najafi2019robustness,alayrac2019arelabels}, although the semi-supervised learning setting provides a natural scenario for real-world applications in which labeling data is costly while unlabeled samples are readily available. 
More importantly, 
the case of multi-label classification, in which each sample can belong to more classes, is only preliminary discussed in the context of adversarial learning in~\cite{8594975}, while using adversarial examples to improve \ad{the accuracy on legitimate (non-adversarial) samples of} some multi-label classifiers is studied in~\cite{wu2017adversarial,babbar2018adversarial}.

In this paper, we focus on multi-label classification and, in particular, in the case in which domain knowledge on the relationships among the considered classes is available. Such knowledge can be naturally expressed by First-Order Logic (FOL) clauses, and, following the learning framework of~\cite{gnecco2015foundations,diligenti2017semantic}, it can be used to improve the classifier by enforcing FOL-based constraints on the unsupervised or partially labeled portions of the training set. 
A well-known intuition in adversarial machine learning suggests that a reliable model of the distribution of the data could be used to spot adversarial examples, being them not sampled from such distribution, but it is not a straightforward procedure~\cite{grosse2017statistical}. 
We borrow such intuition and we intersect it with the idea that semi-supervised examples can help  learn decision boundaries that better follow the marginal data distribution, coherently with the available knowledge~\cite{melacci2011primallapsvm,diligenti2017semantic},
\sm{and we investigate the role of such knowledge in the context of data generated in an adversarial manner.}
\rev{While the generic idea of considering domain information in adversarial attacks has been recently followed by other authors to different extents \cite{A,B,sheatsleyrobustness}, to the best of our knowledge 
we are the first to use domain knowledge expressed in FOL and converted into polynomial constraints to improve adversarial robustness of multi-label classifiers.}

\begin{figure}
\centering
\rev{{\includegraphics[width=0.49\textwidth]{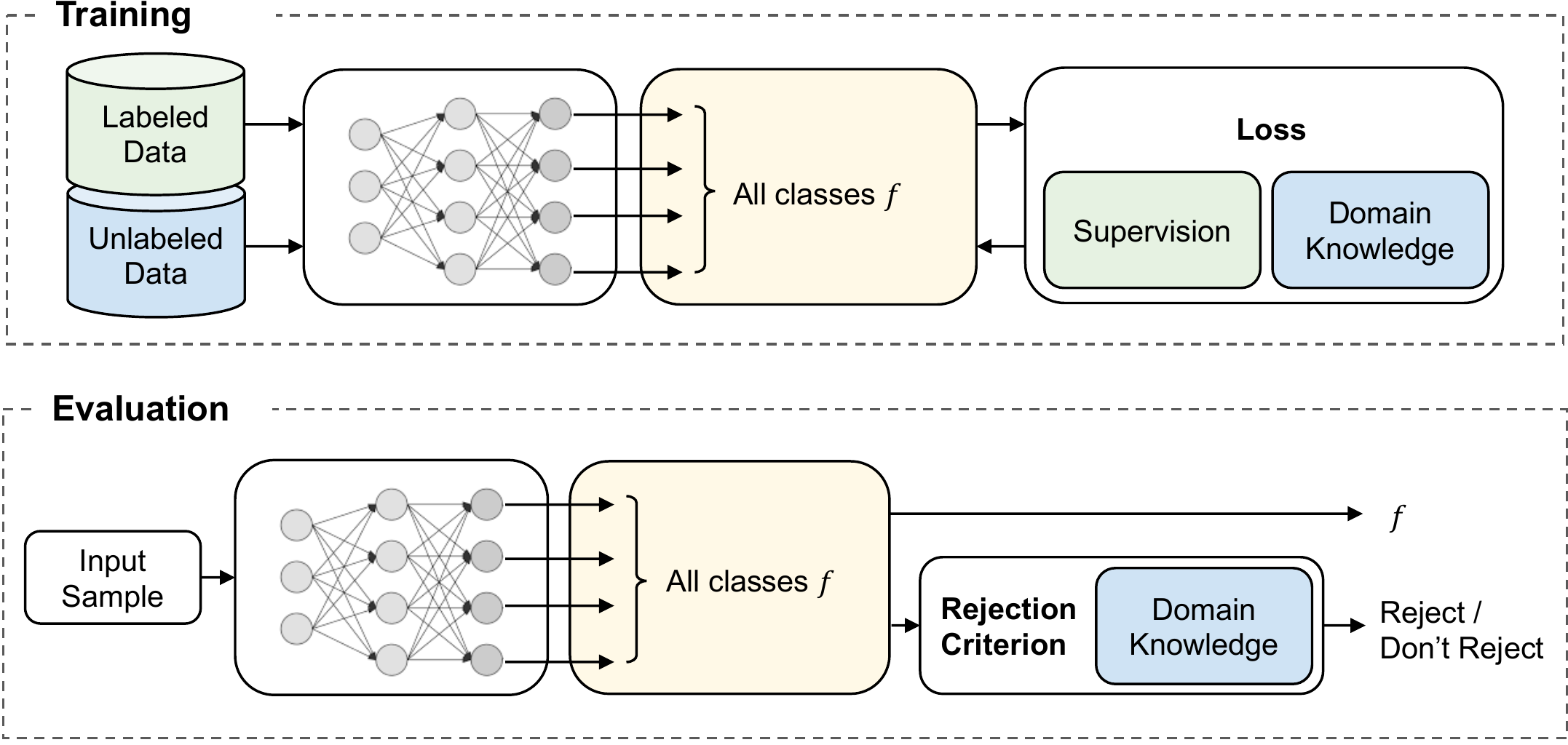}}}
\caption{Leveraging domain knowledge to improve robustness of multi-label classifiers. At training time, domain knowledge is used to enforce constraints on the learning process using unlabeled or partially-labeled data. At evaluation time, domain-knowledge constraints are used to detect and reject samples outside of the training data distribution.} 
\label{fig:schema}
\end{figure}

\sm{In detail,} this paper contributes in showing that domain knowledge is a powerful feature ($i$) to improve robustness of multi-label classifiers, and ($ii$) to help detect adversarial examples. 
\bb{The underlying idea of our approach is conceptually represented in Fig.~\ref{fig:schema}. At training time, domain-knowledge constraints are enforced on the unlabeled (or partially-labeled)  data to learn decision boundaries which better align with the marginal distributions. At test time, the same constraints can be efficiently evaluated on the test samples to identify and reject incoherent predictions, ideally outside of the training data distribution, potentially including adversarial examples. Our approach can be also used in single-classification tasks where domain knowledge and auxiliary classes are present, and can be exploited internally by the classifier to implement the rejection mechanism based on domain-knowledge constraints. We will show some concrete examples of this latter setting in our experiments, reporting comparisons with state-of-the-art adversarial attacks and concurrent defenses developed for single-classification tasks.}
To properly evaluate the robustness of our approach, which remains one of the most challenging problems in adversarial machine learning~\cite{carlini19-arxiv,athalye18,biggio2018wild}, \rev{we propose a novel multi-label attack that can implement both black-box and white-box adaptive attacks, being driven by the domain knowledge in the latter case}. While we show that an adaptive attack having access to the domain knowledge exploited by our classifier can bypass it, even though at the cost of an increased perturbation size, it remains an open issue to understand how hard for an attacker would be to infer such  knowledge in practical cases.
For this reason, we believe that our work can provide a significant contribution towards both evaluating and designing robust multi-label classifiers.



\sm{This paper is organized as follows. Section~\ref{sec:domain} introduces the  notion of learning with domain knowledge, emphasizing its effects in the input space. Section~\ref{sec:adv} shows how domain knowledge can be used to defend against adversarial attacks, together with a knowledge-aware attack procedure. A detailed experimental analysis is reported in Section~\ref{sec:exp}, evaluating the quality of our defense mechanisms, also considering state-of-the-art attacks and existing defense schemes. \minor{Related work is discussed in Section~\ref{sec:rel} and,} finally, conclusions are drawn in Section~\ref{sec:concl}.}

\section{Learning with Domain Knowledge}
\label{sec:domain}
\sm{In the context of this paper we focus on multi-label classification problems with $c$ classes, in which each input $x \in \mathcal{X}$ is associated to one or more classes. We consider the case in which additional {\it domain knowledge} is available for problem at hand, represented by a set of relationships that are known to exist among (a subset of) the $c$ classes. Exploiting such knowledge when training the classifier is the main subject of this section, and it has been shown to improve the generalization skills of the model \cite{gnecco2015foundations,DBLP:journals/tnn/GoriM13,diligenti2017semantic}. The introduction of domain knowledge in the learning process provides precious information only when the training data are not fully labeled, as in the classic semi-supervised framework. Some examples might be partially labeled (i.e., for each data points a subset of the $c$ classes participates to the ground truth) or a portion of the training set might be unsupervised. Of course, if the data are fully labeled then all the class relationships are already encoded in the supervision signal. 
However, in this paper we also consider domain knowledge as a mean to define a criterion that can spot potentially adversarial examples at test time, as we will discuss in  Section~\ref{sec:adv}, and that is also feasible in fully-supervised learning problems.}

\myparagraph{Notation.} Formally, we consider a vector function $f:\ \mathcal{X} \rightarrow \mathbb{R}^{c}$, where $f = \left[ f_1, \ldots, f_c \right]$ and $\mathcal{X} \subseteq \mathbb{R}^{d}$. 
Each function $f_i$ is responsible for implementing a specific task on the input domain $\mathcal{X}$.\footnote{This notion can be trivially extended to the case in which the task functions operate in different domains.} 
\rev{In a classification problem,} function $f_i$ predicts the membership degree of $x$ to the $i$-th class. Moreover, when we restrict the output of $f_i$ to $[0,1]$, we can think of $f_i$ as the fuzzy logic predicate that models the truth degree of belonging to class $i$. In order to simplify the notation, we will frequently make no explicit distinctions between function names, predicate names, class names or between input samples and predicate variables.
Whenever we focus on the predicate-oriented interpretation of each $f_i$, First-Order Logic (FOL) becomes the natural way of describing relationships among the classes, i.e., the most effective type of domain knowledge that could be eventually available in a multi-label problem; e.g., $\forall x \in \mathcal{X},\ f_v(x) \land f_z(x) \Rightarrow f_u(x)$, for some $v,z,u$, meaning that the intersection between the $v$-th class and the $z$-th class is always included in the $u$-one. \rev{Table \ref{tab:notation} reports a summary of the notation described so far and of the main symbols that will be introduced in what follows.}
\begin{table}[ht]
    \centering
    \caption{\rev{List of the main symbols and notations.}}
    \label{tab:notation}
    \rev{\begin{tabular}{@{\hskip0.01pt}l@{\hskip2pt}|@{\hskip3pt}l}
    \toprule
    \textit{Notation} & \scriptsize \textit{Description}\\
    \midrule
     $x \in \mathcal{X} \subseteq \mathbb{R}^{d}$ & \scriptsize Input sample.\\
     $c$ & \scriptsize Number of classes.\\
     $f_i(x)$ & \scriptsize Membership score to the $i$-th class/fuzzy logic predicate.\\
     $\textsc{NAME}(x)$ & \scriptsize Same as $f_i$, using the name of the class instead of $f_i$.\\
     $f = [f_1 \ldots f_c]$ & \scriptsize Vector function collecting all the $f_i$'s.\\
    $f(\cdot, W)$ & \scriptsize Network that implements $f$, with weights collected in $W$.\\
    $\mathcal{K}$ & \scriptsize Domain knowledge (FOL formulas). \\      
    $\ell$ & \scriptsize Number of formulas in $\mathcal{K}.$\\    
     $\land,\ \lor,\ \neg,\ \Rightarrow$ & \scriptsize Logical connectives. \\    
    $\phi(f(x)) = 1$ & \scriptsize T-Norm constraint from a generic FOL formula (in $x$).\\
    $\hat{\phi}(f(x))$ & \scriptsize Penalty ($\geq 0$) associated to constraint $\phi(f(x)) = 1.$\\   
    $\hat{\phi}_h(f(x))$ & \scriptsize Penalty ($\geq 0$) associated to the T-Norm-based  \\
    & \scriptsize constraint from the $h$-th formula in $\mathcal{K}$.\\  
    $\varphi(f, \mathcal{Z}, \mathcal{K}, \mu)$ & \scriptsize Average penalty associated to the violations of the formulas \\
    & \scriptsize  in $\mathcal{K}$ weighed  by the scalars in $\mu$, and evaluated on data $\mathcal{Z}$.\\
    $\mathcal{L},\ \mathcal{V},\ \mathcal{T}$ & \scriptsize Training, validation, and test data (respectively). \\
    $\mathcal{S}$ & \scriptsize Supervision attached to (some) of the data in $\mathcal{L}.$\\
    $\texttt{suploss}$ & \scriptsize Loss function, supervised data only. \\
    $\overline{x}$ & \scriptsize \scriptsize Test sample (unused in training). \\
    $\mu^{\mathcal{L}},\ \mu^{\mathcal{T}}$ & \scriptsize Weights of the FOL formulas when $\varphi$ is evaluated on   \\
    & \scriptsize  training or test data, respectively.\\
    $\lambda$ & \scriptsize Scalar ($\geq 0$) that weighs $\varphi$ in the learning criterion. \\
    $\Omega$ & \scriptsize Rejection criterion. \\
    $\tau$ & \scriptsize Rejection threshold ($>0$).\\
    $\epsilon$ & \scriptsize Upper bound on the norm of the difference between a \\ & \scriptsize  clean example and its perturbed instance.\\
    $l_p(x)$ & \scriptsize Logits ($>-\kappa$) of the positive class with the tiniest output. \\
    $l_n(x)$& \scriptsize Logits ($<\kappa$) of the negative class with the largest output. \\ 
    $\kappa$ & \scriptsize Threshold ($\geq 0$) used to bound the logits. \\
    $\alpha$ & \scriptsize Scalar ($\geq 0$) that weights $\varphi$ in our white-box attack.\\
    \bottomrule
    \end{tabular}}
\end{table}

\myparagraph{Learning from Constraints.} The framework of Learning from Constraints~\cite{gnecco2015foundations,DBLP:journals/tnn/GoriM13,diligenti2017semantic} follows the idea of converting domain knowledge into constraints on the learning problem and it studies, amongst a variety of other knowledge-oriented constraints (see, e.g., Table 2 in \cite{gnecco2015foundations}), the process of handling FOL formulas so that they can be both injected into the learning problem or used as a knowledge verification measure~\cite{DBLP:journals/tnn/GoriM13,diligenti2017semantic}. 
Such knowledge is enforced on those training examples for which either no information or only partial/incomplete labeling is available, thus casting the learning problem in the semi-supervised setting. 
As a result, the multi-label classifier can improve its performance and make predictions on out-of-sample data that are more coherent with the domain knowledge (see, e.g., Table 4 in \cite{gnecco2015foundations}).
In particular, FOL formulas that represent the domain knowledge of the considered problem are converted into numerical constraints using Triangular Norms (T-Norms, \cite{klement2013triangular}), binary functions that generalize the conjunction operator $\land$. Following the previous example, $f_v(x) \land f_z(x) \Rightarrow f_u(x)$ is converted into a bilateral constraint $\phi(f(x))=1$ that, in the case of the product T-Norm, is $1-f_v(x)f_z(x)(1-f_u(x))=1$. The $1$ on the right-hand side of the constraint is due to the fact that the numerical formula must hold true (i.e, $1$), while the left-hand side is in $[0,1]$. We indicate with $\hat{\phi}(f(x))$ the loss function associated to  $\phi(f(x))$. In the simplest case (\sm{the one} followed in this paper) such loss is $\hat{\phi}(f(x)) = 1-\phi(f(x))$, where the minimum value of $\hat{\phi}(f(x))$ is zero. The quantifier $\forall x \in \mathcal{X}$ is translated by enforcing the constraints on a discrete data sample $\mathcal{Z} \subset \mathcal{X}$. The loss function $\varphi(f,\mathcal{Z})$ associated to \sm{all the available FOL formulas} $\mathcal{K}$ is obtained by \sm{aggregating the losses of all the corresponding constraints} and averaging over the data in $\mathcal{Z}$.
Since we usually have $\ell > 1$ formulas whose relative importance could be uneven, we get
\begin{equation}
\varphi(f,\mathcal{Z},\mathcal{K},\mu) = \frac{1}{| \mathcal{Z} |}\sum_{j=1}^{| \mathcal{Z} |} \left( \sum_{h=1}^{\ell} \mu_h \hat{\phi}_h(f(x_j)) \right) \in [0, \gamma]
\label{eq:closs}
\end{equation}
where $\mu$ is the vector that collects the scalar weights $\mu_h > 0$ of the FOL formulas, and $\gamma = \sum_{h=1}^{\ell} \mu_h$.

In this paper, $f$ is implemented with a neural architecture with $c$ output units and weights collected in $W$. We distinguish between the use of Eq. (\ref{eq:closs}) as a loss function in the training stage and its use as a measure to evaluate the constraint fulfillment on out-of-sample data. In detail, the classifier is trained on the training set $\mathcal{L}$ by minimizing
\begin{equation}
\min_{W} \left[ \texttt{suploss}(f(\cdot,W),\mathcal{L},\mathcal{S}) + \lambda \cdot \varphi(f(\cdot,W),\mathcal{L},\mathcal{K},\mu^{\mathcal{L}}) \right]
\label{train}
\end{equation}
where $\mu^{\mathcal{L}}$ is the importance of the FOL formulas at training time, and $\lambda > 0$ modulates the weight of the constraint loss with respect to the supervision loss \texttt{suploss}, being  $\mathcal{S}$ the supervision information attached to some of the data in $\mathcal{L}$. The optimal $\lambda$ is chosen by cross-validation, maximizing the classifier performance. When the classifier is evaluated on a test sample $\overline{x}$, the measure
\begin{equation}
\varphi(f(\cdot,W),\{\overline{x}\},\mathcal{K},\mu^{\mathcal{T}}) \in [0,\gamma^{\mathcal{T}}],
\label{test}
\end{equation}
with weights $\mu^{\mathcal{T}}$ and $\gamma^{\mathcal{T}}=\sum_{h=1}^{\ell}\mu_h^{\mathcal{T}}$, returns a score that indicates the fulfillment of the domain knowledge on $\overline{x}$ (the lower the better). 
Note that $\mu^{\mathcal{L}}$ and $\mu^{\mathcal{T}}$ might not necessarily be equivalent, even if certainly related. In particular, one may differently weigh the importance of some formulas during training to better accommodate the gradient-descent procedure and avoid bad local minima.

It is important to notice that Eq.~\eqref{train} enforces domain knowledge only on the training data $\mathcal{L}$. There are no guarantees that such knowledge will be fulfilled in the whole input space $\mathcal{X}$. This suggests that optimizing Eq.~\eqref{train} yields a stronger fulfillment of knowledge $\mathcal{K}$ over the space regions where the training points are distributed (low values of $\varphi$), while $\varphi$ could return larger values when departing from the distribution of the training data. The constraint enforcement is soft, so that the second term in Eq.~\eqref{train} is not necessarily zero at the end of the optimization. 

\section{Exploiting Domain Knowledge against Adversarial Attacks}
\label{sec:adv}
The basic idea behind this paper is that the constraint loss of Eq.~\eqref{eq:closs} is not only useful to enforce domain knowledge into the learning problem, but also (\textit{i}) to gain some robustness with respect to adversarial attacks and (\textit{ii}) as a tool to detect adversarial examples at no additional training cost, \sm{that are the main directions of this paper, as anticipated in Section~\ref{sec:intro}}.

\begin{figure*}
\centering
\hskip -0.3cm
\fbox{\includegraphics[width=0.23\textwidth]{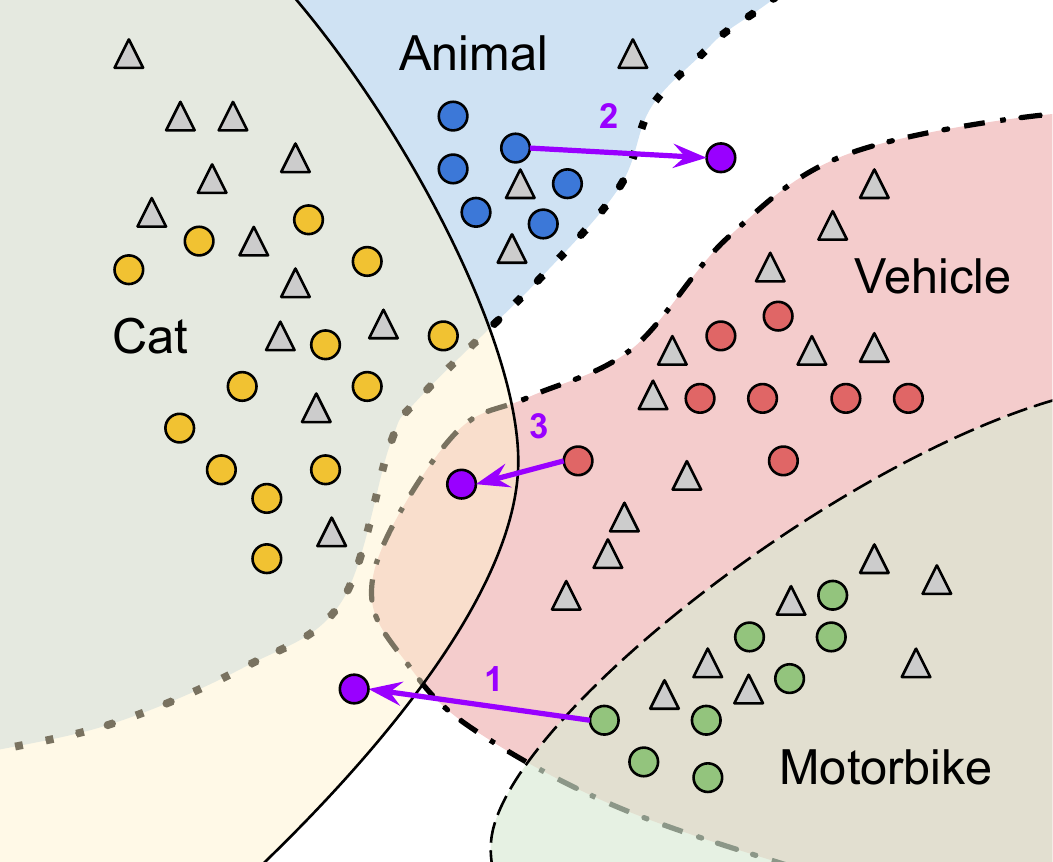}} 
\fbox{\includegraphics[width=0.23\textwidth]{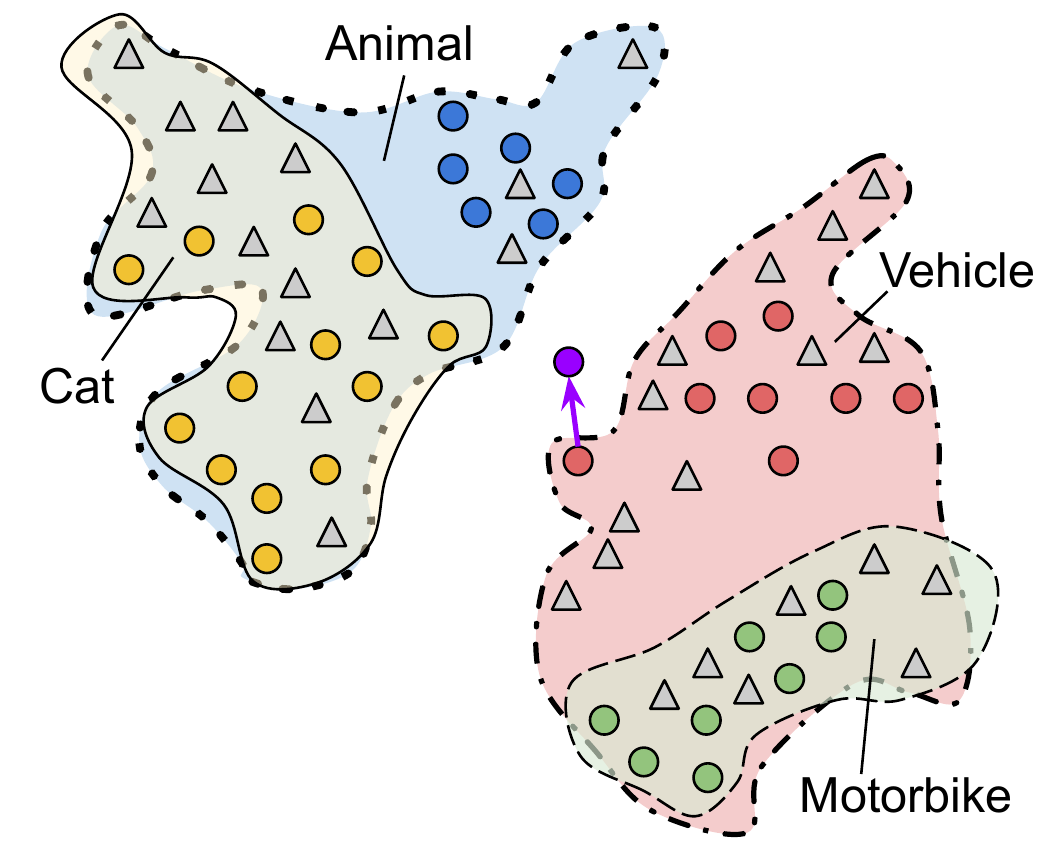}}
\rev{\fbox{\includegraphics[width=0.23\textwidth]{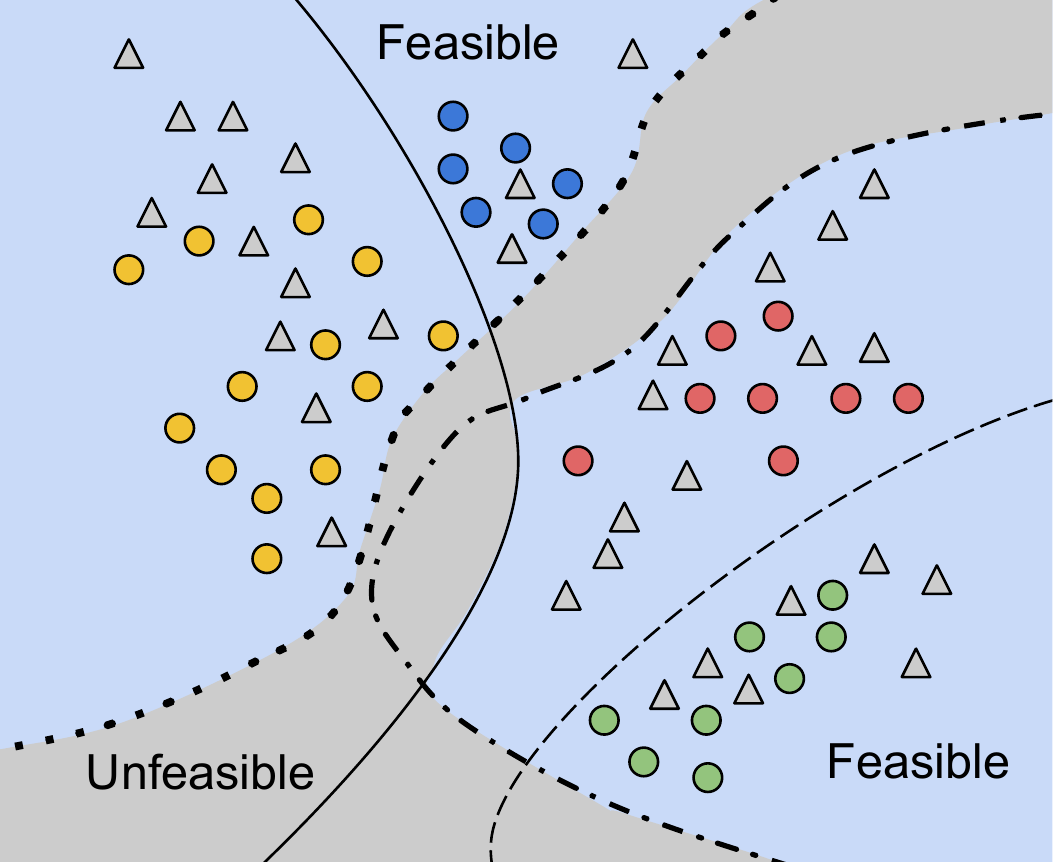}}}
\rev{\fbox{\includegraphics[width=0.23\textwidth]{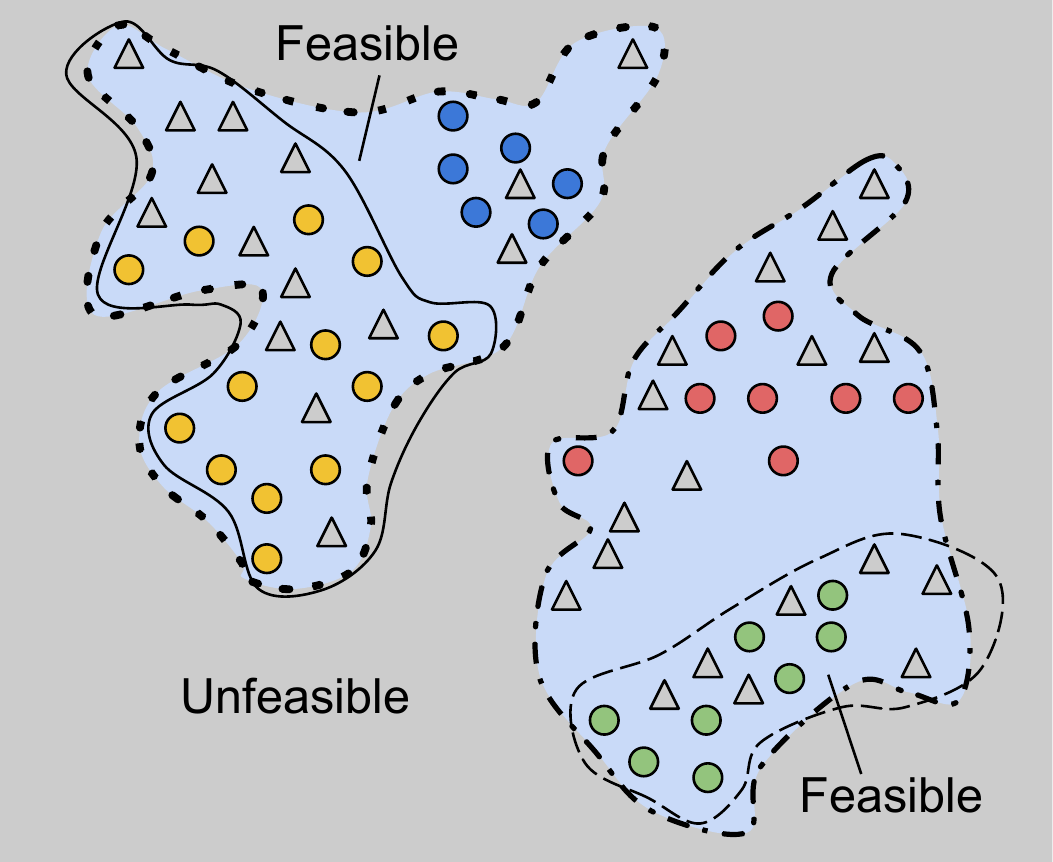}}}
\vskip 1mm
\hskip -0.5cm (a) \hskip 3.8cm (b) \hskip 4.1cm (c) \hskip 4.1cm (d)
\caption{Toy example using the domain knowledge of Eqs.~(\ref{toy1}-\ref{toy4}) on 4 classes: cat (yellow), animal (blue), motorbike (green), vehicle (red). 
Labeled/unlabeled training data are depicted with rounded dots/gray triangles. (a,b) The decision regions for each class are shown in two sample outcomes of the training procedure: (a) open/loose decision boundaries; (b) tight/closed decision boundaries. The white area is associated with no predictions. Some adversarial examples (purple arrows/dots) are detected as they end up in regions that violate the constraints. Moreover, in (c,d) The feasible/unfeasible regions \rev{(blue/gray)} that fulfill/violate the constraints for (a,b) are shown. \rev{Decision boundaries of the classes in (a,b) are also depicted in (c,d)}.} 
\label{fig:toy}
\end{figure*}

\myparagraph{A Paradigmatic Example.} The example in Fig.~\ref{fig:toy} illustrates the main principles followed in this work, in a multi-label classification problem with $4$ classes (cat, animal, motorbike, vehicle) for which the following domain knowledge is available, together with labeled and unlabeled training data:
\begin{eqnarray}
\label{toy1}\hskip -4mm\forall x, & &\hskip -6mm \footnotesize \text{CAT}(x) \Rightarrow \text{ANIMAL}(x) \, , \\
[-0.5mm]\label{toy2}\hskip -4mm\forall x, & &\hskip -6mm \footnotesize \text{MOTORBIKE}(x) \Rightarrow \text{VEHICLE}(x) \, , \\
[-0.5mm]\label{toy3}\hskip -4mm\forall x, & &\hskip -6mm \footnotesize \text{VEHICLE}(x) \Rightarrow \neg \text{ANIMAL}(x) \, , \\
[-0.5mm]\label{toy4}\hskip -4mm\forall x, & &\hskip -6mm \footnotesize \text{CAT}(x) \lor \text{ANIMAL}(x) \lor \text{MOTORBIKE}(x) \lor \text{VEHICLE}(x).
\end{eqnarray}
Such knowledge is converted into numerical constraints, as described in Section~\ref{sec:domain}, while the loss function $\varphi$ is enforced on the training data predictions during classifier training (Eq.~\ref{train}).
Fig.~\ref{fig:toy} shows two examples of the learned classifier. 

Considering point (\textit{i}), in both cases, the decision boundaries are altered on the unlabeled data, enforcing the classifier to take a knowledge-coherent decision over the unlabeled training points and to better cover the marginal distribution of the data. This knowledge-driven regularity improves classifier robustness to adversarial attacks, as we will discuss in Section~\ref{sec:exp}.
Going into further details to illustrate claim (\textit{ii}), in (a) we have the most likely case, in which decision boundaries are not always perfectly tight to the data distribution, and they might be not closed (ReLU networks typically return high-confidence predictions far from the training data~\cite{hein2019relu}). Three different attacks are shown (purple). In attack $1$, an example of motorbike is perturbed to become an element of the cat class, but Eq.~\eqref{toy1} is not fulfilled anymore. In attack $2$, an example of animal is attacked to avoid being predicted as animal. However, it falls in a region where no predictions are yielded, violating Eq.~\eqref{toy4}. Attack number $3$ consists of an adversarial attack to create a fake cat that, however, is also predicted as vehicle, thus violating Eq.~\eqref{toy1} and Eq.~\eqref{toy3}. In (b) we have an ideal and extreme case, with very tight and closed decision boundaries. Some classes are well separated, it is harder to generate adversarial examples by slightly perturbing the available data, while it is easy to fall in regions for which Eq.~\eqref{toy4} is not fulfilled. The pictures in (c\sm{-d}) show the unfeasible regions in which the constraint loss $\varphi$ is significantly larger, thus offering a natural criterion to spot adversarial examples that fall outside of the training data distribution. 

\myparagraph{Domain Knowledge-based Rejection.} Following these intuitions, and motivated by the approach of \cite{hendrycks17baseline,hendrycks2016early}, we define a rejection criterion $\Omega$ as the Boolean expression
\begin{equation}
\Omega(\overline{x}, \tau | f(\cdot,W),\mathcal{K},\mu^{\mathcal{T}}) = \varphi(f(\cdot,W),\{\overline{x}\},\mathcal{K},\mu^{\mathcal{T}}) > \tau
\label{eq:reject}
\end{equation}
where $\tau > 0$ is estimated by cross-validation in order to avoid rejecting (or rejecting a small number of\footnote{10\% in our experiments.}) the examples in the validation set $\mathcal{V}$.
Eq. (\ref{eq:reject}) \sm{evaluates} the constraint loss on the validation data $\mathcal{V}$, using the importance weights $\mu^{\mathcal{T}}$ (that we will discuss in what follows), as in Eq. (\ref{test}). The rationale behind this idea is that those samples for which the constraint loss is larger than what it is on the distribution of the  data \sm{that are available when training/tuning the classifier,} should be rejected. The training samples are the ones over which domain knowledge was enforced during the training stage, while the validation set represents data on which knowledge was not enforced, but that are sampled from the same distribution from which the training set is sampled, making them good candidates for estimating $\tau$. \sm{Notice that $\Omega$ is measured at test time on an already trained classifier, and it can be used independently on the nature of the training data (fully or partially/semi-supervised).}
Differently from ad-hoc detectors, that usually require to train generative models, this rejection procedure comes at no additional training cost.\footnote{Generative models on the fulfillment of the single constraints could be considered too.} 

\myparagraph{Pairing Effect.} The procedure is effective whenever the functions in $f$ are not too strongly paired with respect to $\mathcal{K}$, and we formalize the notion of ``pairing'' as follows.
\begin{definition}
\textit{Pairing.} We consider a classification problem whose training data are distributed accordingly to the probability density $p(x)$. Given $\mathcal{K}$ and $\mu^{\mathcal{T}}$, the functions in $f$ are strongly paired whenever $\zeta(\mathcal{H},\mathcal{L}) = \| \varphi(f(\cdot,W),\mathcal{H},\mathcal{K},\mu^{\mathcal{T}}) - \varphi(f(\cdot,W),\mathcal{L},\mathcal{K},\mu^{\mathcal{T}})  \| \approx 0$, being $\mathcal{H}$ a discrete set of samples uniformly distributed around the support of $p(x)$.
\label{pairing}
\end{definition}
This notion indicates that if the constraint loss is fulfilled in similar ways over the training data distribution and  space areas close to it, then there is no room for detecting those examples that should be rejected. While it is not straightforward to evaluate pairing before training the classifier, %
the soft constraining scheme of Eq. (\ref{train}) allows the classification functions to be paired in a less strong manner that what they would be when using hard constraints.\footnote{See \cite{teso2019does} for a discussion on hard constraints and graphical models in an adversarial context.}
Note that a multi-label system is usually equipped with activation functions that do not structurally enforce any dependencies among the classes (e.g., differently from what happens with softmax), so it is naturally able to respond without assigning the input to any class (white areas in Fig.~\ref{fig:toy}). This property has been recently discussed as a mean for gaining robustness to adversarial examples~\cite{shafahi2018are,bendale2016towards}. The formula in Eq.~\eqref{toy4} is what allows our model to spot examples that might fall in this ``I don't know'' area. Dependencies among classes are only introduced by the constraint loss $\varphi$ in Eq.~\eqref{train} on the training data.

The choice of $\mu^{\mathcal{T}}$ is crucial in the definition of the reject function $\Omega$. On the one hand, in some problems we might have access to the certainty degree of each FOL formula, that could be used to set $\mu^{\mathcal{T}}$, otherwise it seems natural to select an unbiased set of weights $\mu^{\mathcal{T}}$, $\mu_h=1$, $\forall h$. On the other hand, several FOL formulas involve the implication operator $\Rightarrow$, that naturally implements if-then rules (if class $v$ then class $z$) or, equivalently, rules that are about hierarchies, since $\Rightarrow$ models an inclusion (class $v$ included in class $z$). However, whenever the premises are false, the whole formula holds true. It might be easy to trivially fulfill the associated constrains by zeroing all the predicates in the premises, eventually avoiding rejection. As rule of thumb, it is better to select $\mu_h$'s that are larger for those constraints that favor the activation of the involved predicates.

\sm{\myparagraph{Single-label Classifiers.} \rev{The type of domain knowledge described so far usually involves logic formulas that encode relationships among multiple classes, thus it is  naturally associated with multi-label problems. 
Let us focus our attention on multi-label scenarios in which there exists a subset of categories that are known to be mutually exclusive, that we will refer to as \textit{main classes}, while the remaining categories will be referred to as \textit{auxiliary classes}. 
If we restrict the original classification problem to the main classes only, we basically end-up in a single-label scenario. 
Let us assume that the available logic formulas introduce relationships between (some of) the main classes and (some of) the auxiliary ones. As a result, in order to setup our defense mechanism (Eq.~\ref{eq:reject}) or to learn with domain knowledge (Eq.~\ref{train}), predictions on both the main and auxiliary classes must be available, so that the truth degree of the logic formulas can be evaluated.
This consideration can be exploited to design classifiers that expose single label predictions on the main classes, thus acting as single-label classifiers, and include predictions on the auxiliary classes that are not exposed to the user at all, but that are internally used to setup our defense mechanism or to improve the quality of whole classifier when learning in a semi-supervised  context.}
\rev{Formally, } let us assume that the components $\{ f_i, i = 1,\ldots,c \}$ of the vector function  $f$ are partitioned into two disjoint subsets, \rev{where the first one considers the components about the mutually-exclusive main classes and the second subset is about the auxiliary classes}. We define with $f^{v}$ the vector function with the elements in the first subset, while $f^{h}$ is the vector function based on the elements of the second one, \rev{as shown in Fig.~\ref{fig:single}}.
\begin{figure}
\centering
\rev{{\includegraphics[width=0.49\textwidth]{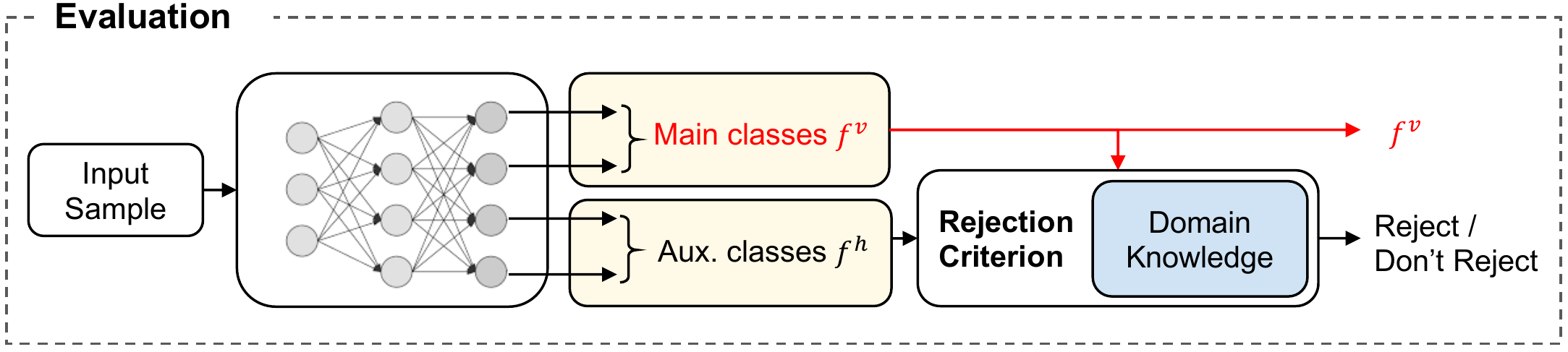}}}
\caption{\rev{Single-label classifier on a set of mutually exclusive classes (\emph{main classes}), computing the class activations by $f^{v}$ and exposing them to the user (red path). It internally computes by $f^{h}$ additional predictions over \emph{auxiliary classes} that are involved in the domain knowledge (together with the main classes). Training considers \emph{all} the classes,  Fig.~\ref{fig:schema}.}} 
\label{fig:single}
\end{figure}
\rev{The system only exposes to the user predictions computed by means of $f^{v}$, while the computations of $f^{h}$ are hidden. Overall, }
the system can still exploit domain knowledge that consists in relationships between the classes associated to $f^{v}$ \rev{(main classes)} and the ones associated to $f^{h}$ \rev{(auxiliary classes)}, or among the ones in $f^{h}$ only, thus leveraging the learning principles that were described in Section~\ref{sec:domain}. Moreover, the system can exploit the hidden predictions and the available knowledge to implement the knowledge-based rejection mechanism that we proposed in this section, \rev{as sketched in Fig.~\ref{fig:single}}.

Due to the single-label nature of the visible portion of the classifier, existing state-of-the-art attacks, specifically designed for single-label models, can be used to fool the classifier in a black-box scenario. In Section~\ref{sec:exp}, when the consider data are compatible with this special setting, we will exploit recent attack procedures to generate adversarial examples and evaluate the proposed knowledge-based rejection mechanism. Of course, differently from what we previously stated about the real multi-label setting, we cannot consider the cost of the rejection mechanism negligible in this case, since the system must learn the functions in $f^{h}$ in order to be able to compute the rejection criterion. 
}

\subsection{Attacking Multi-label Classifiers}

Robustness against adversarial examples is typically evaluated against \emph{black-box} and \emph{white-box} attacks~\cite{biggio2018wild,miller2020adversarial}. In the black-box setting, the attacker is assumed to have only black-box query access to the target model, ignoring the presence of any defense mechanisms \rev{and without having access to any additional domain knowledge and related constraints}. 
However, a surrogate model can be trained on data ideally sampled from the same distribution of that used to train the target model. Within these assumptions, gradient-based attacks can be optimized against the surrogate model, and then transferred/evaluated against the target one~\cite{papernot2016transferability,demontis19-usenix}. 
In the white-box setting, instead, the attacker is assumed to know everything about the target model, including the defense mechanism. White-box attacks are thus expected to \rev{also exploit the available domain knowledge to try to bypass the knowledge-based defense}. 

\sm{The existing literature on the generation of adversarial examples is strongly focused on single-label classification problems (see \cite{miller2020adversarial} and references therein). In such context, the classifier is expected to take a decision that is only about one of the $c$ classes, and, in a nutshell, attacking the classifier boils down to perturb the input in order to make the classifier predict a wrong class. The whole procedure is subject to constraints on the amount of perturbation that the system is allowed to apply.
Formally, given $x \in \mathcal{T}$, being $\mathcal{T}$ the test set, the attack generation procedures in single-label classification commonly solve the following problem,
\begin{equation}
\begin{aligned}
x^{\star} ={} & \arg\min_{x^{\prime}} [-\texttt{suploss}(f(x^{\prime},W),\mathcal{L},\mathcal{S})], \\
 \text{s.t.} \ \ \ & \|x-x^{\prime}\| < \epsilon,
\label{eq:pbx}
\end{aligned}
\end{equation}
being $\|\cdot\|$ an $L_p$-norm and $\epsilon > 0$.
Each $x$ has a unique class label/index attached to it and stored in $\mathcal{S}$, and \texttt{suploss} is usually the cross-entropy loss.
Different attacks and optimization techniques for solving the problem of Eq.~(\ref{eq:pbx}) have been proposed \cite{autoattack}.
While there are no ambiguities on the class on which we want the classifier to reduce its confidence, i.e., the ground-truth (positive) class of the given input $x$, the class that the classifier will predict in input $x^{\star}$ might be given or not, thus each of the remaining $c-1$ classes could  be a valid option. When moving to the multi-label setting, each $x \in \mathcal{T}$ is associated to multiple ground-truth positive classes, collected in set $P_x$, and we indicate with $N_x$ the set of ground-truth negative classes of $x$.
Differently to the previous case, due to the lack of mutual-exclusivity of the predictions, creating an adversarial example out of $x$ is more arbitrary. 
For example, the optimization procedure could focus on making the classifier not able to predict any of the classes in $P_x$, or a subset of them. Similarly, the optimization could focus on making the classifier positively predict one or more classes of $N_x$. 

\rev{Departing from the overwhelming majority of existing attacks for single-label classifiers, we propose a multi-label attack} that focuses
on the classes on which the classifier is less confident (thus easier to attack), that are selected and re-defined during the optimization procedure in function of the way the predictions of the classifier progressively change. Of course, \rev{in the \emph{black-box} case}, this attack is not considering that classes are related, and \rev{it is not taking care that}, perhaps, changing the prediction on a certain class should also trigger a coherent change in other related classes. \rev{Differently, in the \emph{white-box} setting}, the previously introduced domain knowledge and, in particular, the corresponding loss of Eq.~(\ref{eq:closs}) is what encodes such relationships in a differentiable way, so that we can easily exploit it when crafting attacks.
We first introduce the proposed multi-label attack in a {\it black-box} setting, in which domain knowledge is not available. To make gradient computation numerically more robust, as in~\cite{carlini17-sp}, we consider the activations (logits) of the last layer of $f$ to compute the objective function, instead of using the cross-entropy loss.
}
%
Let us define $p = \arg\min_i [f_i(x),\ i \in P_x]$, and $n = \arg\max_i [f_i(x),\ i \in N_x]$, i.e., $p$ ($n$) is the index of the positive (negative) class with the smallest (largest) output score. These are essentially the indices of the classes for which $x$ is closer to the decision boundaries.
Our attack optimizes the following objective,
\begin{equation}
\begin{aligned}
x^{\star} ={} & \arg\min_{x^{\prime}} [  \max(l_{p}(x^{\prime}), -\kappa) -  \min (l_{n}(x^{\prime}), \kappa)] \\
 \text{s.t.} \ \ \ & \|x-x^{\prime}\| < \epsilon,
\label{attacklogitssimple}
\end{aligned}
\end{equation}
where $l_{j}$ is the value of the logit of $f_j$, $\|\cdot\|$ is an $L_p$-norm ($L_2$ in our experiments), and in the case of image data with pixel intensities in $[0,1]$ we also have $x'\in [0,1]$. The scalar $\kappa \geq 0$ is used to threshold the values of the logits, to avoid increasing/decreasing them in an unbounded way (in our experiments, we set $\kappa = 2$).
Optimizing the logit values is preferable to avoid sigmoid saturation. While the definition of Eq.~\eqref{attacklogitssimple} is limited to a pair of classes, we {\it dynamically} update $p$ and $n$ whenever logit $l_{p}$ ($l_{n})$ goes beyond (above) the threshold $-\kappa$ ($\kappa$), thus multiple classes are considered by the attack, compatibly with the maximum number of iterations of the optimizer. This strategy resulted to be more effective than jointly optimizing all the classes in $P_x$ and $N_x$. Moreover, the classes involved in the attack can be a subset of the whole set, as in \cite{8594975}. 
\sm{\rev{In a \textit{white-box} scenario, when the attacker has the use of the domain knowledge, the information in $\mathcal{K}$} provides a comprehensive description on how the predictions of the classifier should be altered over several classes in order to be coherent with the knowledge. \rev{In such a scenario,} we enhance Eq.~(\ref{attacklogitssimple}) to implement what we refer to as} multi-label knowledge-driven adversarial attack (MKA), \rev{including the differentiable knowledge-driven loss $\varphi$ in the objective function,}
\begin{equation}
\begin{aligned}
x^{\star} ={} & \arg\min_{x^{\prime}} [  \max(l_{p}(x^{\prime}), -\kappa) -  \min (l_{n}(x^{\prime}), \kappa)  + \\ 
    & \alpha \cdot \varphi(f,\{x'\},\mathcal{K},\mu^{\mathcal{T}})],\ \ \ \text{s.t.}\ \|x-x^{\prime}\| < \epsilon 
\label{attacklogits}
\end{aligned}
\end{equation}
\sm{in which we set} $\alpha>0$ to enforce domain knowledge and avoid rejection. When crafting adversarial examples, MKA softly enforces the fulfillment of domain knowledge by means of the loss function $\varphi$. For \textit{black-box} attacks, instead, we set $\alpha=0$ \sm{to recover Eq.~(\ref{attacklogitssimple})}. MKA naturally extends the formulation of single-label attacks (when $P_x$ is composed of a single class) and it allows staging both black-box and white-box (adaptive) attacks against our approach.
Eq.~\eqref{attacklogits} is minimized via projected gradient descent ($1000$ samples and $50$ iterations in our experiments).

\rev{\subsection{Impact of Domain  Knowledge and Main Issues}
\label{sec:issues}

Our approach is built around the idea of exploiting the available domain knowledge $\mathcal{K}$ on the target classification problem, both in the cases of rejection and multi-label attack. Several existing works use additional knowledge on the learning problem with different goals, being it represented by logic \cite{neurosym,diligenti2017semantic,gnecco2015foundations,DBLP:journals/tnn/GoriM13}, inherited by knowledge graphs or other external resources \cite{melaccienhancing,yuimproving}, and encoded in multiple ways to face specific tasks \cite{scr,sconr,melaccienhancing}. For instance, Semantic-based Regularization \cite{diligenti2017semantic} and the theory formalized in \cite{gnecco2015foundations} focus on the same approach we use here to convert generic FOL knowledge. On one hand, $\mathcal{K}$ might not always be available, thus limiting the applicability of what we propose and of the other aforementioned approaches. On the other hand, $\mathcal{K}$ is about relationships among classes that, in the case of the universal quantifier, hold $\forall x$. As a result, such knowledge is more generic than specific example-level supervisions. Human experts can produce FOL rules to a lesser effort than what is needed to manually label large batches of examples, since $\mathcal{K}$ naturally represents the type of high-level knowledge on the target domain that a human would develop during a concrete experience on the considered task (e.g., \emph{if A happens, then also B or C are triggered, but not D}). Moreover, we are currently working on methods to extract the type of knowledge that we consider in this paper by means of special neural architectures, with clear connections to Explainable AI \cite{ijcai,barbieroentropybased}.

When the number $\ell$ of FOL formulas in $\mathcal{K}$ is large, a larger number of penalty terms $\hat{\phi}_h$ will be considered in  $\varphi$ of Eq.~(\ref{eq:closs}). Of course, every approach that exploits additional knowledge usually incurs in increased complexity when the knowledge base is large \cite{neurosym,diligenti2017semantic,gnecco2015foundations,DBLP:journals/tnn/GoriM13}. In our case, the T-Norm-based conversion does not represent an issue, since it is computed only once in a pre-processing stage, and, similarly, the output of the network $f(x,W)$ is computed only once in order to evaluate $\varphi$ for a certain sample $x$ and for given weights $W$, independently on the size of $\mathcal{K}$. 
However, the computation of $\varphi$ must be repeated at each iteration of the optimization of Eq.~(\ref{train}) or  Eq.~(\ref{attacklogits}), and  when evaluating whether an input should be rejected or not, Eq.~(\ref{eq:reject}). From the practical point of view, the computational complexity scales almost linearly with $\ell$, but each $\hat{\phi}_h$ has a different structure depending on the FOL formula from which it was generated---roughly speaking, formulas involving more predicates usually yield more complex T-Norm-based polynomials. Several heuristic solutions are indeed possible to overcome these issues. For example, the knowledge base could be sub-selected in order to bound the number of rules in which each class is involved, or a stochastic optimization could be devised to sample the rules included in $\varphi$ at each iteration of the optimization process. However, we remark that, in the experimental activities of this paper, none of the mentioned issues arose.

The way we convert FOL rules into polynomial constraints, described in Section~\ref{sec:domain}, inherits the flexibility of logic in terms of knowledge representation capabilities. Of course, the concrete impact of $\mathcal{K}$ in the rejection mechanisms or in MKA depends on the specific information that is encoded by the FOL rules. For instance, suppose that $f_i(x)=1$ for a certain $x$. The formula $f_i(x) \Rightarrow f_v(x) \lor f_z(x) \lor \ldots \lor f_u(x)$ is ``more likely'' to be fulfilled than the formula with an analogous structure in which $\lor$'s are replaced by $\land$'s. In the former, it is enough for a predicate in the conclusions to be $1$, while in the latter, all the predicates of the conclusions must be jointly true. The rejection criterion or MKA are likely to be more effective in the latter case, but it cannot be strongly stated in advance, since it depends on the concrete way in which $f(\cdot,W)$ is developed by the learning procedure, as discussed in Section~\ref{sec:adv}, and, in the case of MKA, on the difficulty in optimizing Eq.~(\ref{attacklogits}).

When restricting our attention to the rejection function of Eq.~(\ref{eq:reject}), a key element to the success of the proposed criterion is the choice of $\tau$. In Section~\ref{sec:adv} we suggested using data in $\mathcal{V}$ to tune $\tau$, that is a valuable solution, but, of course, it strongly depends on the quality of $\mathcal{V}$, similarly to what happens when tuning other hyper-parameters. More generally, a too small $\tau$ will result in a reject-prone system that does not reject only those inputs that are strongly coherent with the domain knowledge. A too large $\tau$ would end up in not rejecting inputs, being them coherent with $\mathcal{K}$ or not. If further information on the formulas in $\mathcal{K}$ is available, such as their expected importance with respect to the considered task, one could avoid computing an averaged measure as $\varphi$, and evaluate the penalty term $\hat{\phi}_h$ of each single formula against its own reject threshold (i.e., multiple $\tau$'s), that might be selected accordingly to the importance of the formula itself (i.e., smaller $\tau$'s in more important formulas).
}

\section{Experiments}
\label{sec:exp}

\bb{
In this section, we report our experimental analysis, discussing the experimental setup in Section~\ref{sect:exp-settings}, and the results of standard and adversarial evaluations for multi-label classifiers in Section~\ref{sect:exp-multi}. We then show in Section~\ref{sect:exp-single} how our multi-label classifiers can also be adopted to mitigate the impact of adversarial examples in single-label classification tasks, when auxiliary classes are exploited. This allows us to highlight that our approach exhibits competitive performances with respect to other baseline defense methods designed under the same assumptions (i.e., without assuming any specific knowledge of the attacks) and against state-of-the-art attacks that are developed for single-label classification tasks.
}

\subsection{Experimental Settings}
\label{sect:exp-settings}

\myparagraph{Datasets.} We considered three image classification datasets, referred to as ANIMALS, CIFAR-100 and PASCAL-Part respectively. 
The first one is a collection of images of animals, taken from the ImageNet database,\footnote{ANIMALS {\scriptsize\url{http://www.image-net.org/}}, CIFAR-100 {\scriptsize\url{https://www.cs.toronto.edu/~kriz/cifar.html}}} the second one is a popular benchmark composed of RGB images ($32 \times 32$)  belonging to different types of classes (vehicles, flowers, people, etc.),$^{5}$ while the last dataset is composed of images in which both objects (Man, Dog, Car, Train, etc.) and object-parts (Head, Paw, Beak, etc.) are labeled.\footnote{PASCAL-Part: \scriptsize{\url{https://www.cs.stanford.edu/~roozbeh/pascal-parts/pascal-parts.html}}}
\begin{table}
\centering
\caption{Datasets and details on the experimental setting. ``\sm{Classes}'' reports the total number of \sm{categories}, \sm{specifying} the number of main classes in parentheses. \bb{The fraction of labeled ($\%L$) samples, \sm{the level of partial labeling} ($\%P$), along with the number of training ($|\mathcal{L}|$), validation ($|\mathcal{V}|$), and test ($|\mathcal{T}|$) examples are also reported.}}
\label{tab:data}
\begin{tabular}{l|ccc|p{0.65cm}p{0.65cm}p{0.65cm}}
    \toprule
    \textit{Dataset} & \textit{\sm{Classes}} &  \textit{\%L} & \textit{\%P} & $|\mathcal{L}|$ & $|\mathcal{V}|$ &  $|\mathcal{T}|$   \\
    \midrule
    ANIMALS     & $33$ ($7$)& $30\%$& $90\%$ & $5808$    & $1244$    & $1243$    \\
    CIFAR-100   & $120$ ($100$)& $30\%$& $0\%$ & $40000$   & $10000$   & $10000$   \\
    PASCAL-Part& $64$ ($20$)& $30\%$& $70\%$& $7072$    & $1515$    & $1515$    \\
    \bottomrule
\end{tabular}
\end{table}
\begin{table}[t]
\centering
\caption{Values of the hyperparameter $\lambda$ selected via cross-validation in our experiments. Note that baseline models TL and FT do not exploit domain knowledge ($\lambda=0$).}
\label{lambdarox}
\begin{tabular}{l|lll}
\toprule
Model & ANIMALS & CIFAR-100 & PASCAL-Part \\
\midrule
TL+C & $10^{-2}$ & $3$ & $10^{-1}$ \\ 
TL+CC & $1$ & $10$ & $1$ \\ 
\midrule
FT+C & $10^{-2}$ & $3$ & $10^{-1}$ \\
\bottomrule
\end{tabular}
\end{table}
\begin{table}[t]
\begin{center}
\caption{Values of the constraint loss $\varphi$ on the test data $\mathcal{T}$.} 
\label{varphirox}
\resizebox{0.45\textwidth}{!}{
\begin{tabular}{l|lll}
\toprule
Model & ANIMALS & CIFAR-100 & PASCAL-Part \\
\midrule
TL & $0.5833$ {\tiny$\pm 0.0316$} & $1.4440$ {\tiny$\pm  0.0087$} & $2.7286$ {\tiny$\pm 0.0853$} \\
TL+C & $0.2134$ {\tiny$\pm 0.0160$} & $1.0406$ {\tiny$\pm 0.0020$} & $1.7422$ {\tiny$\pm 0.0516$} \\ 
TL+CC & $0.2004$ {\tiny$\pm 0.0097$} & $0.7267$ {\tiny$\pm 0.0020$} & $0.7387$ {\tiny$\pm 0.0151$} \\ 
\midrule
FT & $0.3751$ {\tiny$\pm 0.0169$} & $0.9603$ {\tiny$\pm 0.0041$} & $2.4478$ {\tiny$\pm 0.0723$} \\
FT+C & $0.0897$ {\tiny$\pm 0.0113$} & $0.4449$ {\tiny$\pm 0.0068$} & $0.8434$ {\tiny$\pm 0.0471$} \\
\bottomrule
\end{tabular}}
\end{center}
\end{table}
\begin{table}
\centering
\caption{Multi-label classification results in $\mathcal{T}$, for different models, averaged across different repetitions (standard deviations are $<1\%$). The second row-block is restricted to the main classes (Accuracy or F1). See the main text for details.}
\label{tab:all}
\begin{tabular}{p{1.65cm}l|@{\hskip3pt}c@{\hskip3pt}c@{\hskip3pt}c@{\hskip3pt}|@{\hskip3pt}c@{\hskip3pt}c@{\hskip3pt}}
    \toprule
    Metric & Dataset & \textsc{TL} & \textsc{TL+C} & \textsc{TL+CC} & \textsc{FT} & \textsc{FT+C}\\    
    \midrule
    \multirow{3}{*}{F1 (\%)} & ANIMALS & $98.3 $ & ${98.6}$ &  $98.1$  & $98.6$  & ${99.2}$  \\
    & CIFAR-100 & $52.0$  & ${55.1}$  & $53.1$ & $59.3$ & ${64.0}$  \\
    & PASCAL-Part & $69.5$  & ${70.0}$ & $69.4$  & ${69.1}$  & ${71.0}$ \\
    \midrule
    \multirow{3}{*}{\vtop{\hbox{\strut AccMain (\%)$^{1}$}\hbox{\strut F1Main (\%)$^{2}$}}} & ANIMALS$^{1}$ &  $98.8$  &  ${99.2}$ &  ${99.2}$  &  $98.5$  &  ${99.1}$  \\
    & CIFAR-100$^{1}$ &  $53.3$  & $  {55.6}$ &  $52.8$  &  $60.5$  &  ${61.6}$  \\
    & PASCAL-Part$^{2}$ & $73.8$  & ${75.9}$  &  ${69.5}$  & ${70.4}$  & ${75.0}$  \\
    \bottomrule
\end{tabular}
\end{table}
All datasets are used in a multi-label classification setting, \sm{so that the ground truth of each example is composed by a set of binary class labels.}
In the case of ANIMALS there are $33$ \sm{categories}, where the first $7$ ones, also referred to as ``main'' 
 classes, are about specific \sm{categories of animals} 
 (albatross, cheetah, tiger, giraffe, zebra, ostrich, penguin) while the other $25$ \sm{classes} are about \sm{more generic} features (mammal, bird, carnivore, fly, etc.). The CIFAR-100 dataset is composed of $120$ 
 \sm{classes,} out of which $100$ are fine-grained (``main'' \sm{classes}) and $20$ are superclasses. In the PASCAL-Part dataset, after having processed data as in~\cite{donadello2017logic}, we are left with $64$ \sm{categories}, out of which $20$ are objects (``main'' \sm{classes}) and the remaining $44$ are object-parts. 
We have the use of domain knowledge that holds for all the available examples. In the case of ANIMALS, it is a collection of FOL formulas that were defined in the benchmark of P.H. Winston~\cite{winston1986lisp}, and they involve relationships between animal classes and animal properties, such as $\forall x$ FLY$(x)$ $\land$ LAYEGGS$(x)$ $\Rightarrow$ BIRD$(x)$.
In CIFAR-100, FOL formulas are about the father-son relationships between classes, 
while in PASCAL-Part they either list all the parts belonging to a certain object, i.e., \MakeUppercase{Motorbike$\MakeLowercase{(x)} \Rightarrow$ Wheel$\MakeLowercase{(x)} \lor$  Headlight$\MakeLowercase{(x)} \lor$  Handlebar$\MakeLowercase{(x)} \lor$  Saddle$\MakeLowercase{(x)} $ }, or they list all the objects in which a part can be found, i.e., \MakeUppercase{Handlebar$\MakeLowercase{(x)} \Rightarrow$ Bicycle$\MakeLowercase{(x)} \lor$  Motorbike$\MakeLowercase{(x)} $}. 
we also introduced a disjunction or a mutual-exclusivity constraint among the main classes, and another disjunction among the other  \sm{classes}. See Table~\ref{tab:data} and the supplementary material for more details.
Each dataset was divided into training and test sets (the latter indicated with $\mathcal{T}$). The training set was divided into a learning set ($\mathcal{L}$), used to train the classifiers, and a validation set ($\mathcal{V}$), used to tune the model parameters.
We defined a semi-supervised learning scenario in which only a portion of the training set is labeled, sometimes partially (i.e., only a fraction of the \sm{binary labels} of an example is known), as detailed in Table~\ref{tab:data}. We indicated with $\%L$ the percentage of labeled training data, and with $\%P$ the percentage of \sm{binary class labels} that are unknown for each labeled example.\footnote{When splitting the training data into $\mathcal{L}$ and $\mathcal{V}$, we kept the same percentages  of unknown \sm{binary class labels per example ($\%P$)} in both the splits. Of course, in $\mathcal{V}$ \sm{there are no fully-unlabeled examples ($\%L$ is 100)}. Moreover, when generating partial labels, we ensured that the percentages of discarded positive \sm{(i.e., 1)} and negative \sm{(i.e., 0)} \sm{class labels} were the same.}

\myparagraph{Classifiers.} We compared two neural architectures, based on the popular backbone ResNet50, trained using ImageNet data. In the first network, referred to as \textsc{TL}, we transferred the ResNet50 model and trained the last layer from scratch in order to predict the dataset-specific \sm{multiple classes} (sigmoid activation). The second network, indicated with \textsc{FT}, has the same structure of \textsc{TL}, but we also fine-tuned the last convolutional layer. Each model is based on the product T-Norm, and it was trained for a number of epochs $e$ that we selected as follows: $1000$ epochs in ANIMALS, $300$ (\textsc{TL}) or $100$ (\textsc{FT}) epochs in CIFAR-100, and $500$ (\textsc{TL}) or $250$ (\textsc{FT}) in PASCAL-Part, using minibatches of size $64$. We used the Adam optimizer, with an initial step size of $10^{-5}$, except for \textsc{FT} in CIFAR-100, for which we used $10^{-4}$ to speedup convergence. 
We selected the model at the epoch that led to the largest F1 in $\mathcal{V}$. \sm{We considered unconstrained ($\lambda=0$) and knowledge-constrained ($\lambda > 0$) models. The latter are indicated with the \textsc{+C} (and \textsc{+CC}) suffix. }

\myparagraph{Evaluation Metrics.} To evaluate performance, we considered the (macro) \textit{F1} score and a metric restricted to the main classes.\footnote{We compared the outputs against $0.5$ to obtain binary labels.}
For ANIMALS and CIFAR-100, the main classes are mutually exclusive, so we measured the accuracy in predicting the winning main class (\textit{AccMain}), while in PASCAL-Part we kept the F1 score (\textit{F1Main}) as multiple main classes can be predicted on the same input.

\myparagraph{Hyperparameter Tuning.} In Table~\ref{lambdarox} we report the optimal value of $\lambda \in \{10^{-2},10^{-1},1,3,5,8,10,10^{2} \} $ \sm{for the TL+C and FT+C models} used in our experiments, selected via a 3-fold cross-validation procedure. 
\sm{In the case of \textsc{TL}, we 
also considered a strongly-constrained (\textsc{+CC}) model with inferior performance but higher coherence (greater $\lambda$) among the predicted \sm{categories} (that might lead to a worse fitting of the supervisions).\footnote{\textsc{FT+C} has more learnable weights: constraint loss is already small.}}
 Table~\ref{varphirox} reports the value of the constraint loss $\varphi$ measured on the test set $\mathcal{T}$. We used $\mu^{\mathcal{L}} = \mu^{\mathcal{T}}$, setting each component $\mu^{\cdot}_h$ to $1$, with the exception of the weight of the mutual exclusivity constraint or the disjunction of the main classes, which was set to $10$ to enforce the classifier to take decisions. 

\subsection{Experimental Results on Multi-label Classifiers}
\label{sect:exp-multi}
We discuss here the main experiments related to the evaluation of the considered multi-label classifiers.

\myparagraph{Standard Evaluation.} \sm{In order to assess the behaviors of the classifiers in the considered datasets and the available domain knowledge, we compared classifiers that exploit domain knowledge with the ones that do not exploit it.}
The results of our evaluation are reported in Table~\ref{tab:all}, averaged over the 3 training-test splits. \sm{For each of them, 3 runs were considered, using different initialization of the weights}. 
The introduction of domain knowledge allows the constrained classifiers to slightly outperform the unconstrained ones.

\begin{figure*}[t]
    \centering
    {\scriptsize \hskip 8mm \textsc{ANIMALS} \hskip 4.5cm \textsc{CIFAR-100} \hskip 4.5cm \textsc{PASCAL-Part}} \\ \vskip 0.1cm  
    \includegraphics[width=0.3\textwidth, trim={90 267 115 280}, clip]{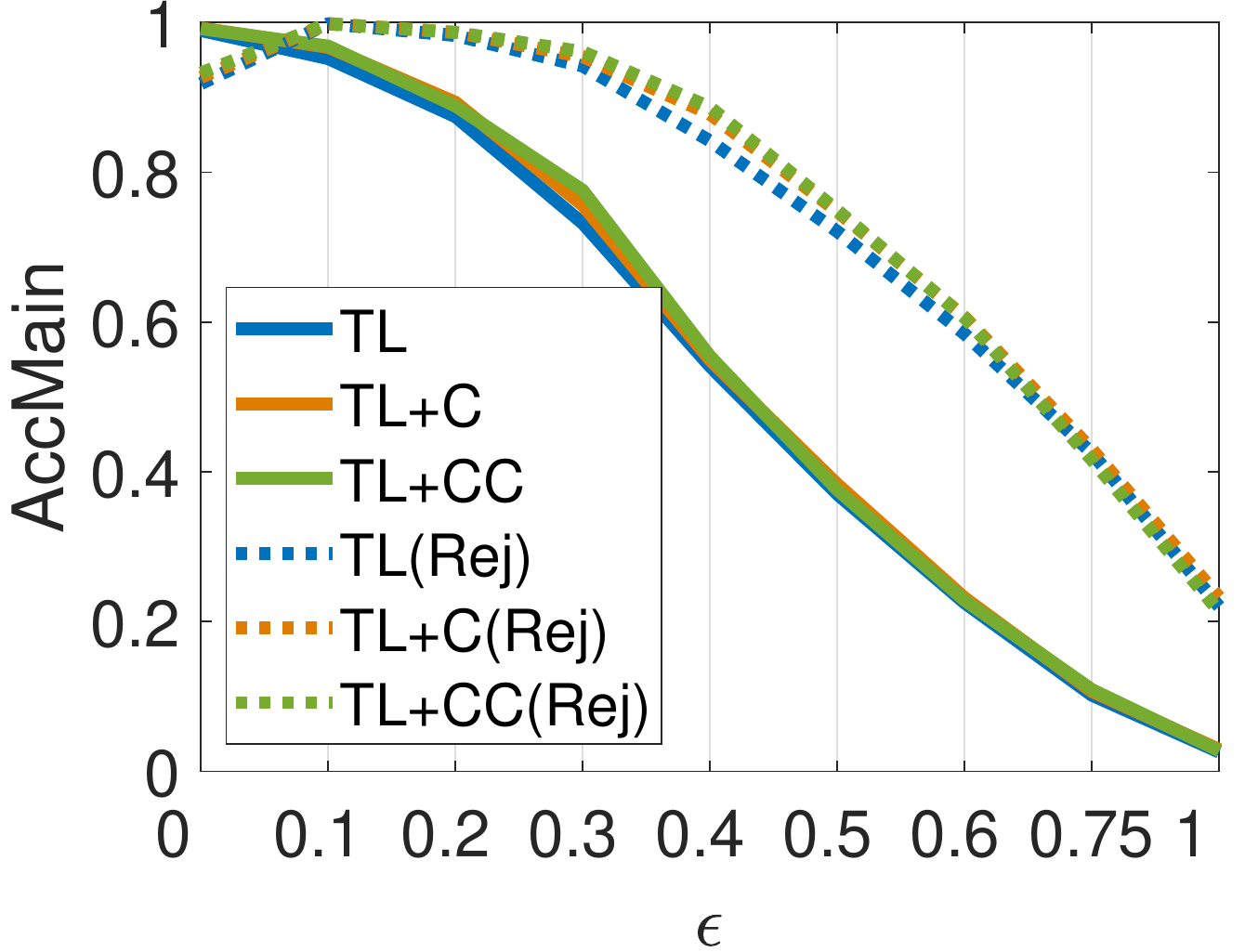}
    \includegraphics[width=0.3\textwidth, trim={90 267 115 280}, clip]{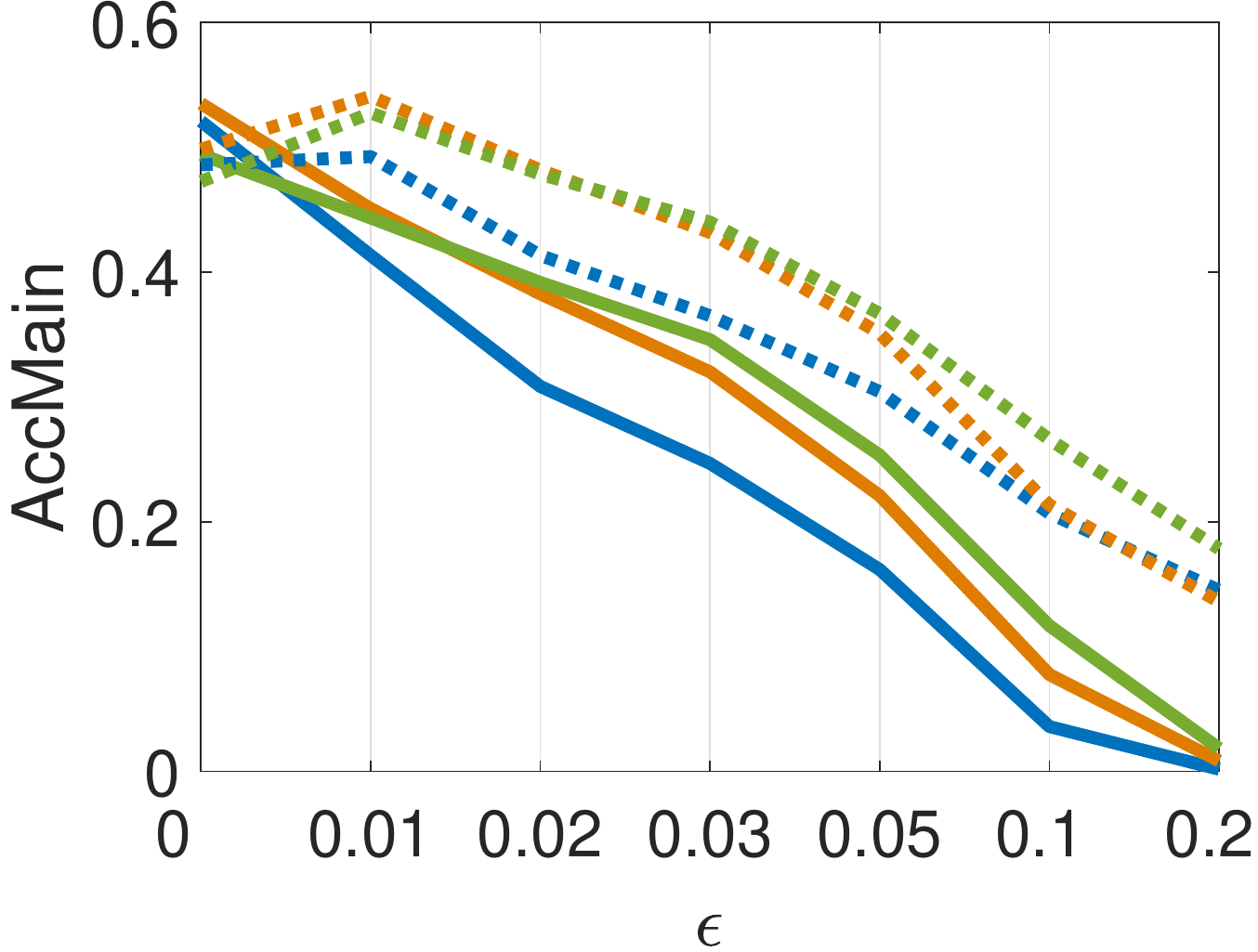}
    \includegraphics[width=0.3\textwidth, trim={90 267 115 280}, clip]{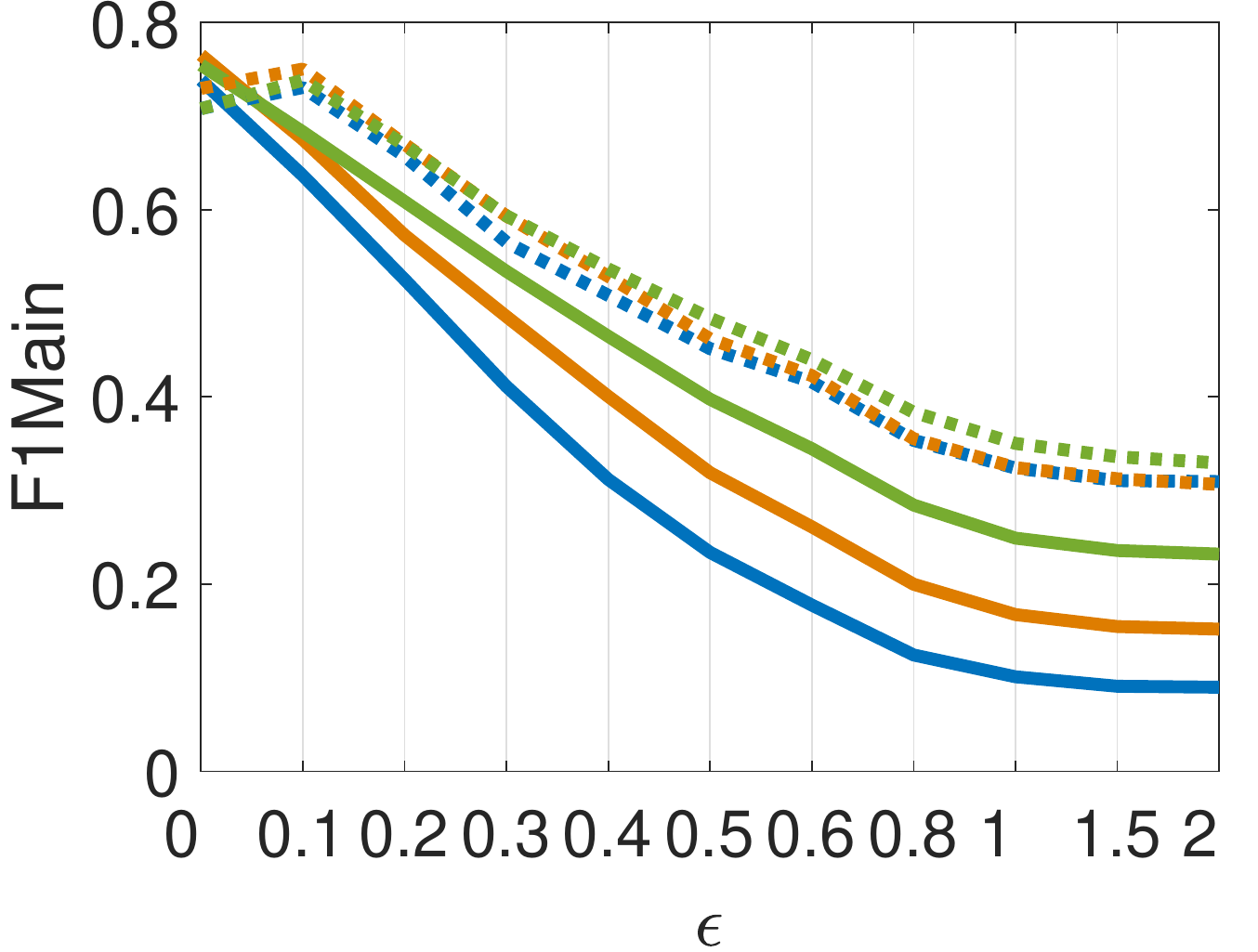}\\
    \includegraphics[width=0.3\textwidth, trim={90 267 115 280}, clip]{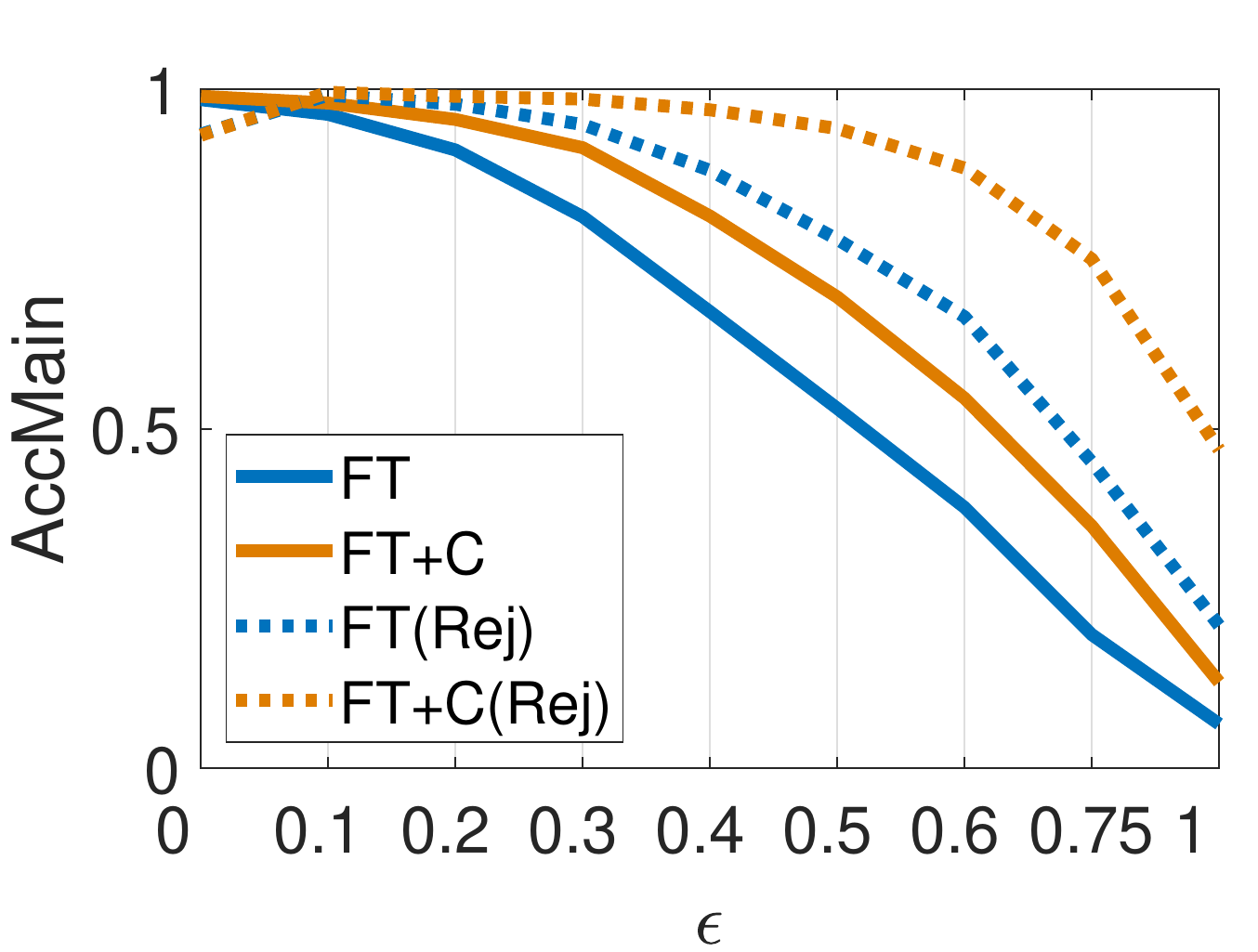}
    \includegraphics[width=0.3\textwidth, trim={90 267 115 280}, clip]{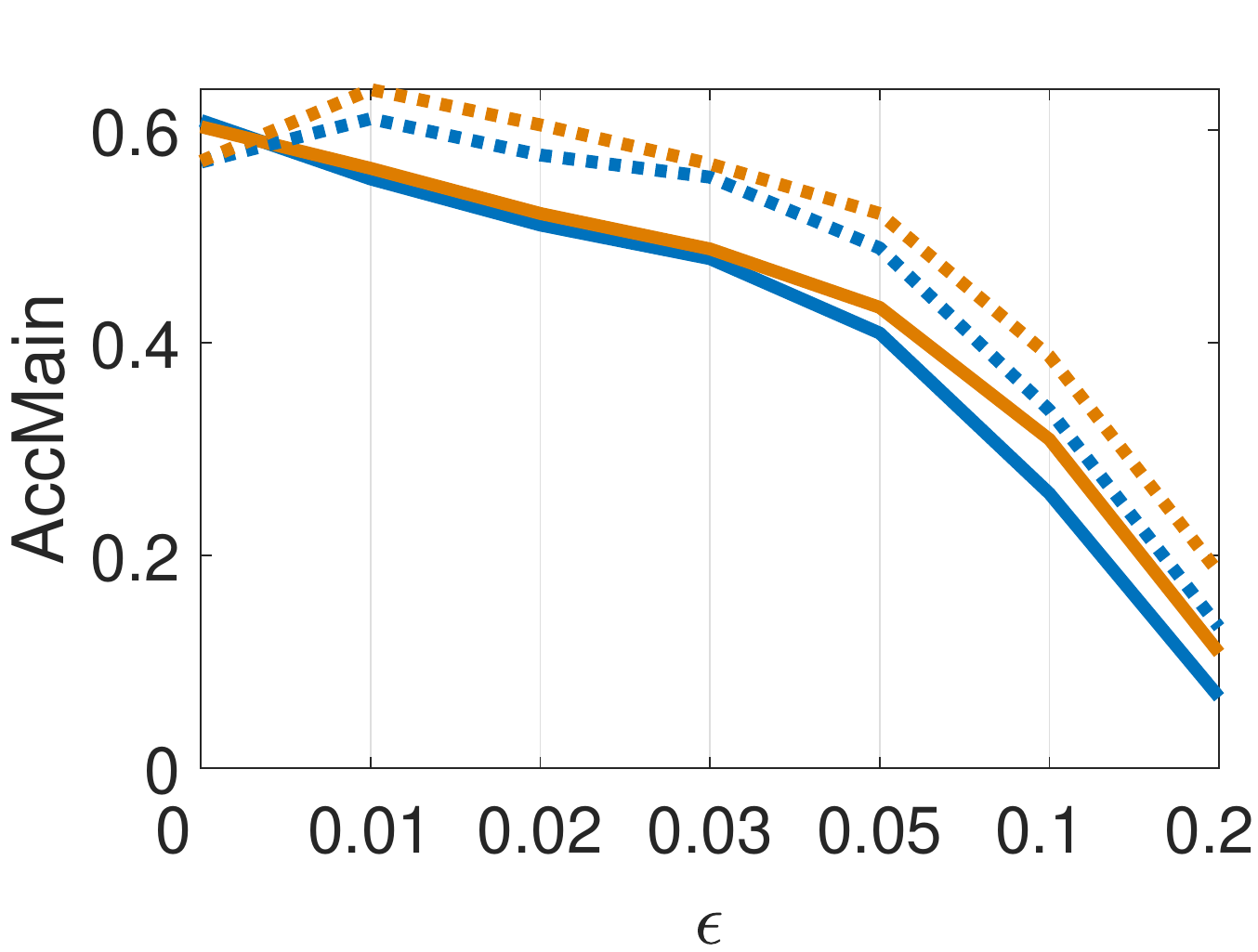}
    \includegraphics[width=0.3\textwidth, trim={90 267 115 280}, clip]{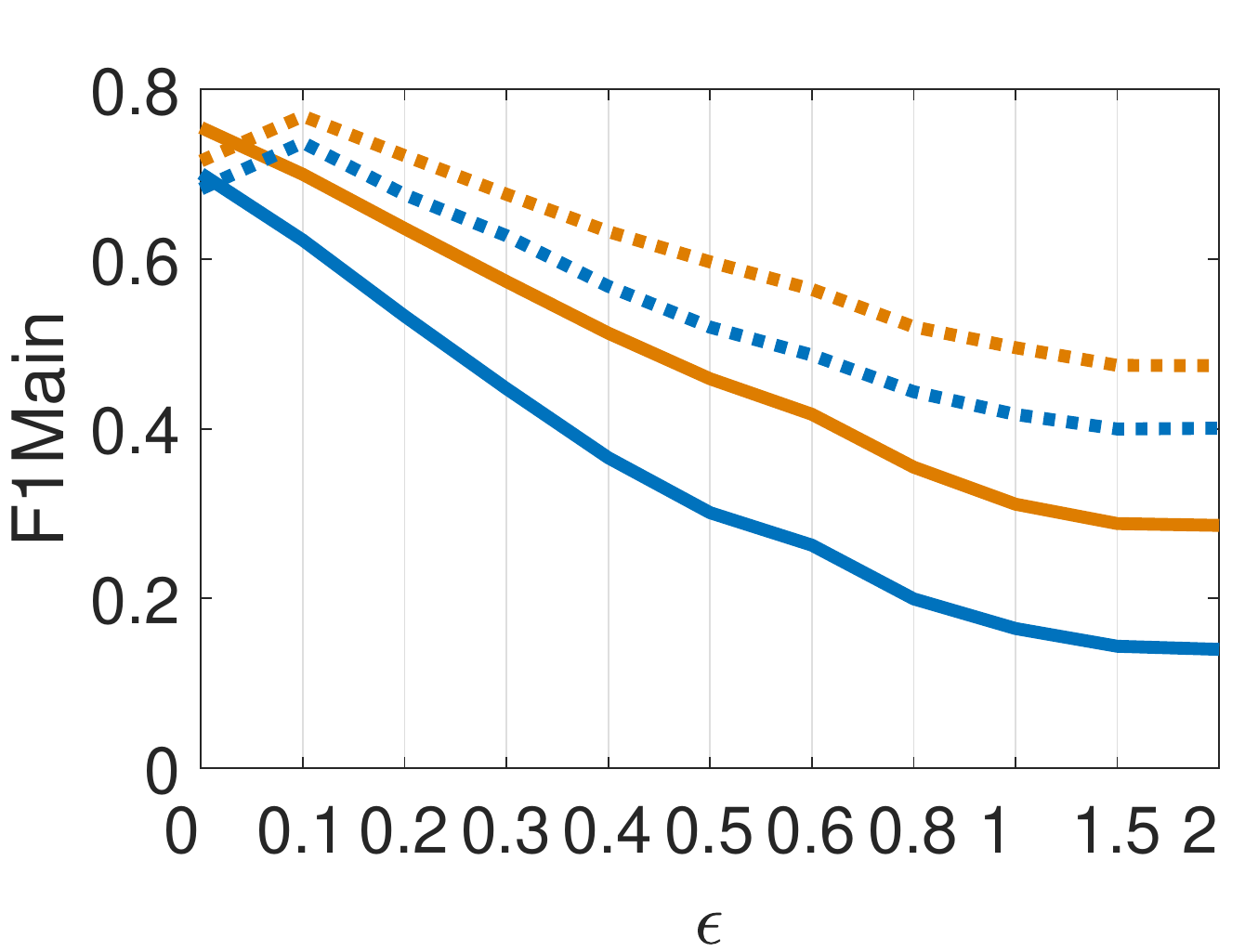}\\      
    \caption{Black-box attacks. Classification quality of vanilla and knowledge-constrained models in function of $\epsilon$. Dotted plots include rejection (Rej) of inputs that are detected to be adversarial.}
    \label{fig:black}
\end{figure*}

\begin{figure*}[t]
    \centering
    {\scriptsize \hskip 8mm \textsc{ANIMALS} \hskip 4.5cm \textsc{\sm{CIFAR-100}}\hskip 4.5cm \textsc{PASCAL-Part}}\\  \vskip -0.00cm   
    \includegraphics[width=0.3\textwidth, trim={90 260 115 305}, clip]{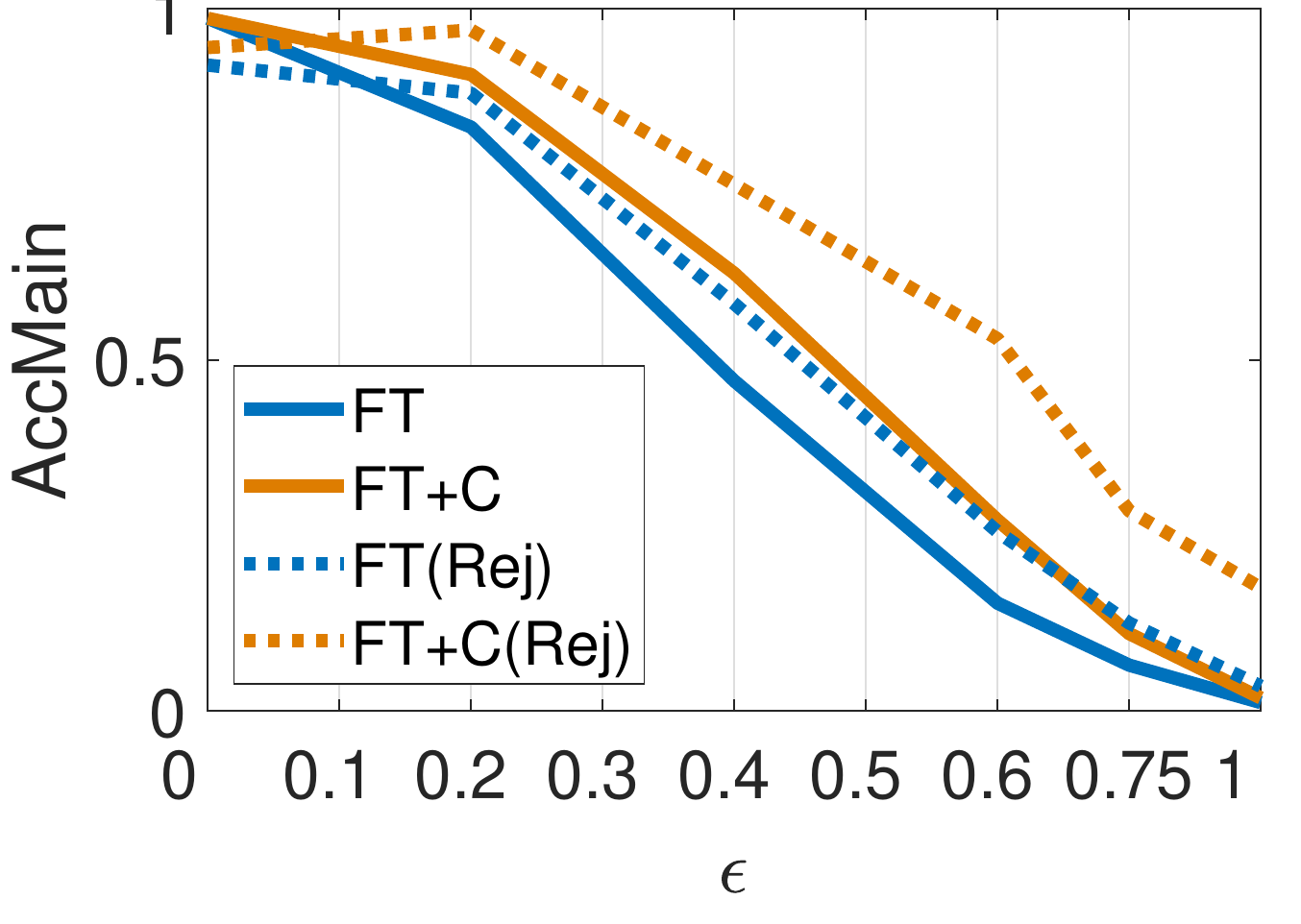}
    \includegraphics[width=0.295\textwidth, trim={90 270 115 290}, clip]{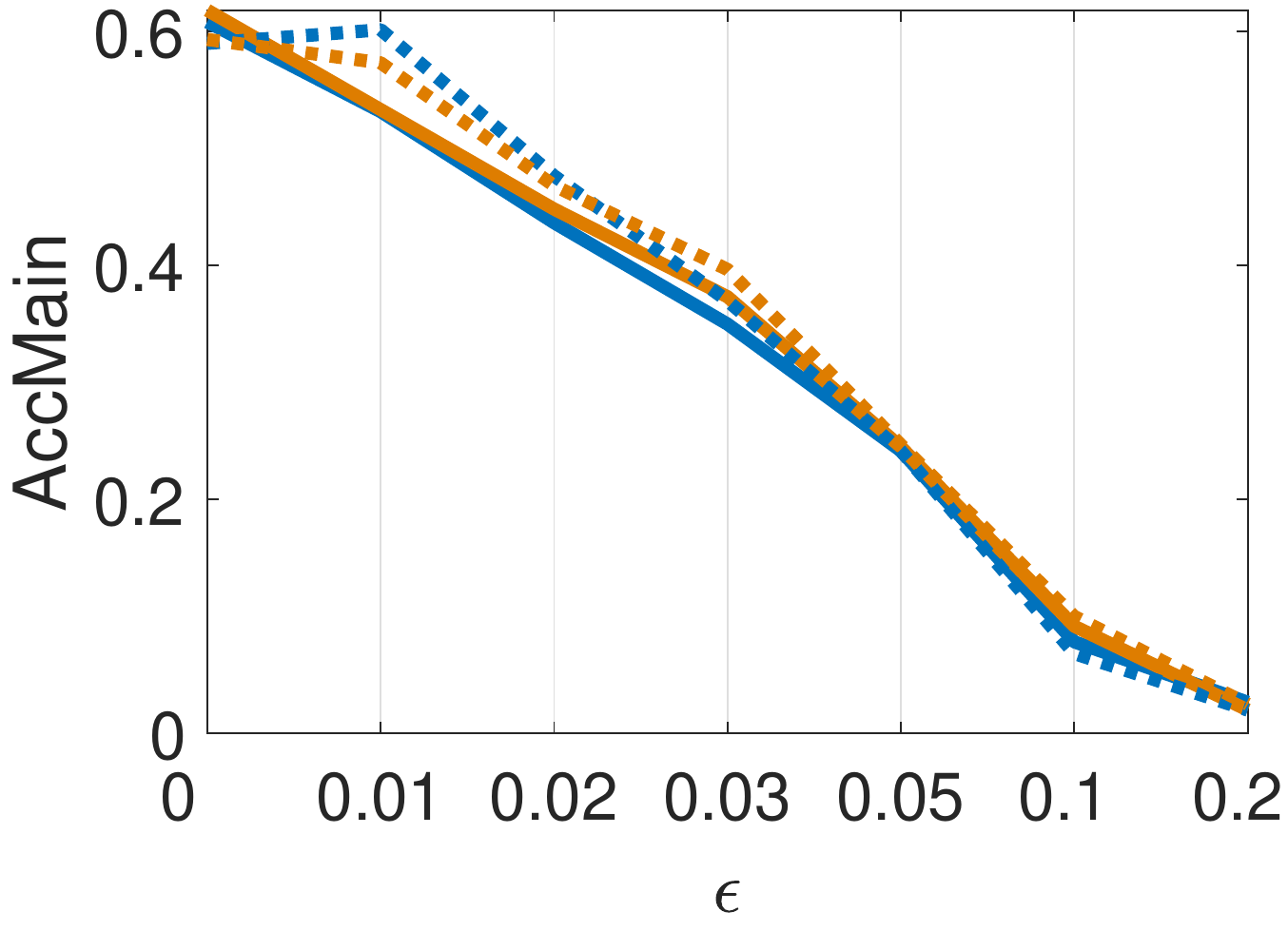}
    \includegraphics[width=0.305\textwidth, trim={90 265 115 295}, clip]{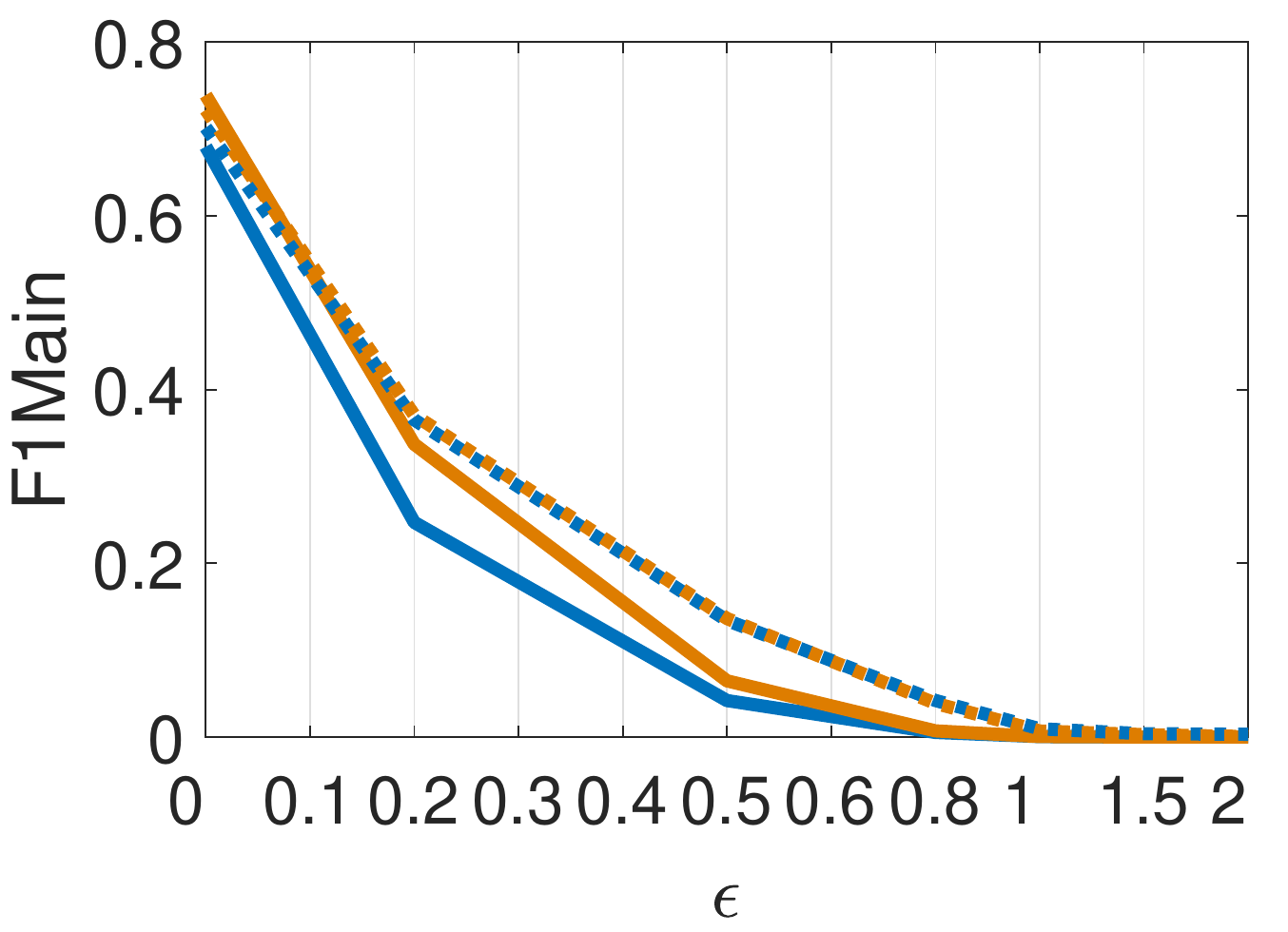}    
    \caption{White-box attacks in the case of the FT classifiers. \sm{Classification quality of vanilla and knowledge-constrained models in function of $\epsilon$. Dotted plots include rejection (Rej) of inputs that are detected to be adversarial.}}
    \label{fig:white}
\end{figure*}

\myparagraph{Adversarial Evaluation.}  To evaluate adversarial robustness, we used the MKA attack procedure described in {Section~\ref{sec:adv}}. 
and we restricted the attack to work \sm{on the already introduced main classes}, being them associated to the most important \sm{categories} of each problem. 
In ANIMALS and CIFAR-100 we assumed the attacker to have access to the information on the mutual exclusivity of the main classes, so that $p$
in Eq.~\eqref{attacklogits} 
is not required to change during the attack optimization. We also set $\kappa=\infty$ to maximize confidence of misclassifications at each given perturbation bound $\epsilon$.  
All the following results are averaged after having attacked twice the model obtained after each of the 3 training runs.

In the \textit{black-box} setting, we assumed the attacker to be also aware of the network architecture of the target classifier, and attacks were generated from a surrogate model trained on a different realization of the training set. 
Fig.~\ref{fig:black} shows the classification quality as a function of the data perturbation bound $\epsilon$, comparing models trained with and without constraints against those implementing the detection/rejection mechanism described in Eq.~\eqref{test}. When using such mechanism, the rejected examples are marked as correctly classified if they are adversarial ($\epsilon>0$), otherwise ($\epsilon=0$) they are marked as points belonging to an unknown class, slightly worsening the performance.
The \textsc{+C}/\textsc{+CC} models show larger accuracy/F1 than the unconstrained ones. Despite the lower results at $\epsilon =0$, models that are more strongly constrained (\textsc{+CC}) resulted to be harder to attack for increasing values of $\epsilon$. When the knowledge-based detector is activated, the improvements with respect to models without rejection are significantly evident. No model is specifically designed to face adversarial attacks and, of course, there are no attempts to reach state-of-the-art results.\footnote{Recall that our rejection mechanism is completely agnostic to the attack; it neither assumes any knowledge of the attack algorithm nor is retrained on adversarial examples. Nevertheless, it can be used as a complementary defense mechanism.} However, the positive impact of exploiting domain knowledge can be observed in all the considered models and datasets, and for almost all the values of $\epsilon$, confirming that such knowledge is not only useful to improve classifier robustness, but also as a mean to detect adversarial examples at no additional training cost. In general, \textsc{FT} models yield better results, due to the larger number of optimized parameters. In ANIMALS the rejection dynamics are providing large improvements in both \textsc{TL} and \textsc{FT}, while the impact of domain knowledge is mostly evident on the robustness of \textsc{FT}. In CIFAR-100, domain knowledge only consists of basic hierarchical relations, with no intersections among child classes or among father classes. By inspecting the classifier, we found that it is pretty frequent for the fooling examples to be predicted with a strongly-activated father class and a (coherent) child class, i.e., we have strongly-paired classes, accordingly to Def.~\ref{pairing}. Differently, the domain knowledge in the other datasets is more structured, yielding better detection quality on average, remarking the importance of the level of detail of such knowledge to counter adversarial examples. In the case of PASCAL-Part, the detection mechanism turned out to better behave in unconstrained classifiers, even if it has a positive impact also on the constrained ones. This is due to the intrinsic difficulty of making predictions on this dataset, especially when considering small object-parts. The false positives have a negative effect in the training stage of the knowledge-constrained classifiers.


To provide a comprehensive, worst-case evaluation of the adversarial robustness of our approach, we also considered a \textit{white-box} adaptive attacker that knows everything about the target model and exploits knowledge of the defense mechanism to bypass it. Of course, this attack always evades detection if the perturbation size $\epsilon$ is sufficiently large. 
We evaluated multiple values of $\alpha$ of Eq.~\eqref{attacklogits}, selecting the one that yielded the lowest values of such objective function. In Fig.~\ref{fig:white} we report the outcome of this analysis for FT models, showing that, even if the accuracy drop is obviously evident for all datasets, in ANIMALS the constrained classifiers require larger perturbations than the unconstrained ones to reduce the performance of the same quantity. \bb{A similar behavior is shown in CIFAR-100, even though only at small $\epsilon$ values. Accordingly, fooling the proposed detection mechanism is not always as trivial as one might expect, even in this worst-case setting.} 
\sm{The impact of the rejection mechanisms is significantly reduced, as expected, but still having a positive impact.}
We refer the reader to the supplementary material for more details about these attacks and their optimization.
Finally, let us point out that the performance drop caused by the white-box attack is much larger than that observed in the black-box case. However, since domain knowledge is not likely to be available to the attacker in many practical settings, it remains an open challenge to develop stronger, practical black-box attacks that are able to infer and exploit such knowledge to bypass our defense mechanism.

\begin{figure*}[!t]
    \centering
    {\scriptsize\textsc{\rev{Black-box}}
    \hskip 0.41\textwidth
    \scriptsize\textsc{\rev{$\ $ White-box}}}\\
    \vskip 1mm
    \rev{{\includegraphics[clip,trim=20 25 20 18,height=0.171\textwidth]{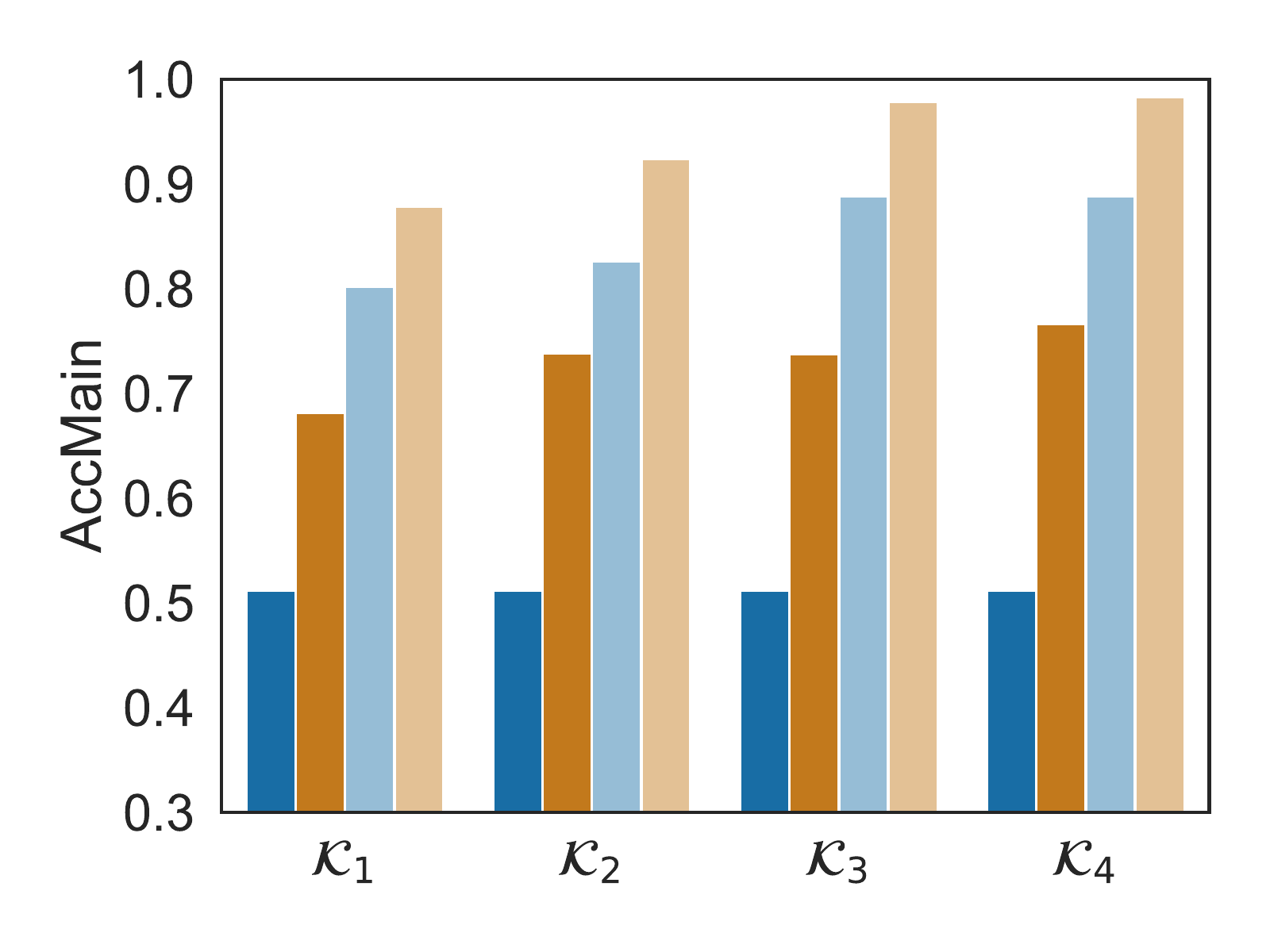}
    \includegraphics[clip,trim=20 22 20 18,height=0.171\textwidth]{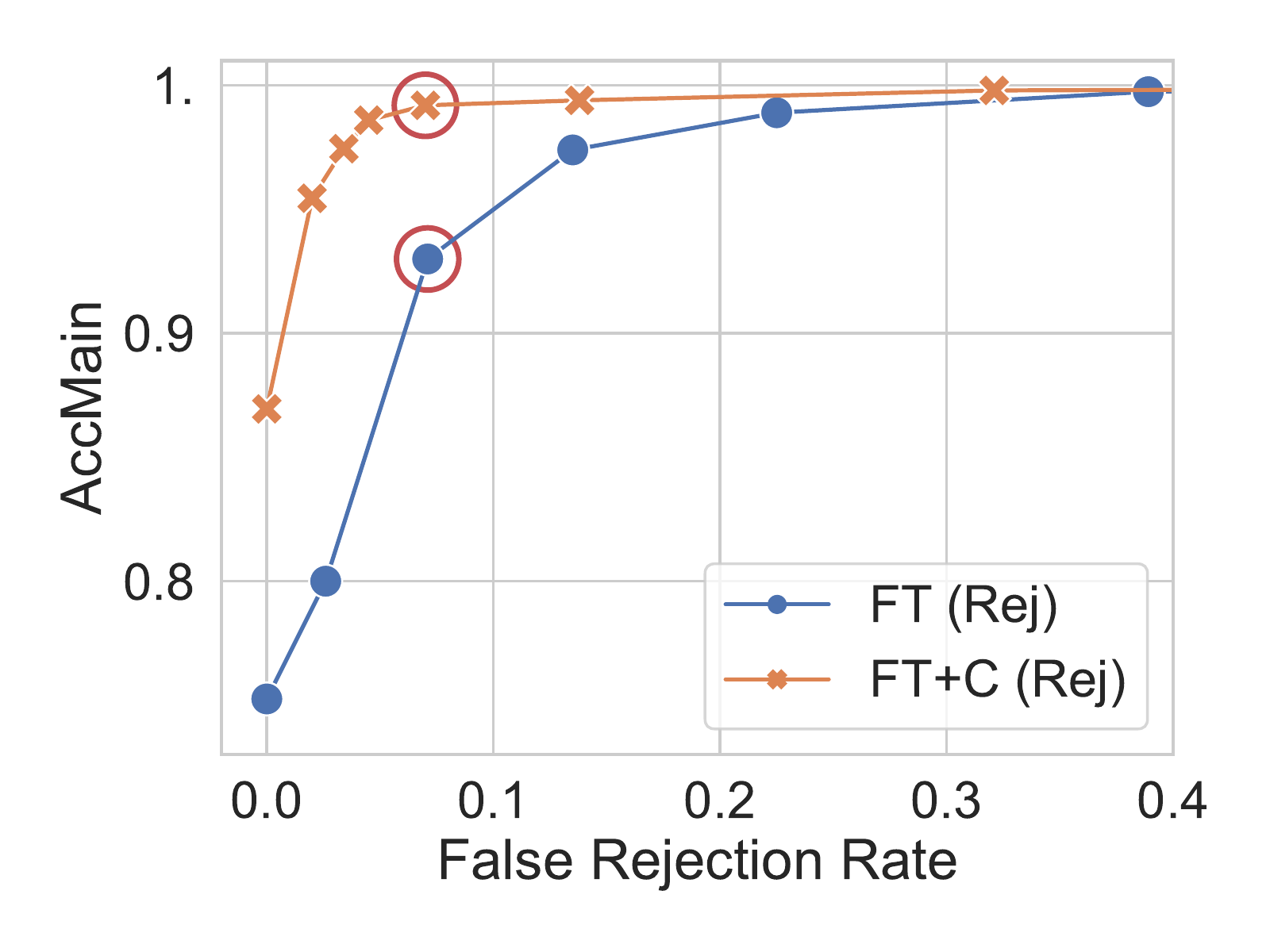}
    \hskip 2.5mm
    \includegraphics[clip,trim=20 25 20 18,height=0.171\textwidth]{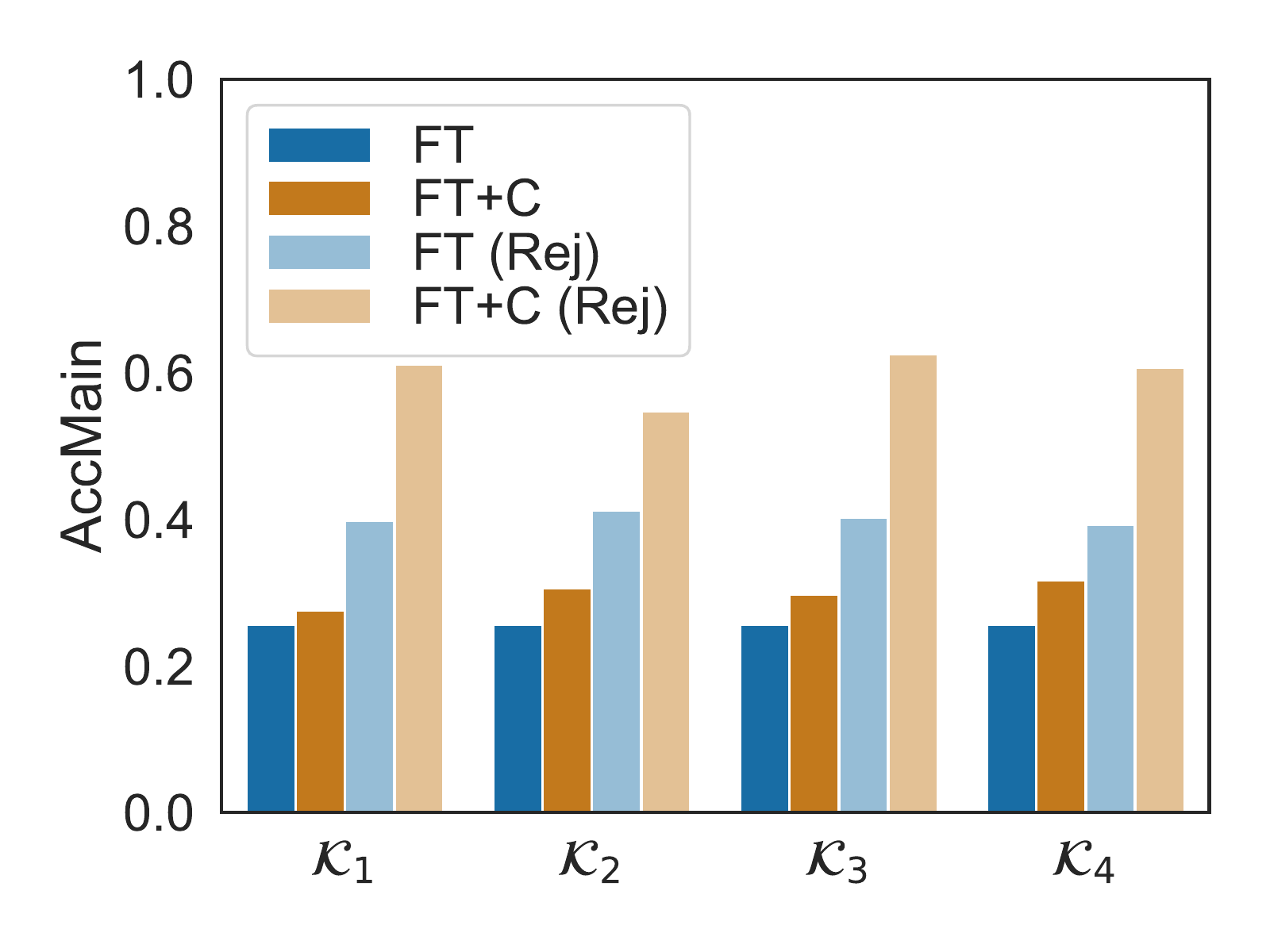}
    \includegraphics[clip,trim=20 22 10 15,height=0.171\textwidth]{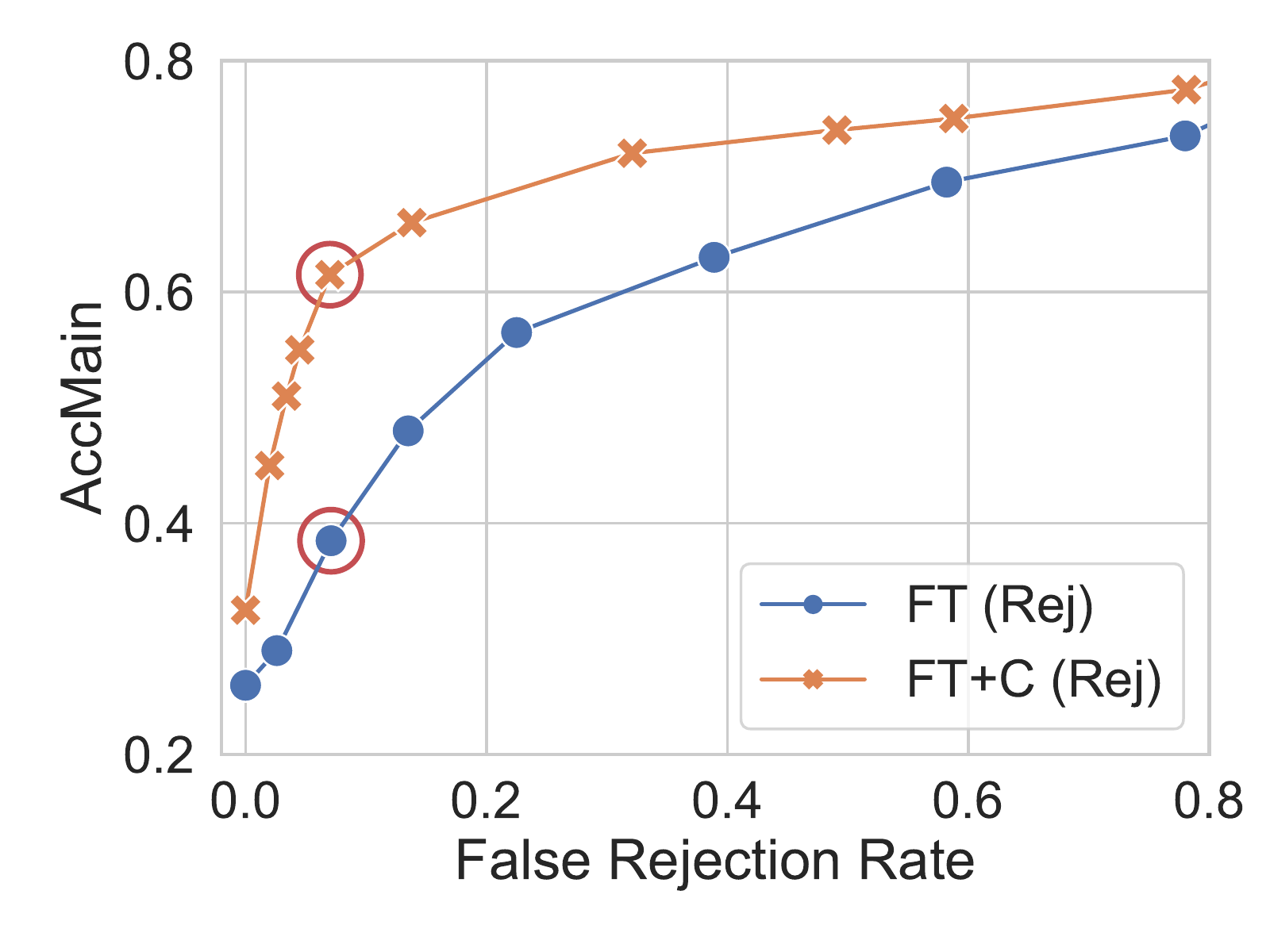}}}\\
    \vskip 1mm
    {\rev{(a) \hskip 0.22\textwidth (b)}}
    \hskip 0.22\textwidth
    {\rev{(c) \hskip 0.22\textwidth (d)}}\\
    \caption{\rev{Further analysis of the proposed approach in the ANIMALS dataset ($\epsilon=0.5$). The first two plots are about the black-box setting, and the last two ones are about the white-box case; (a,c---legend reported only on the latter, for better readability): increasing amounts of domain knowledge $\mathcal{K}_1,\ldots,\mathcal{K}_4$; (b,d): different values of the rejection threshold $\tau$ (from larger to smaller values, left-to-right).}}
    \label{fig:tau}
\end{figure*}

\subsubsection{\minor{In-depth Analysis}}
\rev{\myparagraph{\minor{Incomplete Domain Knowledge.}} We investigated in more detail the relative impact of the domain information on a target problem, simulating the availability of differently sized knowledge bases, $\mathcal{K}_1$, $\mathcal{K}_2$, $\mathcal{K}_3$, $\mathcal{K}_4$, where each $\mathcal{K}_j \subseteq \mathcal{K}$. In particular, we considered the ANIMALS dataset, and we generated $\mathcal{K}_1$, $\mathcal{K}_2$, $\mathcal{K}_3$ by removing some of the FOL formulas of the original $\mathcal{K}$ that was used in the previous experiments (i.e., the one of Table~\ref{animalsrules}, supplementary material), while $\mathcal{K}_4=\mathcal{K}$. This means that some information that belongs to $\mathcal{K}_4$ is actually missing in the other knowledge sets. In detail, we created $\mathcal{K}_1$ by removing the rules that either include the 'mammal' or the 'bird' categories, while $\mathcal{K}_2$ is the outcome of discarding from $\mathcal{K}$ the rules including the 'mammal' category. Similarly, $\mathcal{K}_3$ is obtained removing the rules of the 'bird' category.
We executed a batch of independent experiments, each of them using only one of the generated knowledge bases, and focusing on the same models of Fig.~\ref{fig:black} (bottom) and Fig.~\ref{fig:white}, that were retrained from scratch. 
Fig.~\ref{fig:tau} (a,c) shows the classification quality we obtained in the black (a) and white box (c) settings, using the MKA attack and   $\epsilon=0.5$ (almost the mid of the plots in Figs.~\ref{fig:black}-\ref{fig:white}). In the black-box case, focusing on the models that include rejection (Rej), it is evident how larger knowledge bases yield better results. Interestingly, comparing the outcome of such models with the ones without rejection, we can see that our defense makes the classifier more robust to attacks even when using the smallest amount of knowledge ($\mathcal{K}_1$), confirming the versatility of what we propose. In the white-box setting there are still changes in the accuracy when varying $\mathcal{K}_j$, but they are not so evident and they lack a clear trend. This was expected, since, in this case, the attack procedure is aware of the domain knowledge. However, this result confirms the capability of MKA to craft adversarial examples that lead to knowledge-coherent predictions (to a certain extent) even when varying the level of detail of  the knowledge sets.}

\rev{\myparagraph{\minor{Rejection Threshold.}} In the same experimental setting we also explored the sensitivity of the system to the rejection threshold $\tau$, using the whole knowledge set $\mathcal{K}$. We compared different $\tau$'s that are smaller or greater than the one we selected using validation data (Section~\ref{sec:adv}), indicated here with $\tau^{\star}$. In particular, we evaluated $\tau \in \{10^{\kappa}\tau^{\star},\ \kappa \in [-5,5], \text{integer}\}$, and we measured both the classification quality on the perturbed data, as we did so far, and the rejection rate on the clean test data, where no samples should be rejected. Fig.~\ref{fig:tau} (b,d) reports these two measures on the y-axis and x-axis, respectively, and each marker is about a a specific $\tau$, considering  black (b) and white (d) box settings. The value of $\tau^{\star}$ has been highlighted with a red circle, and only the significant portions of the plots are shown. Too small thresholds (rightmost areas of each plot) lead to systems that frequently reject also clean examples, while too large $\tau$'s (leftmost areas) do not improve the quality of the classifiers, that are fooled by some of the data generated in an adversarial manner within the $\epsilon$-bound. Overall, the $\tau^{\star}$ we selected represents a pretty appropriate trade-off between the two measures.}

\minor{\myparagraph{\minor{Noisy Domain Knowledge.}}
We further extended our analysis by considering the case in which the available domain knowledge is noisy, thus including a small percentage of information that is incorrect in the ANIMALS domain. We simulated three scenarios by altering the original knowledge base $\mathcal{K}$ using three different criteria, yielding the noisy knowledge bases $\tilde{\mathcal{K}}_a$,  $\tilde{\mathcal{K}}_b$, $\tilde{\mathcal{K}}_c$, respectively. The chosen criteria either modify four of the existing rules, making them not coherent with the clean knowledge, or they add four new rules that are not correct in the considered domain, as shown in detail in Table~\ref{tab:ka},  Table~\ref{tab:kb}, and Table~\ref{tab:kc} of the supplementary material, and for which we provide a brief description in the following. In the case of $\tilde{\mathcal{K}}_a$, we selected four existing implications of $\mathcal{K}$, and we altered the premises of two of them and the conclusions of the other two ones. We ensured to inject noise in the main and auxiliary classes in a balanced manner. The same balancing is also kept when generating  $\tilde{\mathcal{K}}_b$, that, however, was obtained by adding four new rules to $\mathcal{K}$. Finally, $\tilde{\mathcal{K}}_c$ is the result of augmenting each main-class-oriented conclusion of four existing implications with a disjunction involving a different, randomly selected, main class. For example, the clean rule  BLACKSTRIPES$(x)\land$UNGULATE$(x)\land\ldots\Rightarrow$ZEBRA$(x)$ is altered by replacing the conclusion ZEBRA$(x)$ with ZEBRA$(x)\lor$TIGER$(x)$. These rules are fulfilled both for configurations that make true their original/clean counterparts, and for other configurations that are actually wrong in the ANIMALS domain. Focusing on the same experimental setup we defined when testing differently sized knowledge bases, we investigated the effects of using each noisy knowledge base both to learn the classifier and/or as a rejection criterion. Fig.~\ref{fig:minor} shows the results of our experience. 
\begin{figure}[!t]
    \centering
    {\scriptsize\textsc{\minor{$\ \quad\quad\quad$Black-box}}
    \hskip 0.15\textwidth
    \scriptsize\textsc{\minor{$\ \quad$ White-box}}}\\
    \vskip 1mm
    \minor{{\includegraphics[clip,trim=20 22 20 18,height=0.171\textwidth]{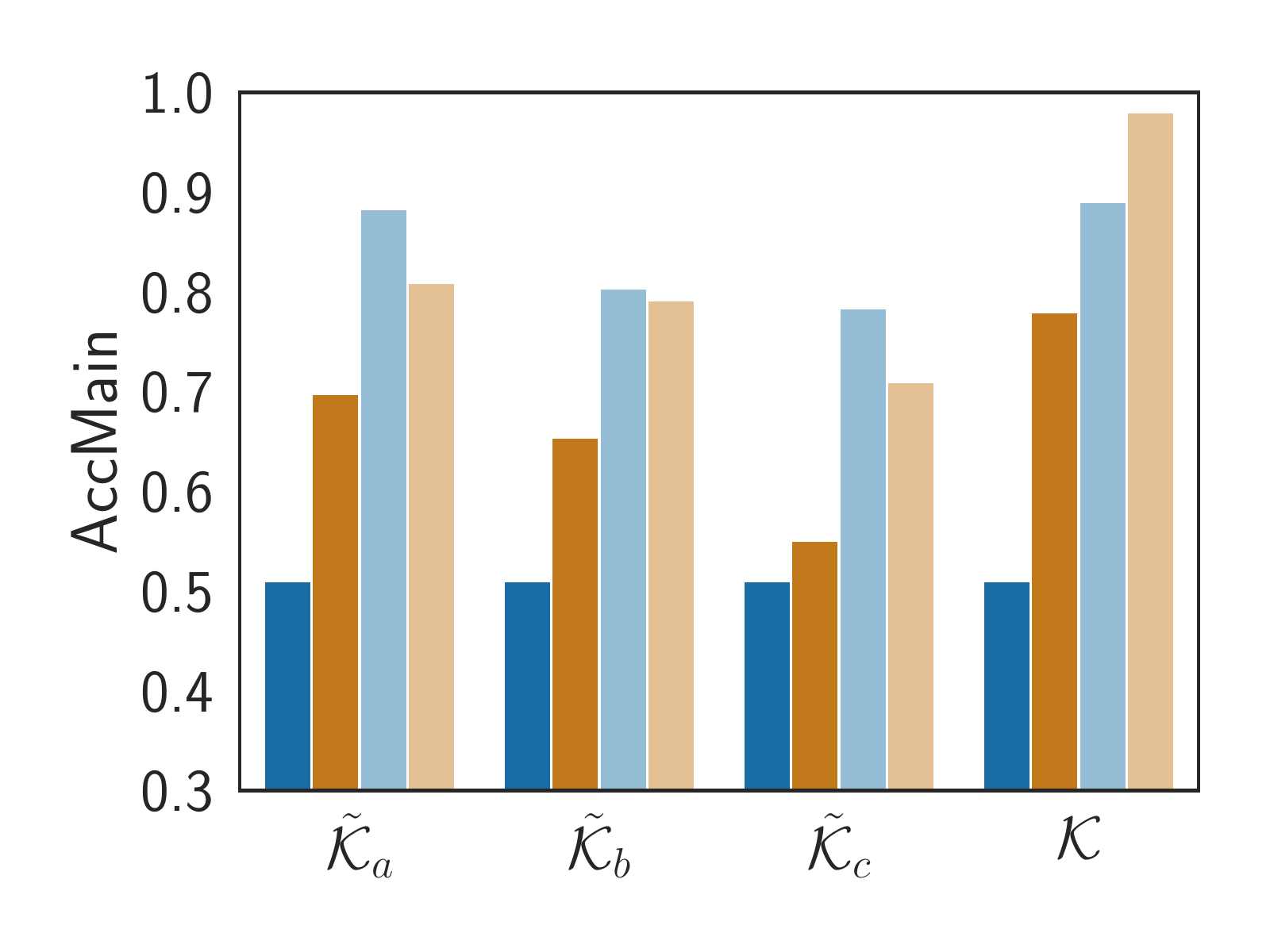}
    \hskip 1mm
    \includegraphics[clip,trim=20 22 20 18,height=0.171\textwidth]{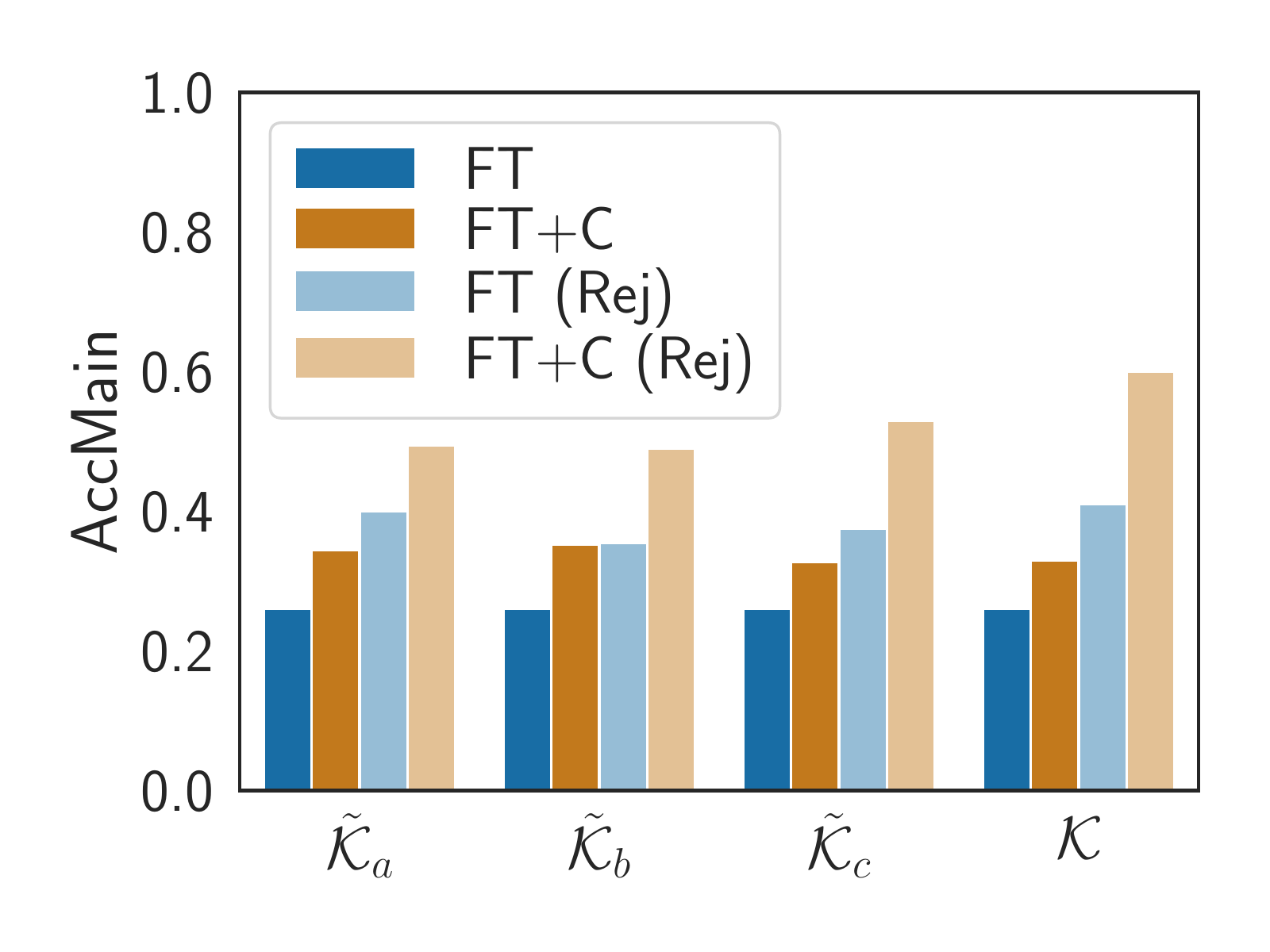}}}
    \caption{\minor{Noisy domain knowledge. Analysis of the proposed approach in the same setup of Fig.~\ref{fig:tau}, when exploiting different noisy knowledge bases $\mathcal{\tilde{K}}_a$, $\mathcal{\tilde{K}}_b$, $\mathcal{\tilde{K}}_c$ (see the paper text for details). $\mathcal{K}$ is the original noise-free knowledge.}}
    \label{fig:minor}
\end{figure}
As expected, learning with a noisy knowledge base reduces the accuracy of the classifier, since the network learns from FOL rules that are partially in contrast with some of the available labeled examples. It is the case of FT+C trained on any noisy  $\mathcal{\tilde{K}}_{\cdot}$, when compared with the case of the clean $\mathcal{K}$ (rightmost set of bars). However, when adding the rejection criterion, we still observe a significant improvement in the performance, even if slightly smaller than in the case of $\mathcal{K}$. Rejection is the outcome of evaluating the average violations of all the available rules, so that the effect of the noisy portion of the knowledge base is partially compensated by the other non-noisy rules. Of course, the final outcome depends on the type of noise we injected into the knowledge base. In the considered experience, adding wrong rules ($\mathcal{\tilde{K}}_b$) led to lower accuracies that when perturbing some of the existing rules ($\mathcal{\tilde{K}}_a$). The case of $\mathcal{\tilde{K}}_c$ is the one that most evidently impacted the classifier performance, with or without rejection. On one hand, we  simply introduced more error-tolerant conclusions; on the other hand, we did it by   altering FOL rules whose conclusions are all about main classes, significantly compromising the way they are related.
Interestingly, for all the noisy knowledge bases, adding the rejection module (Rej) to FT turned out to be better than adding it to FT+C, suggesting that a rejection criterion based on noisy knowledge could be more effective in classifiers that have not been already exposed to such noisy information during the learning stage.
As expected, results collected in the white-box case show that whenever the attacker has full access to the knowledge, being it noisy or not, he can craft MKA attacks with similar outcomes in terms of performance drops. 
Overall, this experience confirms that what we propose can indeed be applied also in case of partially noisy domain knowledge, still increasing the robustness of the classifier, even if to a smaller extent than in case of clean knowledge.
}


\subsection{Experimental Results on Single-label Classifiers}
\label{sect:exp-single}
\bb{\sm{The focus of this paper is on multi-class classification paired with domain knowledge. However, as anticipated in Section~\ref{sec:adv} and qualitatively shown in \rev{Fig.~\ref{fig:single}}, we can consider a special setting in which a single-label classifier internally includes predictors over auxiliary classes that are involved in the knowledge constraints. We experimentally evaluate this setting in the context of the ANIMALS and CIFAR-100 datasets, where the respective main classes (described in Section~\ref{sect:exp-settings}) are mutually exclusive (which is not the case of PASCAL-Part), thus well suited to simulate the setting of \rev{Fig.~\ref{fig:single}}. 
We compared the proposed rejection mechanisms} to a concurrent defense mechanism developed under the same assumptions (i.e., without assuming any knowledge of the attack algorithm), known as Neural Rejection  (NR)~\cite{melis17-vipar,sotgiu20}, and against the state-of-the-art attacks included in the AutoAttack framework~\cite{autoattack}, developed for single-label classification tasks.}

\myparagraph{Compared Defense and Attack Strategies.} 
\bb{The NR defense mechanism, proposed in~\cite{melis17-vipar,sotgiu20}, aims to reject inputs that are far from the training data in a given representation space. The rationale is that points with low support from the training set cannot be reliably classified, and should be thus rejected.
To this end, the output layer of the deep network is replaced with a Support Vector Machine trained using the RBF kernel (SVM-RBF), which enforces the prediction scores to be proportional to the distance between the input sample and the reference prototypes (i.e., the support vectors) in the representation space. Samples are rejected if the prediction scores do not exceed the rejection threshold. Similarly to our approach, this defense mechanism does not make any assumptions on the attack to be detected, other than assuming an anomalous behavior with respect to the observed training data.}

\begin{table*}[!ht]
\centering
\caption{\ad{ANIMALS dataset. Vulnerability analysis of the classifiers against MKA and state-of-art attacks\rev{---classification quality is reported, the same of Figs.~\ref{fig:black}-\ref{fig:white} (first column)}. \sm{For each type of classifier (TL,FT), rows are organized into three groups, that are: models without rejection, with rejection (Rej), classifier equipped with Neural Rejection (NR). For each attack (columns\minor{---see \cite{autoattack} for a description of the compared attacks}), the result of the most robust classifier in the group is highlighted in bold. Models exploiting the proposed rejection (Rej) that overcome NR are marked with *, and vice-versa.}}}
\label{tab:bb_animal}
\begin{tabular}{@{}lc|ccccc|ccccc@{}}
\toprule
        & \multicolumn{1}{l}{} & \multicolumn{5}{c}{White-box attacks ($\epsilon=0.5$)} & \multicolumn{5}{c}{Black-box transfer attacks ($\epsilon=0.5$)} \\ \midrule
  \multicolumn{1}{l}{Model} &
  \multicolumn{1}{c}{$\epsilon=0$} &
  \multicolumn{1}{c}{MKA} &
  \multicolumn{1}{c}{APGD-CE} &
  \multicolumn{1}{c}{APGD-T} &
  \multicolumn{1}{c}{FAB-T} &
  \multicolumn{1}{c}{Square} &
  \multicolumn{1}{c}{MKA} &
  \multicolumn{1}{c}{APGD-CE} &
  \multicolumn{1}{c}{APGD-T} &
  \multicolumn{1}{c}{FAB-T} &
  \multicolumn{1}{c}{Square} \\ \toprule
\multicolumn{1}{l}{TL}       & 99.0  & 25.0    & 17.5      & 14.9     & 20.0    & 96.6     & 45.3 & 29.4    & 29.1   & 83.1  & 98.4   \\
\multicolumn{1}{l}{TL+C}   & 99.3  & \textbf{25.0}   & \textbf{19.0}      & 15.4    & 21.5    & 98.0     & 47.7 & 29.8    & 30.6   & 87.7  & 98.9   \\
\multicolumn{1}{l}{TL+CC}  & 99.3  & 24.5   & 18.1      & \textbf{15.5}     & \textbf{22.7}    & \textbf{98.2}     & \textbf{48.0} & \textbf{30.2}    & \textbf{32.0}   & \textbf{89.5}  & \textbf{99.0}   \\ \midrule
\multicolumn{1}{l}{TL (Rej)}   & 91.8  & 49.8  & 43.3      & 97.0*     & {100.0}*   & {100.0}*    & 85.6 & 53.7    & 98.4   & 99.9  & 99.9   \\ 
\multicolumn{1}{l}{TL+C (Rej)}  & 92.3  & 56.8*  & \textbf{47.8}      & 97.7*     & {100.0}*   & {100.0}*    & 91.2 & \textbf{56.0}    & 98.6   & 99.9  & 99.9   \\
\multicolumn{1}{l}{TL+CC (Rej)} & 92.7  & \textbf{57.5}*  & 45.8      & \textbf{98.1}*     & {100.0}*   & {100.0}*    & \textbf{93.4} & 55.4    & \textbf{98.8}   & \textbf{100.0}* & \textbf{100.0}*  \\ \midrule
\multicolumn{1}{l}{TL (NR)}   & 99.2  & 55.5   & 58.5*      & 96.8     & 99.6    & 99.8     & 98.0* & 71.7*    & 99.3*   & 100.0* & 100.0*  \\ 
 \midrule \midrule
\multicolumn{1}{l}{FT}        & 98.6  & 25.6    & 21.9      & 12.9     & 20.0    & 96.3     & 51.2 & 47.2    & 75.6   & 95.7  & 98.2   \\
\multicolumn{1}{l}{FT+C}  & 99.1  & \textbf{31.7}   & \textbf{51.1}      & \textbf{18.0}     & \textbf{29.5}    & \textbf{97.7}     & \textbf{76.7} & \textbf{57.5}    & \textbf{88.8}   & \textbf{98.3}  & \textbf{98.9}   \\ 
\midrule
\multicolumn{1}{l}{FT (Rej)}  & 92.7  & 39.3*  & 36.8      & 90.0*     & 99.7*    & 99.7*     & 88.9 & 66.9    & 99.6*   & 99.7*  & 99.8*   \\
\multicolumn{1}{l}{FT+C (Rej)}  & 93.2  & \textbf{60.7}*  & \textbf{66.6}*      & \textbf{97.3}*     & \textbf{99.8}*    & \textbf{99.9}*     & \textbf{98.3}* & \textbf{82.2}*    & \textbf{99.9}*   & \textbf{99.9}*  & \textbf{99.9}*   \\ \midrule
\multicolumn{1}{l}{FT (NR)}  & 98.6  & 37.3   & 38.3      & 87.3     & 97.0    & 99.6     & 91.0 & 79.2    & 98.7   & 99.2  & 99.5   \\\bottomrule
\end{tabular}%
\end{table*}

\begin{table*}[!ht]
\centering
\caption{\ad{CIFAR-100 dataset. Vulnerability analysis of the classifiers against MKA and state-of-art attacks\rev{---classification quality is reported, the same of Figs.~\ref{fig:black}-\ref{fig:white} (second column)}. Refer to the caption of Table~\ref{tab:bb_animal} for more details \minor{(see \cite{autoattack} for a description of the compared attacks)}.}}
\label{tab:bb_cifar100}
\begin{tabular}{@{}lc|ccccc|ccccc@{}}
\toprule
        & \multicolumn{1}{l}{} & \multicolumn{5}{c}{White-box attacks ($\epsilon=0.03$)} & \multicolumn{5}{c}{Black-box transfer attacks ($\epsilon=0.03$)} \\ \midrule
  \multicolumn{1}{l}{Model} &
  \multicolumn{1}{c}{$\epsilon=0$} &
  \multicolumn{1}{c}{MKA} &
  \multicolumn{1}{c}{APGD-CE} &
  \multicolumn{1}{c}{APGD-T} &
  \multicolumn{1}{c}{FAB-T} &
  \multicolumn{1}{c}{Square} &
  \multicolumn{1}{c}{MKA} &
  \multicolumn{1}{c}{APGD-CE} &
  \multicolumn{1}{c}{APGD-T} &
  \multicolumn{1}{c}{FAB-T} &
  \multicolumn{1}{c}{Square} \\ \toprule
\multicolumn{1}{l}{TL}      & 51.0  & 21.9    & 22.2      & 21.6     & 22.3    & 51.4     & 23.1 & 23.7    & 23.9   & 39.5  & 52.7   \\
\multicolumn{1}{l}{TL+C}  & 52.9  & \textbf{27.4}      & 24.6      & 24.2    & 25.1    & \textbf{53.3}     & 32.3 & 35.5    & 37.9   & \textbf{48.4}  & \textbf{54.5}   \\
\multicolumn{1}{l}{TL+CC}  & 50.5  & 27.1     & \textbf{25.0}      & \textbf{24.7}     & \textbf{25.3}    & 49.5     & \textbf{35.5} & \textbf{38.6}    & \textbf{40.4}   & 46.9  & 51.5   \\
\midrule
\multicolumn{1}{l}{TL (Rej)}  & 48.1  & 26.9    & 33.2*      & 34.3*     & 34.4    & 59.2*     & 27.6 & 35.1    & 36.9   & 49.1  & 60.2*   \\
\multicolumn{1}{l}{TL+C (Rej)}  & 49.4  & \textbf{31.8}*    & \textbf{35.0}*      & \textbf{35.6}*     & \textbf{36.2}    & \textbf{60.6}*     & 40.7 & 44.8    & 47.0   & \textbf{56.0}*  & \textbf{61.0}*   \\
\multicolumn{1}{l}{TL+CC (Rej)} & 46.1  & 30.8*    & 34.0*      & 34.7*     & 35.4    & 55.7*     & \textbf{45.4} & \textbf{46.3}    & \textbf{47.6}   & 53.5*  & 57.0*   \\
\midrule
\multicolumn{1}{l}{TL (NR)}  & 49.0  & 30.5     & 30.1      & 24.5      & 39.6*     & 45.6      & 49.0* & 48.3*    & 49.0*   & 51.3  & 53.3   \\ 
\midrule \midrule
\multicolumn{1}{l}{FT}       & 59.4  & 29.0    & 26.4      & 26.0     & 26.7    & 57.2     & 48.4 & 49.1    & 49.7   & 55.5  & 59.5   \\
\multicolumn{1}{l}{FT+C}   & 60.0  & \textbf{31.4}    & \textbf{29.6}      & \textbf{28.3}     & \textbf{30.6}    & \textbf{60.1}     & \textbf{51.6} & \textbf{52.2}    & \textbf{52.8}   & \textbf{57.8}  & \textbf{61.0}   \\
\midrule
\multicolumn{1}{l}{FT (Rej)}   & 57.4  & 31.1    & 37.5*      & \textbf{42.0}*     & 41.1    & 66.1*     & 55.1 & 57.2*    & 58.4*   & 62.7*  & 66.2*   \\
\multicolumn{1}{l}{FT+C (Rej)}  & 56.7  & \textbf{37.6}*    & \textbf{37.8}*      & 41.1*     & \textbf{44.6}    & \textbf{67.0}*     & \textbf{60.2}* & \textbf{59.5}*    & \textbf{60.3}*   & \textbf{64.4}*  & \textbf{67.1}*   \\
\midrule
\multicolumn{1}{l}{FT (NR)}  & 59.7  & 36.5       & 35.3      & 30.4     & 50.9*     & 55.1      & 58.0 & 54.2    & 55.7   & 60.0  & 62.7   \\
\bottomrule
\end{tabular}%
\end{table*}

\ad{To compare our defense with NR, we have considered four different state-of-the-art evasion attacks: APGD-CE, APGD-T, FAB-T, and Square, implemented within the framework of AutoAttack~\cite{autoattack}. APGD-CE (APGD-T) is an indiscriminate (targeted) step-free variant of the famous attack called PGD~\cite{madry18-iclr}. Unlike PGD, the step size reduction is not  scheduled a priori but instead governed by the optimization function trend. Moreover, both APGDs attack use momentum. FAB-T is the targeted version of an attack called Fast Adaptive Boundary Attack (FAB)~\cite{croce20-pmlr}, which tries to find the minimum distance sample beyond the boundary of the desired class. The Square attack~\cite{andriushchenko20-eccv}, differently from the previously mentioned ones, is a black-box attack; namely, it can query the classifier obtaining the predicted scores without exploiting any knowledge of the model architecture. By default, APGD-CE makes five random restarts, whereas the targeted versions of the attacks, i.e., APGD-T and FAB-T, run the attack nine times, each setting the target class as one of the nine top classes except the true class.}

\myparagraph{Adversarial Evaluation.} 
\ad{In our experiments, we fixed the maximum allowed perturbation \sm{$\epsilon$ to $0.5$ and \minor{$0.03$} on the ANIMALS and CIFAR-100 datasets, respectively -- the values in the middle of x-axis of Fig\rev{s}.~\ref{fig:black}\rev{-\ref{fig:white}} --} and we used the default value for all the other attacks' parameters. 
Table~\ref{tab:bb_animal} and Table~\ref{tab:bb_cifar100} report the results of this analysis, 
showing the 
\rev{classification quality (same measure of Figs.~\ref{fig:black}-\ref{fig:white})} on the clean (unmodified) test set \sm{$\mathcal{T}$} ($\epsilon=0$) and on the \sm{attacked instances of $\mathcal{T}$} generated \sm{in the same white-box and black-box scenarios described in Section~\ref{sect:exp-multi}.} 
\rev{As expected, white-box attacks are more effective than the black-box ones, reducing the model accuracy in a more evident manner}.
Results confirm that \sm{both} the domain knowledge introduced at training time (+C and +CC) or exploited to implement the proposed rejection mechanism \sm{(Rej)} improve the model robustness against all the considered attacks, and jointly using these strategies further improves it. 
\sm{On average, the performances of the unconstrained classifiers (TL and FT) paired with the proposed knowledge-based rejection are comparable with the ones paired with NR, even though they clearly behave in different manners across the datasets/attacks. Differently, when also considering constrained models (+C and +CC), in most of the cases we can find a classifier with rejection (Rej) that outperforms the unconstrained classifiers equipped with NR. On the clean samples ($\epsilon=0$), the knowledge-based rejection criterion resulted more aggressive than NR.}


MKA and APGD-CE/T are more effective than the other attacks. \sm{On average,} their performances are comparable, and they depend on both the considered model and the dataset. In CIFAR-100, MKA outperforms APGD-CE/T on the black-box transfer scenario and against the model equipped with the proposed rejection mechanism \sm{(Rej)}, whereas on the ANIMALS dataset, APGD-CE usually obtained better results. APGD-CE/T leverages an optimization strategy that is more advanced than the one of MKA that, differently from APGD-CE/T, is designed to be used in multi-label problems too. For example, \ad{APGD-CE/T makes several attack restarts and uses a special type of adaptive step size.}
In the white-box setting, attacks yield a larger reduction of the performances. However, in the case of ANIMALS, the proposed rejection mechanism is still robust to all the attacks, with the exceptions of MKA, that is knowledge aware, and of APGD-CE.
The fine-tuned optimization procedure in APGD-CE allows the attack to create samples that are confidently misclassified, and they end-up in belonging to space regions in which the classification functions are paired (Def.~\ref{pairing}). In CIFAR-100, the rejection mechanism still has a positive impact, even if it is less significant than in ANIMALS.}

\subsubsection{\minor{In-depth Analysis}}
\sm{We further analyzed our results, visualizing the behavior of all the compared attacks in terms of value of the constraint loss of Eq.~(\ref{test}) and of the supervision loss -- first term of Eq.~(\ref{train}).}
\gc{Figs.~\ref{fig:pairing_animals}-\ref{fig:pairing_cifar} \sm{show each generated adversarial example, highlighting them with different markers/colors in function of the corresponding attack procedure, on the ANIMALS and CIFAR-100 datasets, respectively, \rev{black-box (i.e., the constraint loss is measured for the purpose of determining whether to reject or not an example)}. \gc{Samples that are rejected are indicated with crosses, while circles represent the non-rejected ones.}}
The dotted line is about the rejection threshold $\tau$ from Eq.~ (\ref{eq:reject}).

\begin{figure}[ht]
    \centering
    \includegraphics[width=0.93\columnwidth]{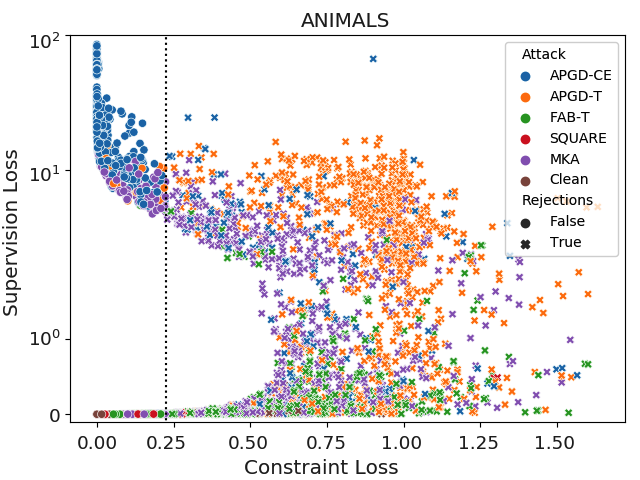}
    \caption{Adversarial data generated ($\epsilon = 0.5$) by different attacks -- ANIMALS, TL+C(Rej), black-box. Examples that are rejected/not-rejected by the proposed knowledge-based criterion are depicted with crosses/circles (``Clean'' indicates unaltered examples from the test set; the vertical line is the reject threshold).}
    \label{fig:pairing_animals}
\end{figure}
\begin{figure}[ht]
    \centering
    \includegraphics[width=0.93\columnwidth]{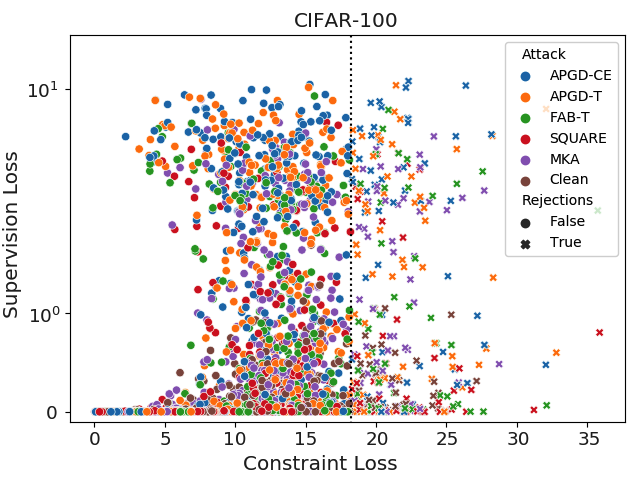}
    \caption{Adversarial data generated ($\epsilon = 0.03$) by different attacks -- CIFAR-100, TL+C(Rej), black-box. See Fig.~\ref{fig:pairing_animals}.}
    \label{fig:pairing_cifar}
\end{figure}

\sm{In line with what we observed in the numerical results, in the case of ANIMALS, Fig.~\ref{fig:pairing_animals}, it is evident how APGD-CE is actually able to craft attacks that strongly increase the supervision loss, still fulfilling the constraints (top-left area). Differently, the other attacks are not able to reach such result, so that their data is localized in high-constraint loss regions, easily rejected by the proposed technique, especially FAB-T, while Square actually fails in generating evident attacks.} 
It is interesting to notice the D-shaped white region over the origin. \sm{It is an area in which constraints are almost fulfilled and the loss function can reach significantly non-null values, but no attacks fall there. This suggests that it is not straightforward to increase the supervision loss without violating the constraints. However, there are more extreme APCG-CE configurations with the largest supervision losses that also fulfill the knowledge (Fig.~\ref{fig:pairing_animals}, top-left area). Of course, this depends on several factors, such as the type of domain knowledge that is available, the way we selected to convert it into polynomial constraints, and the constraint enforcement scheme, thus opening to future improvements. } 
Moving to the CIFAR-100 dataset, Fig.~\ref{fig:pairing_cifar}, 
\rev{we observe different patterns with respect to the case of ANIMALS. This was clearly expected, since the two datasets differ both in terms of the problem they consider, the number of classes and in terms of the known relationships among such classes, described by the dataset-specific domain knowledge and embedded into the constraint loss}. However, we can still observe the D-shaped region over the origin, even if in a less significant manner. On this dataset, the rejection rates are generally lower than ANIMALS. This is mostly due to the fact the constraint loss is larger also on the unaltered data, due to the \rev{already mentioned} different problem and different type of domain knowledge. As a matter of fact, we have also a larger reject threshold $\tau$. In this case, the behavior of the different attack strategies is more coherent, remarking previous considerations on the role of knowledge in shaping the distribution of the attacks.}

\section{Related Work}
\label{sec:rel}
\sm{In addition to the literature described in Section~\ref{sec:intro}, we further emphasize the main differences of what we propose with respect to the most strongly related approaches.}


\myparagraph{\sm{Multi-label adversarial perturbations.}} Most of the work in the adversarial ML area focuses on single-label classification problems. To the best of our knowledge, the first and only study on this problem is the one in \cite{8594975}, in which the authors focus on targeted multi-label adversarial perturbations defining in advance the set of classes on which the attack is targeted (being them positive or negative) and also introducing another set of classes for which the attack is expected not to change the classifier predictions. The framework described in \cite{8594975} is only experimented in a {\it static/targeted} context, i.e., by selecting in advance the sets mentioned above using custom criteria to simulate the attacking scenario artificially. The multi-label attack that we propose in this work is instead {\it dynamic/untargeted} and without the need of defining in advance what are the classes to be considered. Regarding the defenses, \sm{to our best knowledge}, none of the previously proposed ones leverage multi-label classification outputs. 

\myparagraph{Semi-supervised Learning and Adversarial Training.} 
\sm{In the context of adversarial machine learning, unlabeled data are usually employed to improve the robustness of the classifier} by performing adversarial training. The rationale \sm{behind such training scheme} is that if \sm{the available unlabeled} samples are perturbed, then the \sm{predicted class} should not change. 
Miyato et al. \cite{miyato16,miyato2018virtual} and Park et al. \cite{park2018adversarial} exploit adversarial training (virtual adversarial training and adversarial dropout, respectively) to favor regularity around the supervised and unsupervised training data, and to improve the classifier performance. The work in~\cite{akcay2018ganomaly} develops an anomaly detector using adversarial training in the semi-supervised setting. Self-supervised learning is exploited in~\cite{carmon2019unlabeled,najafi2019robustness} to \sm{gain stronger} adversarial robustness. Stability criteria are enforced on unlabeled training data in~\cite{zhai2019adversarially}, whereas the work in~\cite{alayrac2019arelabels} specifically focuses on an unsupervised adversarial training procedure in the context of semi-supervised classification. 
Our model neither exploits adversarial training nor any adversary-aware training criteria aimed at gaining intrinsic regularity. 
We focus on the role of domain knowledge as an indirect means to increase adversarial robustness and, afterward, to detect adversarial examples. 
Therefore, the proposed work is not attack-dependent, and it is faster at training time as it does not require generating adversarial examples. 
We believe that using unlabeled data also to simulate attacks and incorporate them into the training process may further improve robustness.  All the described methods could also be applied jointly with what we propose.

\myparagraph{Rejection-based Approaches for Adversarial Examples.} 
A different line of defenses, complementary with adversarial training, is based on detecting and rejecting samples sufficiently far from the training data in feature space. 
Our approach differs from other adversarial-example detectors~\cite{carlini2017adversarial,ma2018characterizing,samangouei2018defensegan,miller2020adversarial} as it has no additional training cost and negligible runtime cost. 
We are the first to show that domain knowledge can be used to reject adversarial examples and also to propose a detector that exploits unlabeled data.

\rev{\myparagraph{Domain-Agnostic  Methods  and  Semantic Attacks.} 
Recent work in adversarial attacks considers the role of the learning domain and of additional semantic information, even if with different goals to the ones of this paper. The way the learning domain is related to the generation of attacks was recently studied in \cite{A}, that is based on the idea of developing generative adversarial perturbations that turn out to be easily transferable from the source domain (where the attack function is modeled) to another domain. Differently, we focus on knowledge that is domain specific and used both for defending and creating more informed attacks. The knowledge of a set semantic attributes is used to implement the threat model of semantic adversarial attacks in \cite{B}. A generative network is considered, and the attack procedure focuses on altering the activation of such human-understandable attributes, that, in turn, yield visible changes in the input image (e.g., adding glasses to the input face). Differently, our work is built on an $L_p$-norm-bounded perturbation model that  does not enforce the input image to change in a human-understandable manner. Our approach considers a more generic notion of knowledge, that includes information also on the relationships within subsets of logic predicates, and that exploits the power of FOL. Predicate activations are modeled by neural networks and not by scalar variables as for the attributes of \cite{B}.}

\section{Conclusions and Future Work}
\label{sec:concl}
In this paper we investigated the role of domain knowledge in adversarial settings. Focusing on multi-label classification, we injected knowledge expressed by First-Order Logic in the training stage of the classifier, not only with the aim of improving its quality, but also as a mean to build a detector of adversarial examples at no additional cost. We proposed a multi-label attack procedure and showed that knowledge-constrained classifiers can improve their robustness against both black-box and white-box attacks, \sm{and, using the same knowledge, they can detect adversarial attacks}. We believe that these findings will open the investigation of domain knowledge as a feature to further improve the robustness of multi-label classifiers against adversarial attacks.

\sm{The proposed adversarial example rejection scheme is based on the idea of dealing with classifiers that fulfill the knowledge-related constraints over the space regions in which the non-malicious data are distributed,  not guarantying such fulfillment in the rest of the space. 
While this is experimentally evaluated to be a key ingredient to profitably build a rejection strategy, we showed that advanced optimization strategies can fool the defense injecting a stronger perturbation than the one used to fool the undefended system.
In future work we will consider intermixing adversarial training with knowledge constraints, to strengthen the violation of the constraints out of the distribution of the real data. We also plan to design a learnable model that decides whether to reject or not in function of the fulfillment of each specific logic formula, going beyond a simple-but-effective threshold on the cumulative constraint loss.}

\section*{Acknowledgment}
This work was partly supported by the PRIN 2017 project RexLearn, funded by the Italian Ministry of Education, University and Research (grant no. 2017TWNMH2).

\bibliographystyle{IEEEtran}
\bibliography{IEEEabrv,main}

%

\begin{IEEEbiography}[{\includegraphics[width=1in,height=1.25in,clip,keepaspectratio]{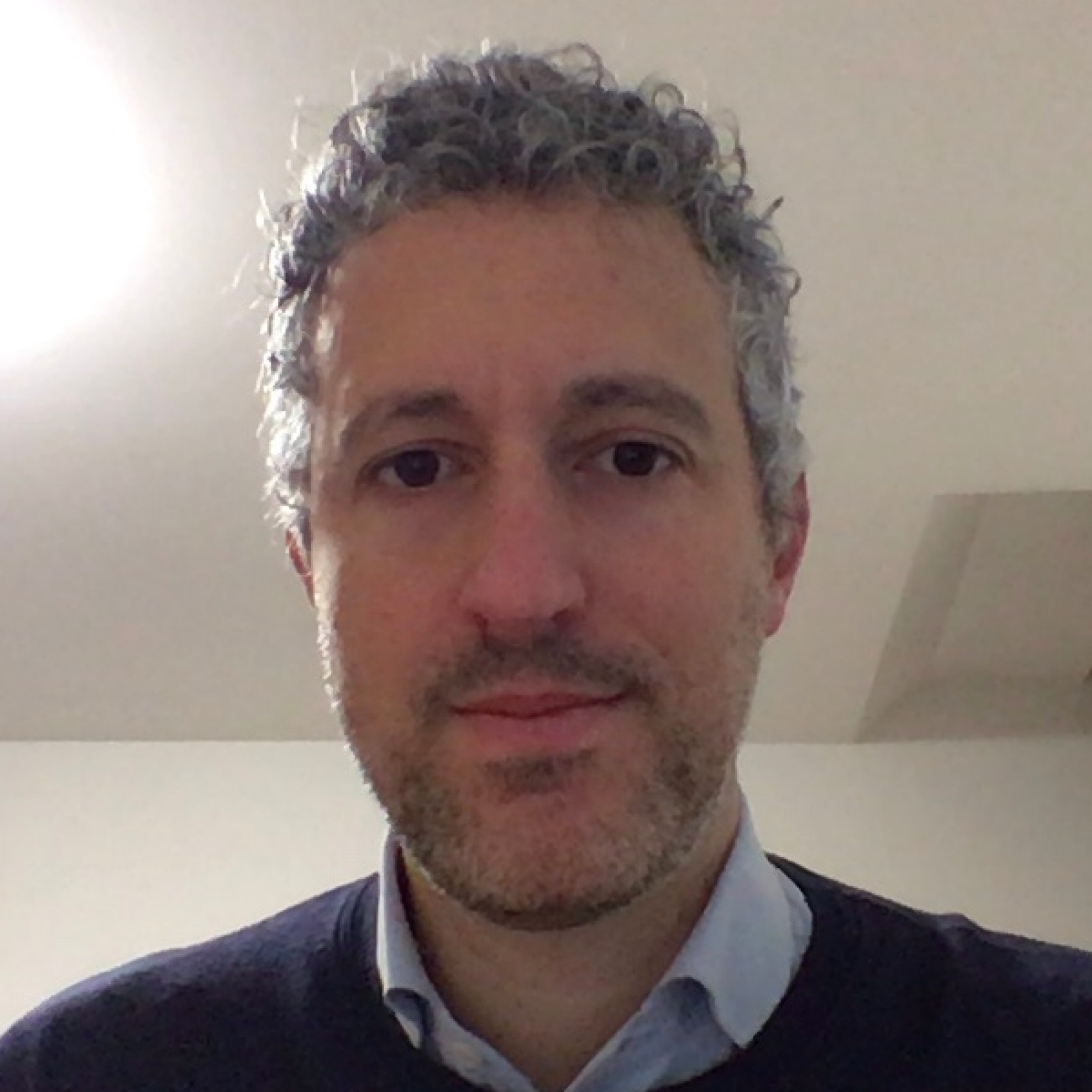}}]{Stefano Melacci} received the M.S. Degree in Computer Engineering (cum laude) and the PhD degree in Computer Science (Information Engineering) from the University of Siena, Italy, in 2006 and 2010, respectively. He worked as Visiting Scientist at the Computer Science and Engineering Department of the Ohio State University, Columbus, USA, and he is currently Associate Professor of the Department of Information Engineering and Mathematics, University of Siena. His research interests include machine learning and pattern recognition, mainly focused on Neural Networks and Kernel Machines, with applications to Computer Vision and Natural Language Processing tasks. He serves as Associate Editor of the IEEE Transactions on Neural Network and Learning Systems.
\end{IEEEbiography}

\begin{IEEEbiography}[{\includegraphics[width=1in, height=1.25in, clip,keepaspectratio]{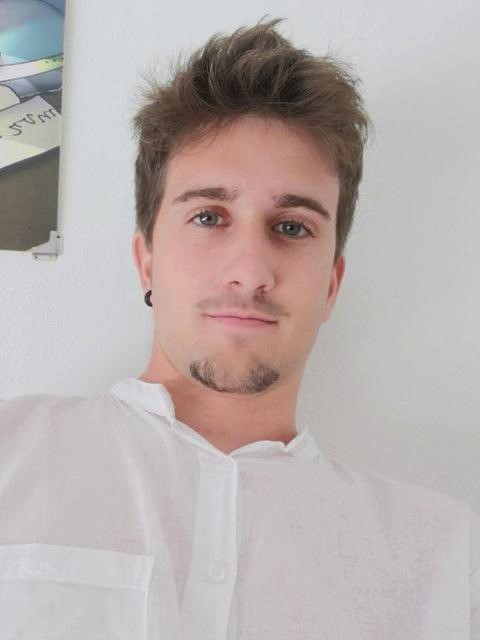}}]{Gabriele Ciravegna} is a PhD student at the University of Siena since 2018 under the supervision of Professor Marco Gori. In 2018 he received the master’s degree in Computer Engineering with honors at the Polytechnic University of Turin. He has always been interested in the machine learning field. Nowadays, he is focused on overcoming the intrinsic limits of machine learning and neural networks, especially in the context of Explainable AI. He presented his works in several international venues such as AAAI, IJCAI, IJCNN. He also serves as reviewer in conferences and journals that are about Neural Networks, such as IEEE Transactions on Neural Networks and Learning Systems.
\end{IEEEbiography}

\begin{IEEEbiography}[{\includegraphics[width=1in,height=1.25in,clip,keepaspectratio]{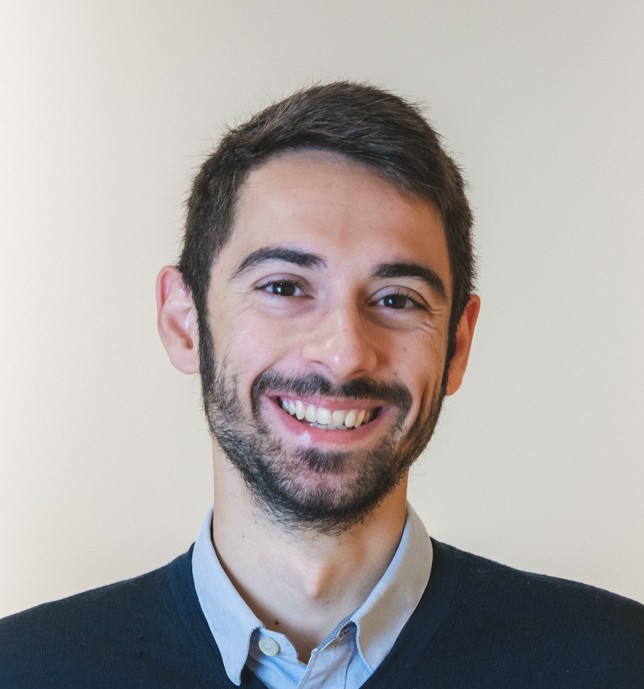}}]{Angelo Sotgiu} is a Ph.D. student in Electronic and Computer Engineering at the University of Cagliari, Italy. He received his M. Sc. in Telecommunication Engineering with honors from the University of Cagliari, Italy, in 2019. His research interests include secure machine learning and computer security. He serves as a reviewer for several journals and conferences in the machine learning and computer security area.
\end{IEEEbiography}

\begin{IEEEbiography}[{\includegraphics[width=1in,height=1.25in,clip,keepaspectratio]{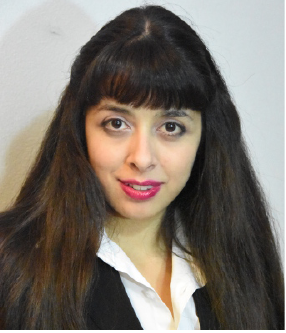}}]{Ambra Demontis} is an Assistant Professor at the University of Cagliari, Italy. She received her M.Sc. degree (Hons.) in Computer Science and her Ph.D. degree in Electronic Engineering and Computer Science from the University of Cagliari, Italy, respectively, in 2014 and 2018.
Her research interests include secure machine learning, kernel methods, and computer security.
In particular, she has provided contributions to the design of secure machine learning systems in the presence of intelligent attackers and highlighted interesting trade-offs between their complexity and security.
She co-organizes the AISec workshop, serves on the program committee of different conferences and workshops, such as IJCAI and DLS, and as a reviewer for several journals, such as TNNLS, TOPS Machine Learning, and Pattern Recognition. She is a Member of the IEEE and the IAPR.
\end{IEEEbiography}

\begin{IEEEbiography}[{\includegraphics[width=1in,height=1.25in,clip,keepaspectratio]{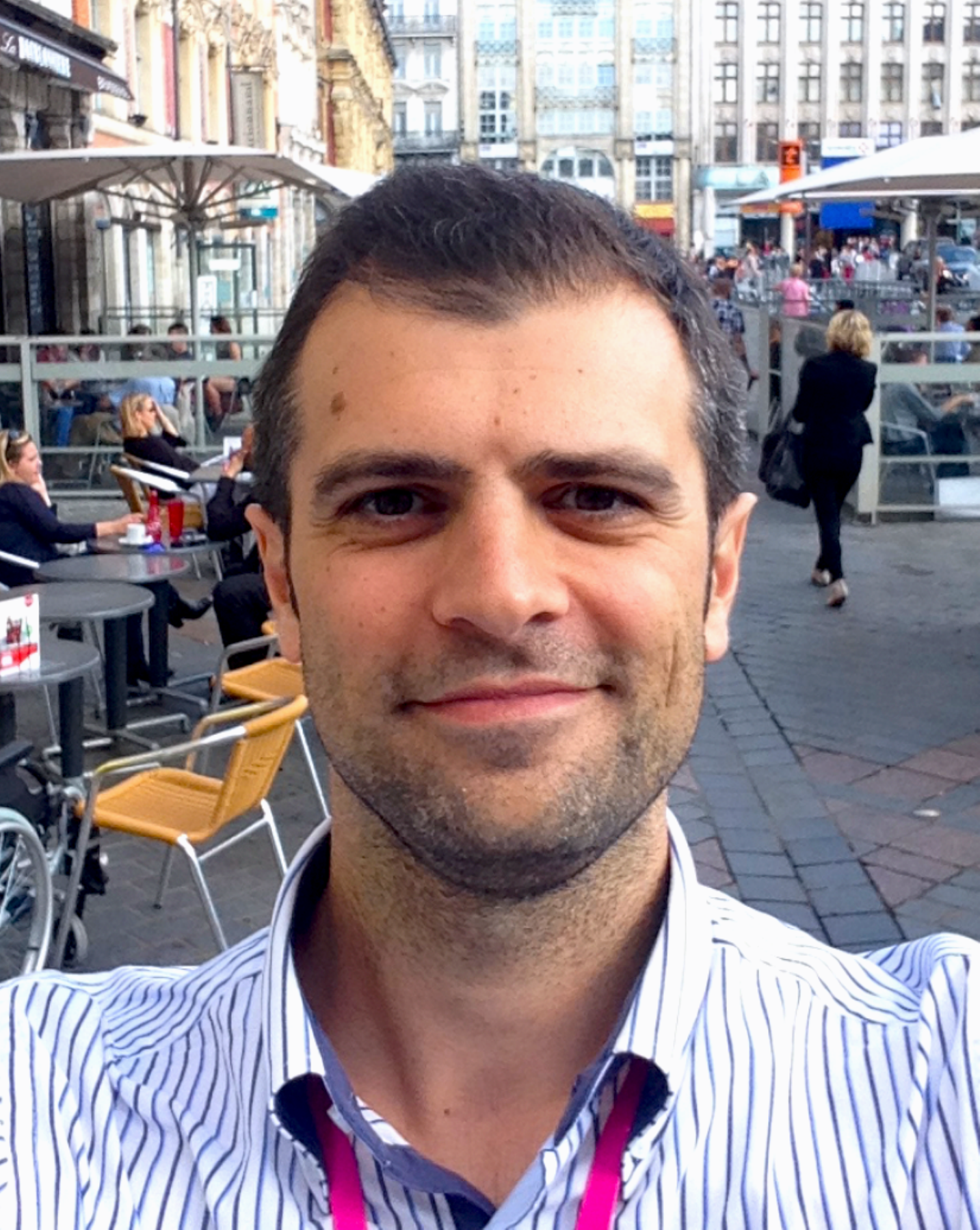}}]{Battista Biggio} (MSc 2006, PhD 2010) is Assistant Professor at the University of Cagliari, Italy, and co-founder of Pluribus One (\url{https://www.pluribus-one.it}). His research interests include machine learning, biometrics and cybersecurity. He has provided pioneering contributions in the area of adversarial machine learning, demonstrating gradient-based evasion and poisoning attacks, and how to mitigate them, playing a leading role in the establishment and advancement of this research field. He regularly serves as a program committee member for the most prestigious conferences and journals in the area of machine learning and computer security (ICML, NeurIPS, ICLR, IEEE SP, USENIX Sec.). He chaired the IAPR TC on Statistical Pattern Recognition Techniques, co-organized the S+SSPR, AISec and DLS workshops, and serves as Associate Editor for Pattern Recognition, IEEE TNNLS and IEEE CIM. Dr. Biggio is a senior member of the IEEE and member of the IAPR and of the ACM.
\end{IEEEbiography}

\begin{IEEEbiography}[{\includegraphics[width=1in,height=1.25in,clip,keepaspectratio]{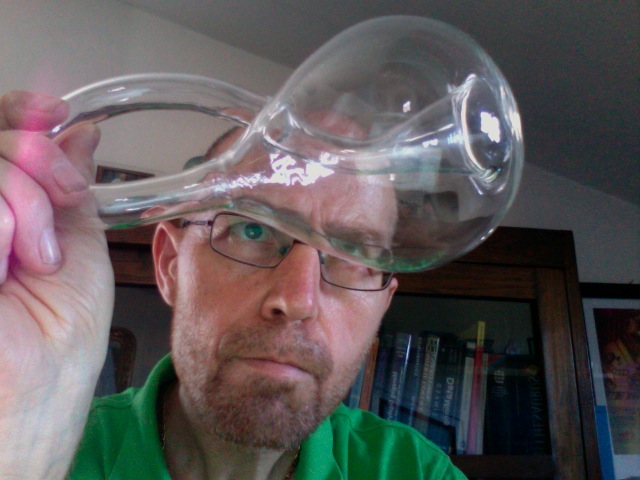}}]{Marco Gori} 
received the Ph.D degree in 1990 from University  of Bologna, Italy, working partly at the School of Computer Science (McGill University, Montreal). 
He is currently full professor at the University of Siena and he is mostly interested in
machine learning with applications to pattern recognition, Web mining, and game playing. He has recently published the monograph ``Machine Learning: A constraint-based approach,'' (MK, 560 pp., 2018), which contains a unified view of his approach to machine learning. 
His pioneering role in neural networks has been emerging especially from the recent interest in Graph Neural Networks, 
that he contributed to introduce in the paper  ``Graph Neural Networks,'' IEEE-TNN, 2009.
Professor Gori has been the chair of the Italian Chapter of the IEEE Computation Intelligence Society and the President of the Italian Association for Artificial Intelligence. He is a Fellow of IEEE, EurAI, and IAPR. He was one the first people involved in European project on Artificial Intelligence CLAIRE, and he is currently a Fellow of Machine Learning association ELLIS. He is in the scientific committee of ICAR-CNR and is the President of the Scientific Committee of FBK-ICT. Dr. Gori is currently holding an international 3IA Chair at the Universit\'e Cote d’Azur.
\end{IEEEbiography}

\begin{IEEEbiography}[{\includegraphics[width=1in,height=1.25in,clip,keepaspectratio]{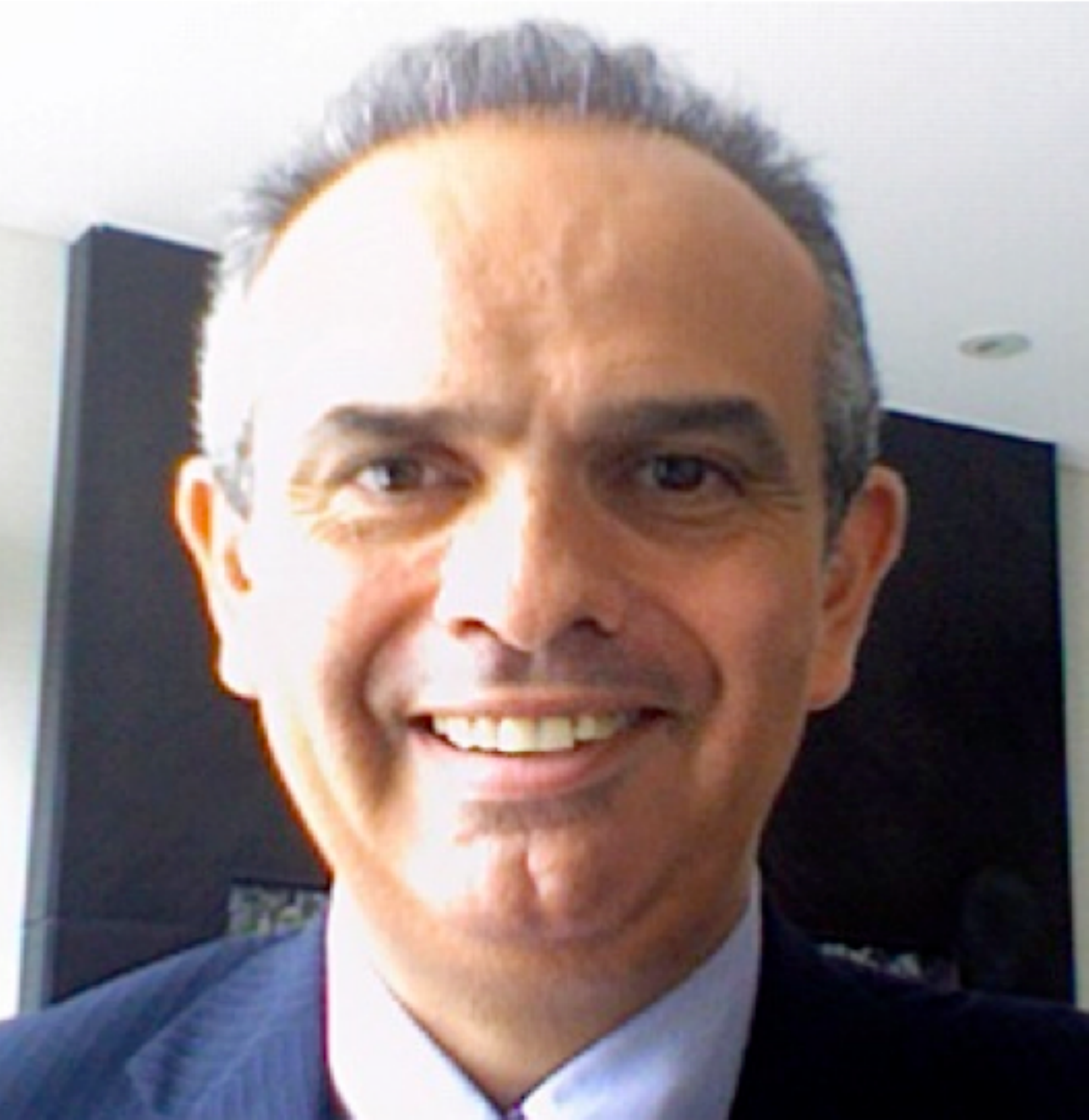}}]{Fabio Roli} received his Ph.D. in Electronic Engineering from the University of Genoa, Italy. He was a research group member of the University of Genoa ('88-'94), and adjunct professor at the University of Trento ('93-'94). In 1995, he joined the Department of Electrical and Electronic Engineering of the University of Cagliari, where he is now Full Professor of Computer Engineering and Director of the Pattern Recognition and Applications laboratory (\url{https://pralab.diee.unica.it/}). He is partner and R\&D manager of the company Pluribus One that he co-founded (\url{https://www.pluribus-one.it}). He has been doing research on the design of pattern recognition and machine learning systems for thirty years. 
He was a very active organizer of international conferences and workshops, and established the popular workshop series on multiple classifier systems. Dr. Roli is Fellow of the IEEE and of the IAPR.
\end{IEEEbiography}





\clearpage

\clearpage
\appendices
\begin{center}
\Large\textbf{Supplementary Material}
\vskip 1mm
\end{center}

We report here some additional details on the attack optimization process and on the parameter settings used in our experiments, along with the complete list of the domain-knowledge constraints available for the considered datasets.

\section{Attack Optimization}

\begin{figure*}[!ht]
    \centering
    \includegraphics[width=0.85\textwidth]{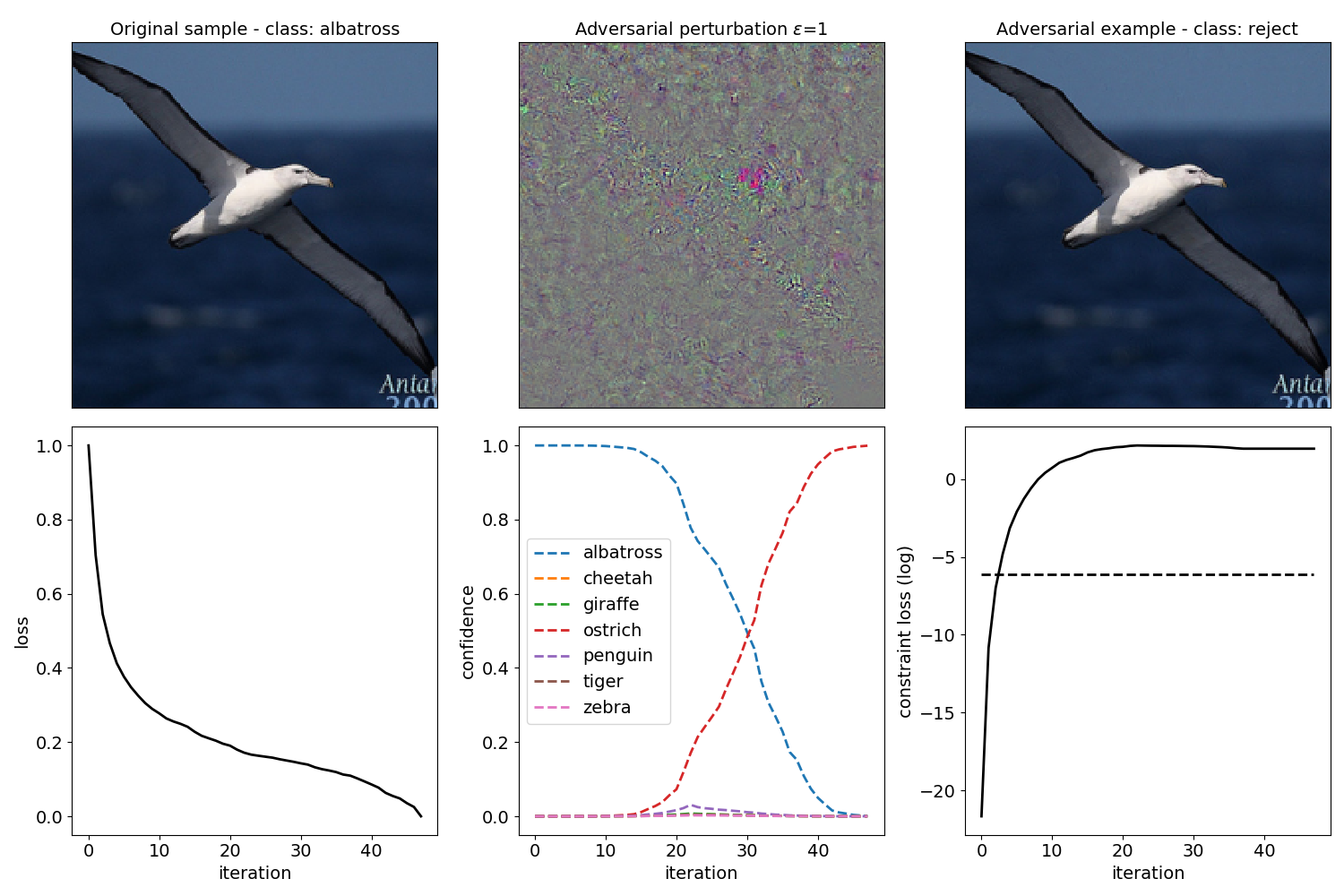}
    \caption{Black-box attack on the ANIMALS dataset. While the attack is able to flip the initial prediction from \textit{albatross} to \textit{ostrich}, the attack is eventually detected as the constraint loss remains above the rejection threshold (dashed black line).}
    \label{angelo1}
\end{figure*}

\begin{figure*}[!ht]
    \centering
    \includegraphics[width=0.85\textwidth]{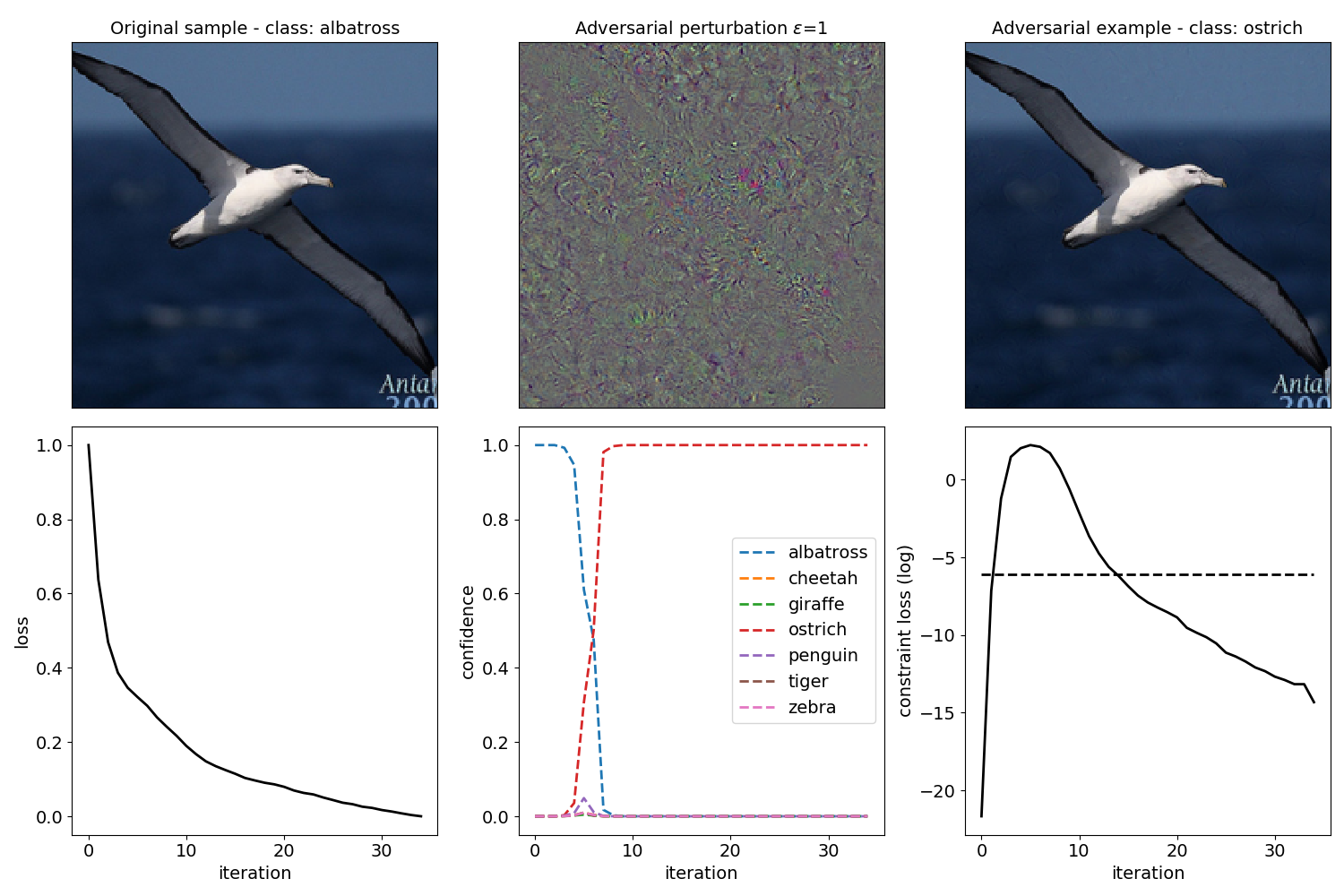}
    \caption{White-box attack on the ANIMALS dataset. The attack is able to flip the initial prediction from \textit{albatross} to \textit{ostrich}, and then starts reducing the constraint loss which eventually falls below the rejection threshold (dashed black line). The attack sample remains thus undetected.}
    \label{angelo2}
\end{figure*}

Our attack optimizes Eq.~\eqref{attacklogits} via projected gradient descent. Black-box attacks are non-adaptive, and thus ignore the defense mechanism. For this reason, the constraint loss term $\varphi$ in our attack is ignored by setting its multiplier $\alpha=0$ and $\kappa=\infty$.
For white-box attacks on ANIMALS and PASCAL-PART, we set $\alpha=0.1$ and $\alpha=1$, respectively, while setting $\kappa=2$. 
These values are chosen to appropriately scale the values of the constraint loss term $\varphi$ w.r.t. the logit difference (i.e., the first term in Eq.~\ref{attacklogits}, lower bounded by $-2\kappa$). This is required to have the sample misclassified while also fulfilling the domain-knowledge constraints. The process is better illustrated in Figs.~\ref{angelo1} and \ref{angelo2}, in which we respectively report the behavior of the black-box and white-box attack optimization on a single image from the ANIMALS dataset, with $\epsilon=1$.
In particular, in each Figure we report the source image, the (magnified) adversarial perturbation, and the resulting adversarial examples, along with some plots describing the optimization process, i.e., how the attack  loss of Eq.~\eqref{attacklogits} is minimized across iterations, and how the softmax-scaled outputs on the main classes and the logarithm of the constraint loss $\varphi$ change accordingly.

In both the black-box and white-box cases, the attack loss is progressively reduced during the iterations of the optimization procedure. While the \textit{albatross} prediction is progressively transformed into \textit{ostrich}, the constraint loss increases across iterations, exceeding the rejection threshold. Thus, the adversarial example is correctly detected. 
Similarly, the white-box attack is able to initially flip the prediction from \textit{albatross} to \textit{ostrich}, allowing the constraint loss to increase. However, after this initial phase, the attack correctly reduces the constraint loss after its initial bump, bringing its value below the rejection threshold.
The system thus fails to detect the corresponding adversarial example.
Finally, it is also worth remarking that, in both cases, the final perturbations do not substantially compromise the source image content, remaining essentially imperceptible to the human eye.


\section{Domain Knowledge}
\label{app:rules}
Each dataset is composed of a set of classes that, for convenience, we associate to logic predicates. 
Such predicates participate in First-Order Logic (FOL) formulas that model the available domain knowledge. 
The FOL formulas that define the domain knowledge of the ANIMALS, CIFAR-100 and PASCAL-Part data are reported in Table~\ref{animalsrules}, Table~\ref{cifar100rules}, and Table~\ref{pascalpartrules}, respectively, where each predicate is indicated with capital letters. In each table (bottom part) we also report those rules that are about activating at least one of the classes of each level of the hierarchy. Following the nomenclature used in the paper, the main classes of the ANIMALS dataset are {\scriptsize ALBATROSS, GIRAFFE, CHEETAH, OSTRICH, PENGUIN, TIGER, ZEBRA}, while the other categories are {\scriptsize MAMMAL, HAIR, MILK, FEATHERS, BIRD, FLY, LAYEGGS, MEAT, CARNIVORE, POINTEDTEETH, CLAWS, FORWARDEYS, HOOFS, UNGULATE, CUD, EVENTOED, TAWNY, BLACKSTRIPES, LONGLEGS, LONGNECK, DARKSPOTS, WHITE, BLACK, SWIM, BLACKWHITE, GOODFLIER}. In the case of the CIFAR-100 dataset, the main classes are the ones associated with the predicates of Table~\ref{cifar100rules} that belong to the premises of the shortest FOL formulas (i.e., the formulas in the form {\scriptsize A$(x)$ $\Rightarrow$ B$(x)$}, where the main class is {\scriptsize A}). Formulas in PASCAL-Part are relationships between objects and object-parts. The same part can belong to multiple objects, and in each objects several parts might be visible. See Table~\ref{pascalpartrules} for the list of classes (main classes are in the premises of the second block of formulas).

In ANIMALS and CIFAR-100, a mutual exclusion predicate is imposed on the main classes. As a matter of fact, in these two datasets, each image is only about a single main class. The $\verb|mutual_excl(p_1, p_2, ..., p_n)|$ predicate defined below, can be devised in different ways. 
The first, straightforward approach consists in considering the disjunction of the true cases in the truth table of the predicate: 
\begin{equation}
\begin{aligned}
    & mutual\_excl(p_1, p_2, ..., p_n) = \\
    & \bigvee^n_{i=0} \left( p_i(x) \land \bigwedge^n_{j=0, j \ne i}\neg p_j(x) \right), \quad i,j \in M,
\end{aligned}
\end{equation}
where $M$ is the set of the main classes, with cardinality $n$ and $p_i(x)$ is the logic predicate corresponding to the $i$-th output of the network $f_i(x)$.  
This formulation of the $\verb|mutual_excl|$ predicate is what we used in the ANIMALS dataset. When there are seveal classes, as in CIFAR-100, this formulation leads to optimization issues, since it turned out to be complicated to find a good balance between the effect of this constraint and the supervision-fitting term. For this reason, the mutual exclusivity in CIFAR-100 was defined as a disjunction of the main classes followed by a set of implications that are used to implement the mutual exclusion of the predicates,
\begin{equation}
\begin{aligned}
    & mutual\_excl(p_1, p_2, ..., p_n) = \\ 
    & \begin{cases}
    \bigvee^n_{i=0}p_i(x), \\ 
    p_i(x) \Rightarrow \bigwedge^n_{j=0, j \ne i}\neg p_j(x), & \forall i \in M
    \end{cases}
\end{aligned}
\end{equation}
that resulted easier to tune, since we have multiple soft constraints that could be eventually violated to accommodate the optimization procedure.

\minor{In the case of the ANIMALS dataset, we also considered a noisy setting in which we artificially altered the FOL rules of Table~\ref{animalsrules} in order to make them not fully coherent with the (real) domain knowledge. We describe the resulting noisy knowledge bases in Table~\ref{tab:ka}, Table~\ref{tab:kb}, and Table~\ref{tab:kc}, reporting only the changes with respect to Table~\ref{animalsrules}. The knowledge base of Table~\ref{tab:ka} has been obtained by altering four of the existing rules, while knowledge of Table~\ref{tab:kb} is the outcome of \textit{adding} four new rules. In both the cases, we considered two implications whose conclusions are about main classes and two other implications whose conclusions are about auxiliary classes. Finally, Table~\ref{tab:kc} is about a noisy knowledge base where we relaxed the main-class-oriented conclusions of four implications. Such knowledge has been created by manually extending the conclusions using the disjunction operator, thus tolerating multiple configurations of the main classes.}

\begin{figure*}[th]
    \centering
    \includegraphics[width=0.8\textwidth, trim={0 100 0 100}, clip]
    {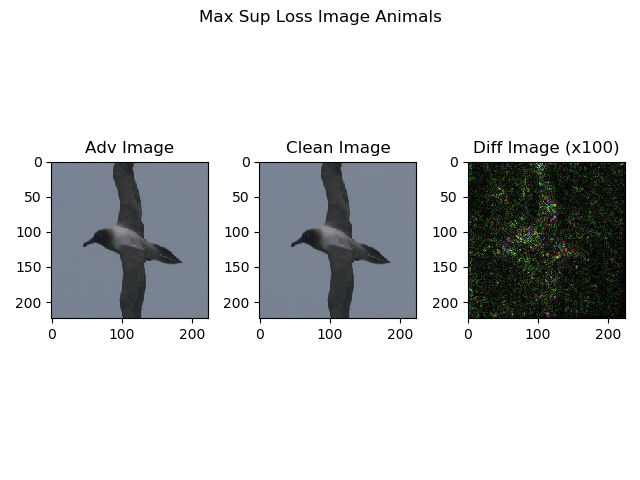}
    \caption{Adversarial examples with highest supervision loss (low constraint loss), APGD-CE attack, ANIMALS dataset.}
    \label{fig:max_sup_loss}
\end{figure*}

\begin{figure*}[th]
    \centering
    \includegraphics[width=0.8\textwidth, trim={0 100 0 100}, clip]
    {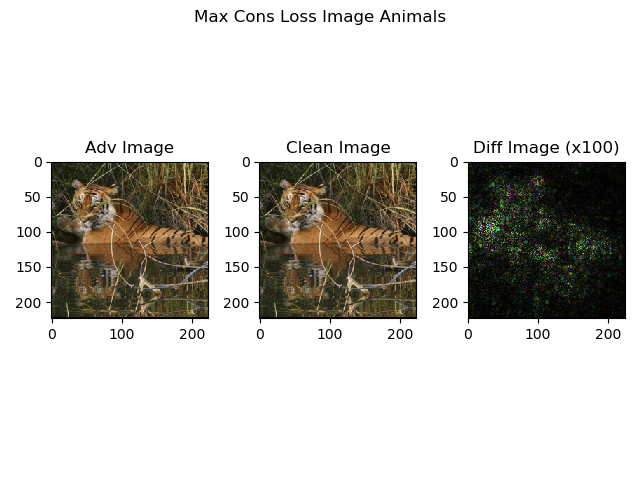}
    \caption{Adversarial examples with highest constraint loss (low supervision loss), APGD-CE attack, ANIMALS dataset.}
    \label{fig:max_cons_loss}
\end{figure*}

\gc{
\section{\sm{Results with SOTA attack strategies}}
In order to better support the experimental analysis of Section~\ref{sect:exp-single}, we report some examples of adverarial examples generated by the APGD-CE algorithm of the AutoAttack \cite{autoattack} library, ANIMALS dataset.
In particular, in Fig.~\ref{fig:max_sup_loss} and Fig.~\ref{fig:max_cons_loss}, we plot the two adversarial examples with highest supervision loss (and low constraint loss) and with highest constraint loss (and low supervision loss) (see also Fig.~\ref{fig:pairing_animals} of the main paper). No evident visual pattern is noticeable to distinguish the two cases.
}
\begin{table*}[!ht]
	\centering
	\caption{Domain knowledge, ANIMALS dataset.}
	\label{animalsrules}
	\vskip 2mm
	\begin{scriptsize}
	\begin{tabular}{|ll|}
		\hline
		& \\
		$\forall x$ &  HAIR$(x)$ $\Rightarrow $ MAMMAL$(x)$\\
		$\forall x$ &  MILK$(x)$ $\Rightarrow$ MAMMAL$(x)$\\
		$\forall x$ &  FEATHER$(x)$ $\Rightarrow$ BIRD$(x)$ \\
		$\forall x$ &  FLY$(x)$ $\land$ LAYEGGS$(x)$ $\Rightarrow$ BIRD$(x)$\\
		$\forall x$ &  MAMMAL$(x)$ $\land$ MEAT$(x)$ $\Rightarrow$ CARNIVORE$(x)$\\
		$\forall x$ &  MAMMAL$(x)$ $\land$ POINTEDTEETH$(x)$ $\land$ CLAWS$(x)$ $\land$ FORWARDEYES$(x)$ $\Rightarrow$ CARNIVORE$(x)$\\
		$\forall x$ &  MAMMAL$(x)$ $\land$ HOOFS$(x)$ $\Rightarrow$ UNGULATE$(x)$\\
		$\forall x$ &  MAMMAL$(x)$ $\land$ CUD$(x)$ $\Rightarrow$ UNGULATE$(x)$\\
		$\forall x$ &  MAMMAL$(x)$ $\land$ CUD$(x)$ $\Rightarrow$ EVENTOED$(x)$\\
		$\forall x$ &  CARNIVORE$(x)$ $\land$ TAWNY$(x)$  $\land$ DARKSPOTS$(x)$ $\Rightarrow$ CHEETAH$(x)$\\
        $\forall x$ &  CARNIVORE$(x)$ $\land$ TAWNY$(x)$  $\land$ BLACKSTRIPES$(x)$ $\Rightarrow$ TIGER$(x)$\\
        $\forall x$ &  UNGULATE$(x)$ $\land$ LONGLEGS$(x)$  $\land$ LONGNECK$(x)$  $\land$ TAWNY$(x)$  $\land$ DARKSPOTS$(x)$ $\Rightarrow$ GIRAFFE$(x)$\\
        $\forall x$ &  BLACKSTRIPES$(x)$ $\land$ UNGULATE$(x)$  $\land$ WHITE$(x)$ $\Rightarrow$ ZEBRA$(x)$\\
        $\forall x$ &  BIRD$(x)$ $\land$ $\neg$FLY$(x)$  $\land$ LONGLEGS$(x)$  $\land$ LONGNECK$(x)$  $\land$ BLACK$(x)$ $\Rightarrow$ OSTRICH$(x)$\\
        $\forall x$ &  BIRD$(x)$ $\land$ $\neg$FLY$(x)$  $\land$ SWIM$(x)$  $\land$ BLACKWHITE$(x)$ $\Rightarrow$ PENGUIN$(x)$\\
        $\forall x$ &  BIRD$(x)$ $\land$ GOODFLIER$(x)$ $\Rightarrow$ ALBATROSS$(x)$\\
        & \\
        \hline
        & \\
        $\forall x$ & \verb|mutual_excl|(ALBATROSS$(x)$, GIRAFFE$(x)$, CHEETAH$(x)$, OSTRICH$(x)$, PENGUIN$(x)$, TIGER$(x)$, ZEBRA$(x)$)\\
        $\forall x$ &  MAMMAL$(x)$ $\lor$  HAIR$(x)$ $\lor$ MILK$(x)$ $\lor$ FEATHERS$(x)$ $\lor$ BIRD$(x)$ $\lor$ FLY$(x)$ $\lor$ LAYEGGS$(x)$ $\lor$ MEAT$(x)$ \\
        & $\lor$ CARNIVORE$(x)$ $\lor$ POINTEDTEETH$(x)$ $\lor$ CLAWS$(x)$ $\lor$ FORWARDEYS$(x)$ $\lor$ HOOFS$(x)$ $\lor$ UNGULATE$(x)$\\
        & $\lor$ CUD$(x)$ $\lor$ EVENTOED$(x)$ $\lor$ TAWNY$(x)$ $\lor$ BLACKSTRIPES$(x)$ $\lor$ LONGLEGS$(x)$ $\lor$ LONGNECK$(x)$ \\
        & $\lor$ DARKSPOTS$(x)$ $\lor$ WHITE$(x)$ $\lor$ BLACK$(x)$ $\lor$ SWIM$(x)$ $\lor$ BLACKWHITE$(x)$ $\lor$ GOODFLIER$(x)$\\        
		& \\
		\hline
	\end{tabular}
	\end{scriptsize}
\end{table*}

\clearpage

\onecolumn
\captionof{table}{Domain knowledge, CIFAR-100 dataset.}
\begin{scriptsize}
\begin{longtable}{|ll|}
  		\hline
  		& \\
  		$\forall x$ &  AQUATIC MAMMALS$(x)$ $\Rightarrow$ (BEAVER$(x)$ $\lor$ DOLPHIN$(x)$ $\lor$ OTTER$(x)$ $\lor$ SEAL$(x)$ $\lor$ WHALE$(x)$) \\
  		$\forall x$ &  BEAVER$(x)$ $\Rightarrow$ AQUATIC MAMMALS$(x)$ \\
  		$\forall x$ &  DOLPHIN$(x)$ $\Rightarrow$ AQUATIC MAMMALS$(x)$ \\
  		$\forall x$ &  OTTER$(x)$ $\Rightarrow$ AQUATIC MAMMALS$(x)$ \\
  		$\forall x$ &  SEAL$(x)$ $\Rightarrow$ AQUATIC MAMMALS$(x)$ \\
  		$\forall x$ &  WHALE$(x)$ $\Rightarrow$  AQUATIC MAMMALS$(x)$ \\
		& \\
  		$\forall x$ &  FISH$(x)$ $\Rightarrow$ (AQUARIUM FISH$(x)$ $\lor$ FLATFISH$(x)$ $\lor$ RAY$(x)$ $\lor$ SHARK$(x)$ $\lor$ TROUT$(x)$) \\
  		 $\forall x$ &  AQUARIUM\_FISH$(x)$ $\Rightarrow$ FISH$(x)$ \\
  		 $\forall x$ &  FLATFISH$(x)$ $\Rightarrow$ FISH$(x)$ \\
  		 $\forall x$ &  RAY$(x)$ $\Rightarrow$ FISH$(x)$ \\
  		 $\forall x$ &  SHARK$(x)$ $\Rightarrow$ FISH $(x)$ \\
  		 $\forall x$ &  TROUT$(x)$ $\Rightarrow$ FISH$(x)$ \\
  		& \\
  		$\forall x$ &  FLOWERS$(x)$ $\Rightarrow$ (ORCHID$(x)$ $\lor$ POPPY$(x)$ $\lor$ ROSE$(x)$ $\lor$ SUNFLOWER$(x)$ $\lor$ TULIP$(x)$) \\
  		$\forall x$ &  ORCHID$(x)$ $\Rightarrow$ FLOWERS$(x)$ \\
  		$\forall x$ &  POPPY$(x)$ $\Rightarrow$ FLOWERS$(x)$ \\
  		$\forall x$ &  ROSE$(x)$ $\Rightarrow$ FLOWERS$(x)$ \\
  		$\forall x$ &  SUNFLOWER$(x)$ $\Rightarrow$ FLOWERS$(x)$ \\
  		$\forall x$ &  TULIP$(x)$ $\Rightarrow$ FLOWERS$(x)$ \\
  		& \\
  		$\forall x$ &  FOOD\_CONTAINERS$(x)$ $\Rightarrow$ (BOTTLE$(x)$ $\lor$ BOWL$(x)$ $\lor$ CAN$(x)$ $\lor$ CUP$(x)$ $\lor$ PLATE$(x)$)  \\
  		$\forall x$ &  BOTTLE$(x)$ $\Rightarrow$ FOOD\_CONTAINERS $(x)$ \\ 
  		$\forall x$ &  BOWL$(x)$ $\Rightarrow$ FOOD\_CONTAINERS $(x)$ \\
  		$\forall x$ &  CAN$(x)$ $\Rightarrow$ FOOD\_CONTAINERS $(x)$ \\ 	
  		$\forall x$ &  CUP$(x)$ $\Rightarrow$ FOOD\_CONTAINERS $(x)$ \\
  		$\forall x$ &  PLATE$(x)$ $\Rightarrow$ FOOD\_CONTAINERS $(x)$ \\
  		& \\
  		$\forall x$ &  FRUIT\_AND\_VEGETABLES$(x)$ $\Rightarrow$ (APPLE$(x)$ $\lor$ MUSHROOM$(x)$ $\lor$ ORANGE$(x)$ $\lor$ PEAR$(x)$ \\
  		& $\lor$ SWEET\_PEPPER$(x)$) \\
  		$\forall x$ &  APPLE$(x)$ $\Rightarrow$ FRUIT\_AND\_VEGETABLES$(x)$ \\ 
  		$\forall x$ &  MUSHROOM$(x)$ $\Rightarrow$ FRUIT\_AND\_VEGETABLES$(x)$ \\ 
  		$\forall x$ &  ORANGE$(x)$ $\Rightarrow$ FRUIT\_AND\_VEGETABLES$(x)$ \\ 
  		$\forall x$ &  PEAR$(x)$ $\Rightarrow$ FRUIT\_AND\_VEGETABLES$(x)$ \\ 
  		$\forall x$ &  SWEET\_PEPPER$(x)$ $\Rightarrow$ FRUIT\_AND\_VEGETABLES$(x)$ \\

  		$\forall x$ &  HOUSEHOLD\_ELECTRICAL\_DEVICES$(x)$ $\Rightarrow$ (CLOCK$(x)$ $\lor$ KEYBOARD$(x)$ $\lor$ LAMP$(x)$ \\
  		& $\lor$ TELEPHONE$(x)$ $\lor$ TELEVISION$(x)$) \\
  		$\forall x$ &  CLOCK$(x)$ $\Rightarrow$ HOUSEHOLD\_ELECTRICAL\_DEVICES$(x)$ \\ 
  		$\forall x$ &  KEYBOARD$(x)$ $\Rightarrow$ HOUSEHOLD\_ELECTRICAL\_DEVICES$(x)$ \\ 
  		$\forall x$ &  LAMP$(x)$ $\Rightarrow$ HOUSEHOLD\_ELECTRICAL\_DEVICES$(x)$ \\ 
  		$\forall x$ &  TELEPHONE$(x)$ $\Rightarrow$ HOUSEHOLD\_ELECTRICAL\_DEVICES$(x)$ \\ 
  		$\forall x$ &  TELEVISION$(x)$ $\Rightarrow$ HOUSEHOLD\_ELECTRICAL\_DEVICES$(x)$\\
  		& \\
  		$\forall x$ &  HOUSEHOLD\_FURNITURE$(x)$ $\Rightarrow$ (BED$(x)$ $\lor$  CHAIR$(x)$ $\lor$ COUCH$(x)$ $\lor$ TABLE$(x)$ $\lor$ WARDROBE$(x)$) \\
  		$\forall x$ &  BED$(x)$ $\Rightarrow$ HOUSEHOLD\_FURNITURE$(x)$ \\ 
  		$\forall x$ &  CHAIR$(x)$ $\Rightarrow$ HOUSEHOLD\_FURNITURE$(x)$ \\ 
  		$\forall x$ &  COUCH$(x)$ $\Rightarrow$ HOUSEHOLD\_FURNITURE$(x)$ \\ 
  		$\forall x$ &  TABLE$(x)$ $\Rightarrow$ HOUSEHOLD\_FURNITURE$(x)$ \\ 
  		$\forall x$ &  WARDROBE$(x)$ $\Rightarrow$ HOUSEHOLD\_FURNITURE$(x)$ \\ 
  		& \\
  		$\forall x$ &  INSECTS$(x)$ $\Rightarrow$ (BEE$(x)$ $\lor$  BEETLE$(x)$ $\lor$  BUTTERFLY$(x)$ $\lor$  CATERPILLAR$(x)$ $\lor$  COCKROACH$(x)$) \\
  		$\forall x$ &  BEE$(x)$ $\Rightarrow$ INSECTS$(x)$ \\ 
  		$\forall x$ &  BEETLE$(x)$ $\Rightarrow$ INSECTS$(x)$ \\ 
  		$\forall x$ &  BUTTERFLY$(x)$ $\Rightarrow$ INSECTS$(x)$ \\ 
  		$\forall x$ &  CATERPILLAR$(x)$ $\Rightarrow$ INSECTS$(x)$ \\ 
   		$\forall x$ &  COCKROACH$(x)$ $\Rightarrow$ INSECTS$(x)$ \\
  		& \\ 
  		$\forall x$ &  LARGE\_CARNIVORES$(x)$ $\Rightarrow$ (BEAR$(x)$ $\lor$ LEOPARD$(x)$ $\lor$ LION$(x)$ $\lor$ TIGER $(x)$ $\lor$ WOLF$(x)$) \\
  		$\forall x$ &  BEAR$(x)$ $\Rightarrow$ LARGE\_CARNIVORES$(x)$ \\ 
  		$\forall x$ &  LEOPARD$(x)$ $\Rightarrow$ LARGE\_CARNIVORES$(x)$ \\ 
  		$\forall x$ &  LION$(x)$ $\Rightarrow$ LARGE\_CARNIVORES$(x)$ \\ 
  		$\forall x$ &  TIGER$(x)$ $\Rightarrow$ LARGE\_CARNIVORES$(x)$ \\ 
  		$\forall x$ &  WOLF$(x)$ $\Rightarrow$ LARGE\_CARNIVORES$(x)$ \\ 
  		& \\
  		$\forall x$ &  LARGE\_MAN-MADE\_OUTDOOR\_THINGS$(x)$ $\Rightarrow$ (BRIDGE$(x)$ $\lor$ CASTLE$(x)$ $\lor$ HOUSE$(x)$ $\lor$ ROAD$(x)$ \\
  		& $\lor$ SKYSCRAPER$(x)$) \\
  		$\forall x$ &  BRIDGE$(x)$ $\Rightarrow$ LARGE\_MAN-MADE\_OUTDOOR\_THINGS$(x)$ \\ 
  		$\forall x$ &  CASTLE$(x)$ $\Rightarrow$ LARGE\_MAN-MADE\_OUTDOOR\_THINGS$(x)$ \\
  		$\forall x$ &  HOUSE$(x)$ $\Rightarrow$ LARGE\_MAN-MADE\_OUTDOOR\_THINGS$(x)$ \\
  		$\forall x$ &  ROAD$(x)$ $\Rightarrow$ LARGE\_MAN-MADE\_OUTDOOR\_THINGS$(x)$ \\ 
  		$\forall x$ &  SKYSCRAPER$(x)$ $\Rightarrow$ LARGE\_MAN-MADE\_OUTDOOR\_THINGS$(x)$ \\

  		$\forall x$ &  LARGE\_NATURAL\_OUTDOOR\_SCENES$(x)$ $\Rightarrow$ (CLOUD$(x)$ $\lor$ FOREST$(x)$ $\lor$ MOUNTAIN$(x)$ \\
  		& $\lor$ PLAIN$(x)$ $\lor$ SEA$(x)$) \\
  		$\forall x$ &  CLOUD$(x)$ $\Rightarrow$ LARGE\_NATURAL\_OUTDOOR\_SCENES$(x)$ \\ 
  		$\forall x$ &  FOREST$(x)$ $\Rightarrow$ LARGE\_NATURAL\_OUTDOOR\_SCENES$(x)$ \\ 
  		$\forall x$ &  MOUNTAIN$(x)$ $\Rightarrow$ LARGE\_NATURAL\_OUTDOOR\_SCENES$(x)$ \\ 
  		$\forall x$ &  PLAIN$(x)$ $\Rightarrow$ LARGE\_NATURAL\_OUTDOOR\_SCENES$(x)$ \\ 
  		$\forall x$ &  SEA$(x)$ $\Rightarrow$ LARGE\_NATURAL\_OUTDOOR\_SCENES$(x)$ \\ 
  		& \\
  		$\forall x$ &  LARGE\_OMNIVORES\_AND\_HERBIVORES$(x)$ $\Rightarrow$ (CAMEL$(x)$ $\lor$  CATTLE$(x)$ $\lor$  CHIMPANZEE$(x)$ \\
  		& $\lor$  ELEPHANT$(x)$ $\lor$  KANGAROO$(x)$)\\
  		$\forall x$ &  CAMEL$(x)$ $\Rightarrow$ LARGE\_OMNIVORES\_AND\_HERBIVORES$(x)$ \\ 
  		$\forall x$ &  CATTLE$(x)$ $\Rightarrow$ LARGE\_OMNIVORES\_AND\_HERBIVORES$(x)$ \\ 
  		$\forall x$ &  CHIMPANZEE$(x)$ $\Rightarrow$ LARGE\_OMNIVORES\_AND\_HERBIVORES$(x)$ \\ 
  		$\forall x$ &  ELEPHANT$(x)$ $\Rightarrow$ LARGE\_OMNIVORES\_AND\_HERBIVORES$(x)$ \\ 
  		$\forall x$ &  KANGAROO$(x)$ $\Rightarrow$ LARGE\_OMNIVORES\_AND\_HERBIVORES$(x)$ \\ 
  		& \\
  		$\forall x$ &  MEDIUM\_MAMMALS$(x)$ $\Rightarrow$ (FOX$(x)$ $\lor$  PORCUPINE$(x)$ $\lor$  POSSUM$(x)$ $\lor$  RACCOON$(x)$ \\
  		& $\lor$  SKUNK$(x)$)\\
  		$\forall x$ &  FOX$(x)$ $\Rightarrow$ MEDIUM\_MAMMALS$(x)$ \\ 
  		$\forall x$ &  PORCUPINE$(x)$ $\Rightarrow$ MEDIUM\_MAMMALS$(x)$ \\ 
  		$\forall x$ &  POSSUM$(x)$ $\Rightarrow$ MEDIUM\_MAMMALS$(x)$ \\ 
  		$\forall x$ &  RACCOON$(x)$ $\Rightarrow$ MEDIUM\_MAMMALS$(x)$ \\ 
  		$\forall x$ &  SKUNK$(x)$ $\Rightarrow$ MEDIUM\_MAMMALS$(x)$ \\ 
  		& \\
  		$\forall x$ &  NON-INSECT\_INVERTEBRATES$(x)$ $\Rightarrow$ (CRAB$(x)$ $\lor$  LOBSTER$(x)$ $\lor$  SNAIL$(x)$ $\lor$ SPIDER$(x)$ \\
  		& $\lor$ WORM$(x)$) \\
  		$\forall x$ &  CRAB$(x)$ $\Rightarrow$ NON-INSECT\_INVERTEBRATES$(x)$ \\ 
  		$\forall x$ &  LOBSTER$(x)$ $\Rightarrow$ NON-INSECT\_INVERTEBRATES$(x)$ \\ 
  		$\forall x$ &  SNAIL$(x)$ $\Rightarrow$ NON-INSECT\_INVERTEBRATES$(x)$ \\ 
  		$\forall x$ &  SPIDER$(x)$ $\Rightarrow$ NON-INSECT\_INVERTEBRATES$(x)$ \\ 
  		$\forall x$ &  WORM$(x)$ $\Rightarrow$ NON-INSECT\_INVERTEBRATES$(x)$ \\
  		& \\
  		$\forall x$ &  PEOPLE$(x)$ $\Rightarrow$ (BABY$(x)$ $\lor$ MAN$(x)$ $\lor$ WOMAN$(x)$ $\lor$ BOY$(x)$ $\lor$ GIRL$(x)$)\\
  		$\forall x$ &  BABY$(x)$ $\Rightarrow$ PEOPLE$(x)$ \\ 
  		$\forall x$ &  BOY$(x)$ $\Rightarrow$ PEOPLE$(x)$ \\ 
  		$\forall x$ &  GIRL$(x)$ $\Rightarrow$ PEOPLE$(x)$ \\

  		$\forall x$ &  MAN$(x)$ $\Rightarrow$ PEOPLE$(x)$ \\ 
  		$\forall x$ &  WOMAN$(x)$ $\Rightarrow$ PEOPLE$(x)$ \\ 
  		& \\
  		$\forall x$ &  REPTILES$(x)$ $\Rightarrow$ (CROCODILE$(x)$ $\lor$  DINOSAUR$(x)$ $\lor$  LIZARD$(x)$ $\lor$ SNAKE$(x)$ $\lor$  TURTLE$(x)$) \\
  		$\forall x$ &  CROCODILE$(x)$ $\Rightarrow$ REPTILES$(x)$ \\ 
  		$\forall x$ &  DINOSAUR$(x)$ $\Rightarrow$ REPTILES$(x)$ \\ 
  		$\forall x$ &  LIZARD$(x)$ $\Rightarrow$ REPTILES$(x)$ \\ 
  		$\forall x$ &  SNAKE$(x)$ $\Rightarrow$ REPTILES$(x)$ \\ 
  		$\forall x$ &  TURTLE$(x)$ $\Rightarrow$ REPTILES$(x)$ \\
  		& \\
  		$\forall x$ &  SMALL\_MAMMALS$(x)$ $\Rightarrow$ (HAMSTER$(x)$ $\lor$  MOUSE$(x)$ $\lor$  RABBIT$(x)$ $\lor$ SHREW$(x)$ $\lor$  SQUIRREL$(x)$)\\
  		$\forall x$ &  HAMSTER$(x)$ $\Rightarrow$ SMALL\_MAMMALS$(x)$ \\ 
  		$\forall x$ &  MOUSE$(x)$ $\Rightarrow$ SMALL\_MAMMALS$(x)$ \\ 
  		$\forall x$ &  RABBIT$(x)$ $\Rightarrow$ SMALL\_MAMMALS$(x)$ \\ 
  		$\forall x$ &  SHREW$(x)$ $\Rightarrow$ SMALL\_MAMMALS$(x)$ \\ 
  		$\forall x$ &  SQUIRREL$(x)$ $\Rightarrow$ SMALL\_MAMMALS$(x)$ \\
  		& \\
  		$\forall x$ &  TREES$(x)$ $\Rightarrow$ (MAPLE\_TREE$(x)$ $\lor$  OAK\_TREE$(x)$ $\lor$  PALM\_TREE$(x)$ $\lor$ PINE\_TREE$(x)$ \\
  		& $\lor$  WILLOW\_TREE$(x)$) \\
  		& \\
  		$\forall x$ &  MAPLE\_TREE$(x)$ $\Rightarrow$ TREES$(x)$ \\ 
  		$\forall x$ &  OAK\_TREE$(x)$ $\Rightarrow$ TREES$(x)$ \\ 
  		$\forall x$ &  PALM\_TREE$(x)$ $\Rightarrow$ TREES$(x)$ \\ 
  		$\forall x$ &  PINE\_TREE$(x)$ $\Rightarrow$ TREES$(x)$ \\ 
  		$\forall x$ &  WILLOW\_TREE$(x)$ $\Rightarrow$ TREE$(x)$ \\
  		& \\
  		$\forall x$ &  VEHICLES1$(x)$ $\Rightarrow$ (BIKE$(x)$ $\lor$  BUS$(x)$ $\lor$  MOTORBIKE$(x)$ $\lor$ PICKUP\_TRUCK$(x)$ $\lor$  TRAIN$(x)$) \\
  		$\forall x$ &  BIKE$(x)$ $\Rightarrow $ VEHICLES1$(x)$ \\ 
  		$\forall x$ &  BUS$(x)$ $\Rightarrow$ VEHICLES1$(x)$ \\ 
  		$\forall x$ &  MOTORBIKE$(x)$ $\Rightarrow$ VEHICLES1$(x)$ \\ 
  		$\forall x$ &  PICKUP$(x)$ $\Rightarrow$ VEHICLES1$(x)$ \\ 
  		$\forall x$ &  TRAIN$(x)$ $\Rightarrow$ VEHICLES1$(x)$ \\ 
  		& \\
  		$\forall x$ &  VEHICLES2$(x)$ $\Rightarrow$ (LAWN MOWER$(x)$ $\lor$  ROCKET$(x)$ $\lor$  STREETCAR$(x)$ $\lor$  TANK$(x)$ $\lor$  TRACTOR$(x)$)\\		
  		$\forall x$ &  LAWN MOWER$(x)$ $\Rightarrow$ VEHICLES2$(x)$ \\ 
  		$\forall x$ &  ROCKET$(x)$ $\Rightarrow$ VEHICLES2$(x)$ \\

  		$\forall x$ &  STREETCAR$(x)$ $\Rightarrow$ VEHICLES2$(x)$ \\ 
  		$\forall x$ &  TANK$(x)$ $\Rightarrow$ VEHICLES2$(x)$ \\ 
  		$\forall x$ &  TRACTOR$(x)$ $\Rightarrow$ VEHICLES2$(x)$ \\ 
  		
  		& \\

  	    \hline
  	    & \\
  	    $\forall x$ &  \verb|mutual_excl|( APPLE$(x)$, AQUARIUM FISH$(x)$, BABY$(x)$, BEAR$(x)$, BEAVER $(x)$,  BED$(x)$, BEE$(x)$,  \\ 
  		& BEETLE$(x)$, BICYCLE$(x)$, BOTTLE$(x)$, BOWL $(x)$, BOY$(x)$, BRIDGE$(x)$, BUS$(x)$,  \\
	  	& BUTTERFLY$(x)$, CAMEL$(x)$, CAN$(x)$, CASTLE$(x)$,   CATERPILLAR$(x)$ , CATTLE$(x)$, CHAIR$(x)$\\ 
	  	& CHIMPANZEE$(x)$,  CLOCK$(x)$, CLOUD$(x)$ , COCKROACH$(x)$, COUCH$(x)$, CRAB$(x)$, \\
	  	& CROCODILE$(x)$ , CUP$(x)$, DINOSAUR$(x)$, DOLPHIN$(x)$, ELEPHANT$(x)$, FLATFISH$(x)$,  \\ 
	  	& FOREST$(x)$, FOX$(x)$, GIRL$(x)$, HAMSTER$(x)$, HOUSE$(x)$, KANGAROO$(x)$, KEYBOARD$(x)$,  \\
	  	& LAMP$(x)$ ,  LAWN\_MOWER$(x)$, LEOPARD$(x)$, LION$(x)$, LIZARD$(x)$, LOBSTER$(x)$, MAN$(x)$,  \\
	  	& MAPLE\_TREE$(x)$ , MOTORCYCLE$(x)$, MOUNTAIN$(x)$, MOUSE$(x)$, MUSHROOM$(x)$,  \\
	  	& OAK\_TREE$(x)$, ORANGE$(x)$, ORCHID$(x)$, OTTER$(x)$, PALM\_TREE$(x)$, PEAR$(x)$,  \\
	  	& PICKUP\_TRUCK$(x)$ , PINE\_TREE$(x)$, PLAIN$(x)$, PLATE$(x)$, POPPY$(x)$, PORCUPINE$(x)$,  \\
	  	& POSSUM$(x)$, RABBIT$(x)$, RACCOON$(x)$, RAY$(x)$, ROAD$(x)$, ROCKET$(x)$, ROSE$(x)$, SEA$(x)$,  \\
	  	& SEAL$(x)$, SHARK$(x)$, SHREW$(x)$, SKUNK$(x)$	$\lor$ SKYSCRAPER$(x)$, SNAIL$(x)$, SNAKE$(x)$,  \\ 
	  	& SPIDER$(x)$, SQUIRREL$(x)$,  STREETCAR$(x)$, SUNFLOWER$(x)$, SWEET\_PEPPER$(x)$, TABLE$(x)$,  \\
	  	& TANK$(x)$, TELEPHONE$(x)$, TELEVISION$(x)$, TIGER$(x)$,  TRACTOR$(x)$, TRAIN$(x)$, TROUT$(x)$,  \\
	  	& TULIP$(x)$, TURTLE$(x)$, WARDROBE$(x)$, WHALE$(x)$, WILLOW\_TREE$(x)$, WOLF$(x)$\\
	  	& WOMAN$(x)$, WORM$(x)$ )\\
  		& \\
  		$\forall x$ &  \verb|mutual_excl|( AQUATIC MAMMALS$(x)$,  FISH$(x)$,  FLOWERS$(x)$, FOOD CONTAINERS$(x)$, \\ 
  		& FRUIT AND VEGETABLES$(x)$,  HOUSEHOLD ELECTRICAL $(x)$,  HOUSEHOLD FURNITURE$(x)$, \\ 
  		& INSECTS$(x)$ ,   LARGE CARNIVORES$(x)$,   MAN-MADE OUTDOOR $(x)$,\\ 
  		& NATURAL OUTDOOR SCENES$(x)$,   OMNIVORES AND HERBIVORES$(x)$,  MEDIUM MAMMALS$(x)$, \\
  		& INVERTEBRATES$(x)$ ,  PEOPLE$(x)$ ,  REPTILES$(x)$ ,  SMALL MAMMALS$(x)$,  TREES$(x)$, \\
  		& VEHICLES1$(x)$,  VEHICLES2$(x)$ )\\
  		& \\
  		\hline
  
\end{longtable}
\label{cifar100rules}
\end{scriptsize}

\begin{table*}[!ht]
	\centering
	\caption{Domain knowledge, PASCAL-Part dataset.}
	\label{pascalpartrules}
	\vskip 2mm
	\begin{scriptsize}
	\begin{tabular}{|ll|}
  		\hline
  		& \\
$\forall x$ & \MakeUppercase{Screen$\MakeLowercase{(x)} \Rightarrow$ (Tvmonitor)} \\
$\forall x$ & \MakeUppercase{Coach$\MakeLowercase{(x)} \Rightarrow$ (Train$\MakeLowercase{(x)} $) } \\
 $\forall x$  & \MakeUppercase{Torso$\MakeLowercase{(x)} \Rightarrow$ (Person$\MakeLowercase{(x)} \lor$  Horse$\MakeLowercase{(x)} \lor$  Cow$\MakeLowercase{(x)} \lor$  Dog$\MakeLowercase{(x)} \lor$  Bird$\MakeLowercase{(x)} \lor$  Cat$\MakeLowercase{(x)} \lor$  Sheep$\MakeLowercase{(x)} $) } \\
$\forall x$  & \MakeUppercase{Leg$\MakeLowercase{(x)} \Rightarrow$ (Person$\MakeLowercase{(x)} \lor$  Horse$\MakeLowercase{(x)} \lor$  Cow$\MakeLowercase{(x)} \lor$  Dog$\MakeLowercase{(x)} \lor$  Bird$\MakeLowercase{(x)} \lor$  Cat$\MakeLowercase{(x)} \lor$  Sheep$\MakeLowercase{(x)} $) } \\
 $\forall x$  & \MakeUppercase{Head$\MakeLowercase{(x)} \Rightarrow$ (Person$\MakeLowercase{(x)} \lor$  Horse$\MakeLowercase{(x)} \lor$  Cow$\MakeLowercase{(x)} \lor$  Dog$\MakeLowercase{(x)} \lor$  Bird$\MakeLowercase{(x)} \lor$  Cat$\MakeLowercase{(x)} \lor$  Sheep$\MakeLowercase{(x)} $) } \\
 $\forall x$  & \MakeUppercase{Ear$\MakeLowercase{(x)} \Rightarrow$ (Person$\MakeLowercase{(x)} \lor$  Horse$\MakeLowercase{(x)} \lor$  Cow$\MakeLowercase{(x)} \lor$  Dog$\MakeLowercase{(x)} \lor$  Cat$\MakeLowercase{(x)} \lor$  Sheep$\MakeLowercase{(x)} $) } \\
 $\forall x$  & \MakeUppercase{Eye$\MakeLowercase{(x)} \Rightarrow$ (Person$\MakeLowercase{(x)} \lor$  Cow$\MakeLowercase{(x)} \lor$  Dog$\MakeLowercase{(x)} \lor$  Bird$\MakeLowercase{(x)} \lor$  Cat$\MakeLowercase{(x)} \lor$  Horse$\MakeLowercase{(x)} \lor$  Sheep$\MakeLowercase{(x)} $) } \\
 $\forall x$  & \MakeUppercase{Ebrow$\MakeLowercase{(x)} \Rightarrow$ (Person$\MakeLowercase{(x)} $) } \\
 $\forall x$  & \MakeUppercase{Mouth$\MakeLowercase{(x)} \Rightarrow$ (Person$\MakeLowercase{(x)} $) } \\
 $\forall x$  & \MakeUppercase{Hair$\MakeLowercase{(x)} \Rightarrow$ (Person$\MakeLowercase{(x)} $) } \\
 $\forall x$  & \MakeUppercase{Nose$\MakeLowercase{(x)} \Rightarrow$ (Person$\MakeLowercase{(x)} \lor$  Dog$\MakeLowercase{(x)} \lor$  Cat$\MakeLowercase{(x)} $) } \\
 $\forall x$  & \MakeUppercase{Neck$\MakeLowercase{(x)} \Rightarrow$ (Person$\MakeLowercase{(x)} \lor$  Horse$\MakeLowercase{(x)} \lor$  Cow$\MakeLowercase{(x)} \lor$  Dog$\MakeLowercase{(x)} \lor$  Bird$\MakeLowercase{(x)} \lor$  Cat$\MakeLowercase{(x)} \lor$  Sheep$\MakeLowercase{(x)} $) } \\
 $\forall x$  & \MakeUppercase{Arm$\MakeLowercase{(x)} \Rightarrow$ (Person$\MakeLowercase{(x)} $) } \\
 $\forall x$  & \MakeUppercase{Muzzle$\MakeLowercase{(x)} \Rightarrow$ (Horse$\MakeLowercase{(x)} \lor$  Cow$\MakeLowercase{(x)} \lor$  Dog$\MakeLowercase{(x)} \lor$  Sheep$\MakeLowercase{(x)} $) } \\
 $\forall x$  & \MakeUppercase{Hoof$\MakeLowercase{(x)} \Rightarrow$ (Horse$\MakeLowercase{(x)} $) } \\
 $\forall x$  & \MakeUppercase{Tail$\MakeLowercase{(x)} \Rightarrow$ (Horse$\MakeLowercase{(x)} \lor$  Cow$\MakeLowercase{(x)} \lor$  Dog$\MakeLowercase{(x)} \lor$  Bird$\MakeLowercase{(x)} \lor$  Sheep$\MakeLowercase{(x)} \lor$  Cat$\MakeLowercase{(x)} \lor$  Aeroplane$\MakeLowercase{(x)} $) } \\
 $\forall x$  & \MakeUppercase{Bottle Body$\MakeLowercase{(x)} \Rightarrow$ (Bottle$\MakeLowercase{(x)} $) } \\
 $\forall x$  & \MakeUppercase{Paw$\MakeLowercase{(x)} \Rightarrow$ (Dog$\MakeLowercase{(x)} \lor$  Cat$\MakeLowercase{(x)} $) } \\
 $\forall x$  & \MakeUppercase{Aeroplane Body$\MakeLowercase{(x)} \Rightarrow$ (Aeroplane$\MakeLowercase{(x)} $) } \\
 $\forall x$  & \MakeUppercase{Wing$\MakeLowercase{(x)} \Rightarrow$ (Aeroplane$\MakeLowercase{(x)} \lor$  Bird$\MakeLowercase{(x)} $) } \\
 $\forall x$  & \MakeUppercase{Wheel$\MakeLowercase{(x)} \Rightarrow$ (Aeroplane$\MakeLowercase{(x)} \lor$  Car$\MakeLowercase{(x)} \lor$  Bicycle$\MakeLowercase{(x)} \lor$  Bus$\MakeLowercase{(x)} \lor$  Motorbike$\MakeLowercase{(x)} $) } \\
 $\forall x$  & \MakeUppercase{Stern$\MakeLowercase{(x)} \Rightarrow$ (Aeroplane$\MakeLowercase{(x)} $) } \\
 $\forall x$  & \MakeUppercase{Cap$\MakeLowercase{(x)} \Rightarrow$ (Bottle$\MakeLowercase{(x)} $) } \\
 $\forall x$  & \MakeUppercase{Hand$\MakeLowercase{(x)} \Rightarrow$ (Person$\MakeLowercase{(x)} $) } \\
 $\forall x$  & \MakeUppercase{Frontside$\MakeLowercase{(x)} \Rightarrow$ (Car$\MakeLowercase{(x)} \lor$  Bus$\MakeLowercase{(x)} \lor$  Train$\MakeLowercase{(x)} $) } \\
 $\forall x$  & \MakeUppercase{Rightside$\MakeLowercase{(x)} \Rightarrow$ (Car$\MakeLowercase{(x)} \lor$  Bus$\MakeLowercase{(x)} \lor$  Train$\MakeLowercase{(x)} $) } \\
 $\forall x$  & \MakeUppercase{Roofside$\MakeLowercase{(x)} \Rightarrow$ (Car$\MakeLowercase{(x)} \lor$  Bus$\MakeLowercase{(x)} \lor$  Train$\MakeLowercase{(x)} $) } \\
 $\forall x$  & \MakeUppercase{Backside$\MakeLowercase{(x)} \Rightarrow$ (Car$\MakeLowercase{(x)} \lor$  Bus$\MakeLowercase{(x)} \lor$  Train$\MakeLowercase{(x)} $) } \\
 $\forall x$  & \MakeUppercase{Leftside$\MakeLowercase{(x)} \Rightarrow$ (Car$\MakeLowercase{(x)} \lor$  Train$\MakeLowercase{(x)} \lor$  Bus$\MakeLowercase{(x)} $) } \\
 $\forall x$  & \MakeUppercase{Door$\MakeLowercase{(x)} \Rightarrow$ (Car$\MakeLowercase{(x)} \lor$  Bus$\MakeLowercase{(x)} $) } \\
 $\forall x$  & \MakeUppercase{Mirror$\MakeLowercase{(x)} \Rightarrow$ (Car$\MakeLowercase{(x)} \lor$  Bus$\MakeLowercase{(x)} $) } \\
 $\forall x$  & \MakeUppercase{Headlight$\MakeLowercase{(x)} \Rightarrow$ (Car$\MakeLowercase{(x)} \lor$  Bus$\MakeLowercase{(x)} \lor$  Train$\MakeLowercase{(x)} \lor$  Motorbike$\MakeLowercase{(x)} \lor$  Bicycle$\MakeLowercase{(x)} $) } \\
 $\forall x$  & \MakeUppercase{Motorbike$\MakeLowercase{(x)} \Rightarrow$ (Wheel$\MakeLowercase{(x)} \lor$  Headlight$\MakeLowercase{(x)} \lor$  Handlebar$\MakeLowercase{(x)} \lor$   Saddle$\MakeLowercase{(x)} $) } \\
 $\forall x$  & \MakeUppercase{Window$\MakeLowercase{(x)} \Rightarrow$ (Car$\MakeLowercase{(x)} \lor$  Bus$\MakeLowercase{(x)} $) } \\
 $\forall x$  & \MakeUppercase{Plate$\MakeLowercase{(x)} \Rightarrow$ (Car$\MakeLowercase{(x)} \lor$  Bus$\MakeLowercase{(x)} $) } \\
 $\forall x$  & \MakeUppercase{Engine$\MakeLowercase{(x)} \Rightarrow$ (Aeroplane$\MakeLowercase{(x)} $) } \\
 $\forall x$  & \MakeUppercase{Foot$\MakeLowercase{(x)} \Rightarrow$ (Person$\MakeLowercase{(x)} \lor$  Bird$\MakeLowercase{(x)} $) } \\
 $\forall x$  & \MakeUppercase{Chainwheel$\MakeLowercase{(x)} \Rightarrow$ (Bicycle$\MakeLowercase{(x)} $) } \\
 $\forall x$  & \MakeUppercase{Saddle$\MakeLowercase{(x)} \Rightarrow$ (Bicycle$\MakeLowercase{(x)} \lor$  Motorbike$\MakeLowercase{(x)} $) } \\
 $\forall x$  & \MakeUppercase{Handlebar$\MakeLowercase{(x)} \Rightarrow$ (Bicycle$\MakeLowercase{(x)} \lor$  Motorbike$\MakeLowercase{(x)} $) } \\
 $\forall x$  & \MakeUppercase{Train Head$\MakeLowercase{(x)} \Rightarrow$ (Train$\MakeLowercase{(x)} $) } \\
 $\forall x$  & \MakeUppercase{Beak$\MakeLowercase{(x)} \Rightarrow$ (Bird$\MakeLowercase{(x)} $) } \\
 $\forall x$  & \MakeUppercase{Pot$\MakeLowercase{(x)} \Rightarrow$ (Pottedplant$\MakeLowercase{(x)} $) } \\
 $\forall x$  & \MakeUppercase{Plant$\MakeLowercase{(x)} \Rightarrow$ (Pottedplant$\MakeLowercase{(x)} $) } \\
 $\forall x$  & \MakeUppercase{Horn$\MakeLowercase{(x)} \Rightarrow$ (Cow$\MakeLowercase{(x)} \lor$  Sheep$\MakeLowercase{(x)} $) } \\
& \\
\hline
& \\

$\forall x$ & \MakeUppercase{Tvmonitor$\MakeLowercase{(x)} \Rightarrow$ (Screen$\MakeLowercase{(x)} $) } \\
$\forall x$ & \MakeUppercase{Train$\MakeLowercase{(x)} \Rightarrow$ (Coach$\MakeLowercase{(x)} \lor$  Leftside$\MakeLowercase{(x)} \lor$  Train Head$\MakeLowercase{(x)} \lor$  Headlight$\MakeLowercase{(x)} \lor$  Frontside$\MakeLowercase{(x)}$} \\
& $\lor$\MakeUppercase{Rightside$\MakeLowercase{(x)} \lor$  Backside$\MakeLowercase{(x)} \lor$ Roofside$\MakeLowercase{(x)} $) } \\
$\forall x$ & \MakeUppercase{Person$\MakeLowercase{(x)} \Rightarrow$ (Torso$\MakeLowercase{(x)} \lor$  Leg$\MakeLowercase{(x)} \lor$  Head$\MakeLowercase{(x)} \lor$  Ear$\MakeLowercase{(x)} \lor$  Eye$\MakeLowercase{(x)} \lor$  Ebrow$\MakeLowercase{(x)} \lor$  Mouth$\MakeLowercase{(x)} \lor$  Hair$\MakeLowercase{(x)}$} \\  
& $\lor$\MakeUppercase{Nose$\MakeLowercase{(x)} \lor$  Neck$\MakeLowercase{(x)} \lor$ Arm$\MakeLowercase{(x)} \lor$  Hand$\MakeLowercase{(x)} \lor$  Foot$\MakeLowercase{(x)} $) } \\
$\forall x$ & \MakeUppercase{Horse$\MakeLowercase{(x)} \Rightarrow$ (Head$\MakeLowercase{(x)} \lor$  Ear$\MakeLowercase{(x)} \lor$  Muzzle$\MakeLowercase{(x)} \lor$  Torso$\MakeLowercase{(x)} \lor$  Neck$\MakeLowercase{(x)} \lor$  Leg$\MakeLowercase{(x)} \lor$  Hoof$\MakeLowercase{(x)} \lor$  Tail$\MakeLowercase{(x)} \lor$  Eye$\MakeLowercase{(x)} $) } \\
$\forall x$ & \MakeUppercase{Cow$\MakeLowercase{(x)} \Rightarrow$ (Head$\MakeLowercase{(x)} \lor$  Ear$\MakeLowercase{(x)} \lor$  Eye$\MakeLowercase{(x)} \lor$  Muzzle$\MakeLowercase{(x)} \lor$  Torso$\MakeLowercase{(x)} \lor$  Neck$\MakeLowercase{(x)} \lor$  Leg$\MakeLowercase{(x)} \lor$  Tail$\MakeLowercase{(x)} \lor$  Horn$\MakeLowercase{(x)} $) } \\
$\forall x$ & \MakeUppercase{Bottle$\MakeLowercase{(x)} \Rightarrow$ (Bottle Body$\MakeLowercase{(x)} \lor$  Cap$\MakeLowercase{(x)} $) } \\
$\forall x$ & \MakeUppercase{Dog$\MakeLowercase{(x)} \Rightarrow$ (Head$\MakeLowercase{(x)} \lor$  Ear$\MakeLowercase{(x)} \lor$  Torso$\MakeLowercase{(x)} \lor$  Neck$\MakeLowercase{(x)} \lor$  Leg$\MakeLowercase{(x)} \lor$  Paw$\MakeLowercase{(x)} \lor$  Eye$\MakeLowercase{(x)} \lor$  Muzzle$\MakeLowercase{(x)} $ } \\
& $\lor$ \MakeUppercase{Nose$\MakeLowercase{(x)} \lor$  Tail$\MakeLowercase{(x)} $) } \\
$\forall x$ & \MakeUppercase{Aeroplane$\MakeLowercase{(x)} \Rightarrow$ (Aeroplane Body$\MakeLowercase{(x)} \lor$  Wing$\MakeLowercase{(x)} \lor$  Wheel$\MakeLowercase{(x)} \lor$  Stern$\MakeLowercase{(x)} \lor$  Engine$\MakeLowercase{(x)} \lor$  Tail$\MakeLowercase{(x)} $) } \\
$\forall x$ & \MakeUppercase{Car$\MakeLowercase{(x)} \Rightarrow$ (Frontside$\MakeLowercase{(x)} \lor$  Rightside$\MakeLowercase{(x)} \lor$  Door$\MakeLowercase{(x)} \lor$  Mirror$\MakeLowercase{(x)} \lor$  Headlight$\MakeLowercase{(x)} \lor$  Wheel$\MakeLowercase{(x)} $}\\
& $\lor$ \MakeUppercase{ Window$\MakeLowercase{(x)} \lor$  Plate$\MakeLowercase{(x)} \lor $ Roofside$\MakeLowercase{(x)} \lor$  Backside$\MakeLowercase{(x)} \lor$  Leftside$\MakeLowercase{(x)} $) } \\
$\forall x$ & \MakeUppercase{Bus$\MakeLowercase{(x)} \Rightarrow$ (Plate$\MakeLowercase{(x)} \lor$  Frontside$\MakeLowercase{(x)} \lor$  Rightside$\MakeLowercase{(x)} \lor$  Door$\MakeLowercase{(x)} \lor$  Mirror$\MakeLowercase{(x)} \lor$  Headlight$\MakeLowercase{(x)} $} \\ 
& $\lor$  \MakeUppercase{Window$\MakeLowercase{(x)} \lor$  Wheel$\MakeLowercase{(x)} \lor$ Leftside$\MakeLowercase{(x)} \lor$  Backside$\MakeLowercase{(x)} \lor$  Roofside$\MakeLowercase{(x)} $) } \\
$\forall x$ & \MakeUppercase{Bicycle$\MakeLowercase{(x)} \Rightarrow$ (Wheel$\MakeLowercase{(x)} \lor$  Chainwheel$\MakeLowercase{(x)} \lor$  Saddle$\MakeLowercase{(x)} \lor$  Handlebar$\MakeLowercase{(x)} \lor$  Headlight$\MakeLowercase{(x)} $) } \\
$\forall x$ & \MakeUppercase{Bird$\MakeLowercase{(x)} \Rightarrow$ (Head$\MakeLowercase{(x)} \lor$  Eye$\MakeLowercase{(x)} \lor$  Beak$\MakeLowercase{(x)} \lor$  Torso$\MakeLowercase{(x)} \lor$  Neck$\MakeLowercase{(x)} \lor$  Leg$\MakeLowercase{(x)} \lor$  Foot$\MakeLowercase{(x)} \lor$  Tail$\MakeLowercase{(x)} \lor$  Wing$\MakeLowercase{(x)} $) } \\
$\forall x$ & \MakeUppercase{Cat$\MakeLowercase{(x)} \Rightarrow$ (Head$\MakeLowercase{(x)} \lor$  Ear$\MakeLowercase{(x)} \lor$  Eye$\MakeLowercase{(x)} \lor$  Nose$\MakeLowercase{(x)} \lor$  Torso$\MakeLowercase{(x)} \lor$  Neck$\MakeLowercase{(x)} \lor$  Leg$\MakeLowercase{(x)} \lor$  Paw$\MakeLowercase{(x)} \lor$  Tail$\MakeLowercase{(x)} $) } \\
$\forall x$ & \MakeUppercase{Motorbike$\MakeLowercase{(x)} \Rightarrow$ (Wheel$\MakeLowercase{(x)} \lor$  Headlight$\MakeLowercase{(x)} \lor$  Handlebar$\MakeLowercase{(x)} \lor$  Saddle$\MakeLowercase{(x)} $) } \\
$\forall x$ & \MakeUppercase{Sheep$\MakeLowercase{(x)} \Rightarrow$ (Head$\MakeLowercase{(x)} \lor$  Ear$\MakeLowercase{(x)} \lor$  Eye$\MakeLowercase{(x)} \lor$  Muzzle$\MakeLowercase{(x)} \lor$  Torso$\MakeLowercase{(x)} \lor$  Neck$\MakeLowercase{(x)} \lor$  Leg$\MakeLowercase{(x)} \lor$  Tail$\MakeLowercase{(x)} \lor$  Horn$\MakeLowercase{(x)} $) } \\
$\forall x$ & \MakeUppercase{Pottedplant$\MakeLowercase{(x)} \Rightarrow$ (Pot$\MakeLowercase{(x)} \lor$  Plant$\MakeLowercase{(x)} $) } \\
& \\
\hline
& \\
$\forall x $ & \MakeUppercase{ Tvmonitor$\MakeLowercase{(x)} \lor $  Train$\MakeLowercase{(x)} \lor $  Person$\MakeLowercase{(x)} \lor $  Boat$\MakeLowercase{(x)} \lor $  Horse$\MakeLowercase{(x)} \lor $  Cow$\MakeLowercase{(x)} \lor $  Bottle$\MakeLowercase{(x)} \lor $  Dog$\MakeLowercase{(x)} $} \\
& \MakeUppercase{$\lor $  Aeroplane$\MakeLowercase{(x)} \lor $  Car$\MakeLowercase{(x)} \lor $  Bus$\MakeLowercase{(x)} \lor $  Bicycle$\MakeLowercase{(x)} \lor $  Table$\MakeLowercase{(x)} \lor $  Chair$\MakeLowercase{(x)} \lor $  Bird$\MakeLowercase{(x)} \lor $  Cat$\MakeLowercase{(x)} $} \\
& \MakeUppercase{$\lor $  Motorbike$\MakeLowercase{(x)} \lor $  Sheep$\MakeLowercase{(x)} \lor $  Sofa$\MakeLowercase{(x)} \lor $  Pottedplant$\MakeLowercase{(x)}$}\\
& \\
\hline  
\end{tabular}
\end{scriptsize}
\end{table*}
\clearpage

\begin{table*}[!ht]
	\centering
	\caption{\minor{First noisy domain knowledge ($\mathcal{\tilde{K}}_a$), ANIMALS dataset, obtained by altering the clean knowledge of Table~\ref{animalsrules}. We report only the altered rules, highlighting the changes that make them not-coherent with the ANIMALS domain.}}
	\label{tab:ka}
	\vskip 2mm
	\begin{scriptsize}
	\minor{\begin{tabular}{|ll|}
		\hline
		& \\
		$\forall x$ &  FEATHER$(x)$ $\Rightarrow$ \sout{BIRD$(x)$} \textbf{MAMMAL}$\textbf{(x)}$ \\
		$\forall x$ &  MAMMAL$(x)$ $\land$ \sout{MEAT$(x)$} \textbf{BIRD}$\textbf{(x)}$ $\Rightarrow$ CARNIVORE$(x)$\\
		$\forall x$ &  CARNIVORE$(x)$ $\land$ TAWNY$(x)$  $\land$ \sout{DARKSPOTS$(x)$} $\Rightarrow$ CHEETAH$(x)$\\
        $\forall x$ &  BLACKSTRIPES$(x)$ $\land$ UNGULATE$(x)$  $\land$ WHITE$(x)$ $\Rightarrow$ \sout{ZEBRA$(x)$} \textbf{TIGER}$\textbf{(x)}$\\
		& \\
		\hline
	\end{tabular}}
	\end{scriptsize}
\end{table*}

\begin{table*}[!ht]
	\centering
	\caption{\minor{Second noisy domain knowledge ($\mathcal{\tilde{K}}_b$), ANIMALS dataset, obtained by adding new rules to the clean knowledge of Table~\ref{animalsrules}. We report only the added rules, that were explicitly created to be not-coherent with the ANIMALS domain.}}
	\label{tab:kb}
	\vskip 2mm
	\begin{scriptsize}
	\minor{\begin{tabular}{|ll|}
		\hline
		& \\
		$\forall x$ &  FLY$(x)$ $\Rightarrow $ MAMMAL$(x)$\\
		$\forall x$ &  MAMMAL$(x)$ $\land$ EVENTOED$(x)$ $\Rightarrow$ FEATHER$(x)$\\
		$\forall x$ &  BLACKSTRIPES$(x)$ $\land$ WHITE$(x)$ $\Rightarrow$ PENGUIN$(x)$\\
		$\forall x$ &  CARNIVORE$(x)$ $\land$ DARKSPOTS$(x)$ $\Rightarrow$ TIGER$(x)$\\
		& \\
		\hline
	\end{tabular}}
	\end{scriptsize}
\end{table*}

\begin{table*}[!ht]
	\centering
	\caption{\minor{Third noisy domain knowledge ($\mathcal{\tilde{K}}_c$), ANIMALS dataset,  obtained by altering the clean knowledge of Table~\ref{animalsrules}. We report only the altered rules, highlighting the changes that make them not-fully-coherent with the ANIMALS domain. They all involve main-class-oriented conclusions.}}
	\label{tab:kc}
	\vskip 2mm
	\begin{scriptsize}
	\minor{\begin{tabular}{|ll|}
		\hline
		& \\
		$\forall x$ &  CARNIVORE$(x)$ $\land$ TAWNY$(x)$  $\land$ DARKSPOTS$(x)$ $\Rightarrow$ (CHEETAH$(x)$ \textbf{$\lor$}  \textbf{GIRAFFE}$\textbf{(x)}$)\\
        $\forall x$ &  UNGULATE$(x)$ $\land$ LONGLEGS$(x)$  $\land$ LONGNECK$(x)$  $\land$ TAWNY$(x)$  $\land$ DARKSPOTS$(x)$ $\Rightarrow$ (GIRAFFE$(x)$ \textbf{$\lor$}  \textbf{ZEBRA}$\textbf{(x)}$)\\
        $\forall x$ &  BLACKSTRIPES$(x)$ $\land$ UNGULATE$(x)$  $\land$ WHITE$(x)$ $\Rightarrow$ (ZEBRA$(x)$ \textbf{$\lor$}  \textbf{TIGER}$\textbf{(x)}$)\\
        $\forall x$ &  BIRD$(x)$ $\land$ $\neg$FLY$(x)$  $\land$ SWIM$(x)$  $\land$ BLACKWHITE$(x)$ $\Rightarrow$ (PENGUIN$(x)$ \textbf{$\lor$}  \textbf{OSTRICH}$\textbf{(x)}$)\\
        & \\
        \hline
	\end{tabular}}
	\end{scriptsize}
\end{table*}


%



%


\end{document}